\documentclass[10pt,twoside]{report} 
\usepackage[latin1]{inputenc} 
\usepackage[dvips]{graphicx}
\usepackage[finnish,english]{babel} 
\usepackage{bibalias,amsmath,amssymb}
\usepackage[footnotesize]{caption}
\usepackage[b5paper,dvips]{geometry} 
\usepackage[round]{natbib}
\usepackage{fancyheadings} 
\usepackage{chngpage}
\usepackage{setspace}
\usepackage{verbatim} 
\usepackage{url}

\def\B{\mathbf{B}}

\def\E{\mathbb{E}}

\def\I{\mathbf{I}}

\def\L{{\cal L}}
\def\M{\mathbf{M}}
\def\N{\mathcal{N}}
\def\Nd{N}
\def\Nm{\mathcal{N}_m}

\def\Rn{R^{(n)}}

\def\T{\mathbf{T}}
\def\U{\mathbf{U}}
\def\Vb{\mathbf{V}}
\def\X{\mathbf{X}}
\def\Xcal{\mathcal{X}}
\def\Y{\mathbf{Y}}
\def\Ycal{\mathcal{Y}}
\def\Z{\mathbf{Z}}
\def\W{\mathbf{W}}
\def\Wx{\mathbf{W}_x}
\def\Wy{\mathbf{W}_y}

\def\d{\mathbf{d}}
\def\dt{d_t}

\def\gi{g_i}
\def\gt{g_t}
\def\gc{g_c}

\def\m{\mathbf{m}}
\def\mj{\mathbf{m}_j}

\def\mtj{m_{tj}}
\def\n{\mathbf{n}}

\def\srn{\s_{r}^{(n)}}
\def\s{\mathbf{s}}

\def\sij{s_{ij}}

\def\stj{s_{tj}}
\def\scj{s_{cj}}

\def\vx{\mathbf{v}_x}
\def\vy{\mathbf{v}_y}

\def\wrn{w_{r}^{(n)}}
\def\x{\mathbf{x}}
\def\xn{\x^{(n)}}

\def\xi{\x_i}
\def\y{\mathbf{y}}
\def\yi{\y_i}

\def\z{\mathbf{z}}

\def\alp{{\boldsymbol \alpha}}
\def\alphaj{\alpha_j}
\def\betaj{\beta_j}
\def\alphahatj{\hat{\alpha}_j}

\def\betahatj{\hat{\beta}_j}

\def\epsilonij{\mathbf{\varepsilon}_{ij}}
\def\epsiloncj{\varepsilon_{cj}}
\def\epsilontj{\varepsilon_{tj}}

\def\Epsx{\boldsymbol{\varepsilon}_x}
\def\Epsy{\boldsymbol{\varepsilon}_y}

\def\Epsilon{\boldsymbol{\varepsilon}}
\def\Epsilonj{\boldsymbol{\varepsilon}_j}

\def\bth{\boldsymbol{\theta}}
\def\lambdaj{\lambda_j}

\def\bmu{\boldsymbol{\mu}}

\def\muj{\mathbf{\mu}_j}

\def\Psib{\mathbf{\Psi}}

\def\bpi{{\boldsymbol \pi}}

\def\bSigma{\boldsymbol{\Sigma}}
\def\sigt{\sigma_T^2}

\def\Sigrn{\bSigma_{r}^{(n)}}

\def\TauSq{\boldsymbol{\tau}^2}

\def\taus{\{\tau_j^2\}}
\def\taujSq{\mathbf{\tau}_j^2}

\def\0{\mathbf{0}}

\def\Real{\mathbb{R}}

\begin{document}

\thispagestyle{empty}

\lhead[\fancyplain{}{\small\sl\leftmark}]{}
\rhead[]{\fancyplain{}{\small\sl\rightmark}}

\thispagestyle{empty}

 \noindent
 {\small TKK Dissertations in Information and Computer Science}\\
 {\small Espoo 2010 \hfill TKK-ICS-D19}\\
\vspace*{60mm}

{\noindent\large\bf PROBABILISTIC ANALYSIS OF THE HUMAN\\TRANSCRIPTOME
WITH SIDE INFORMATION}

 \vspace*{2mm}

 \noindent
 Leo Lahti\\
 \vspace*{12.4mm}

{\noindent\small Dissertation for the degree of Doctor of Science in
Technology to be presented with due permission of the Faculty of
Information and Natural Sciences for public examination and debate in
Auditorium AS1 at the Aalto University School of Science and
Technology (Espoo, Finland) on the 17th of December 2010 at 13
o'clock.}

\vspace*{\stretch{1}}
\vspace*{5mm}
 \noindent
Aalto University School of Science and Technology\\
Faculty of Information and Natural Sciences\\
Department of Information and Computer Science\\

 \noindent
Aalto-yliopiston teknillinen korkeakoulu\\
Informaatio- ja luonnontieteiden tiedekunta\\
Tietojenkäsittelytieteen laitos\\

\newpage
\thispagestyle{empty}
\vspace*{58mm}

\noindent
Distribution:\\[2mm]
Aalto University School of Science and Technology\\
Faculty of Information and Natural Sciences\\
Department of Information and Computer  Science\\
P.O.Box 15400\\
FI-00076 Aalto \\
FINLAND\\[2mm]
Tel. +358-9-470 23272\\
Fax  +358-9-470 23277\\
Email: {\tt series@ics.tkk.fi}\\[2mm]

\noindent
Copyright \copyright 2010 Leo Lahti\\
First Edition. Some Rights Reserved.\\
http://www.iki.fi/Leo.Lahti (leo.lahti@iki.fi)\\[2mm]

\noindent
\includegraphics[width=4.5cm]{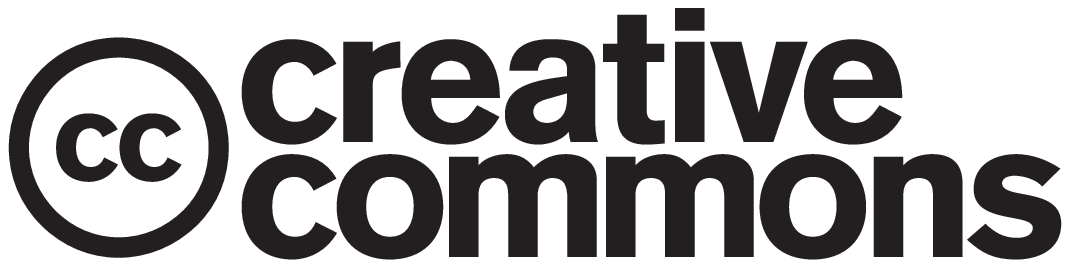}

\noindent This thesis is licensed under the terms of {\it Creative
  Commons Attribution 3.0 Unported} license available from
  http://www.creativecommons.org/. Accordingly, you are free to copy,
  distribute, display, perform, remix, tweak, and build upon this work
  even for commercial purposes, assuming that you give the original
  author credit. See the licensing terms for details. For Appendices
  and Figures, consult the separate copyright notices.\\[2mm]

\noindent
ISBN 978-952-60-3367-9 (Print)\\
ISBN 978-952-60-3368-6 (Online)\\
ISSN 1797-5050 (Print)\\
ISSN 1797-5069 (Online)\\
URL:  {\tt http://lib.tkk.fi/Diss/2010/isbn9789526033686/}\\

\noindent
Multiprint Oy\\
Espoo 2010

\newpage

\pagenumbering{roman}

\thispagestyle{empty}
\noindent{{\bf {\Large ABSTRACT}}}
\vspace{0.5cm}

\noindent Lahti, L. (2010): \newblock {\bf Probabilistic analysis of
the human transcriptome with side information} Doctoral thesis, Aalto
University School of Science and Technology, Dissertations in
Information and Computer Science, TKK-ICS-D19, Espoo, Finland.\\

\noindent \textbf{Keywords:} data integration, exploratory data
analysis, functional genomics, probabilistic modeling,
transcriptomics\\

Recent advances in high-throughput measurement technologies and
efficient sharing of biomedical data through community databases have
made it possible to investigate the complete collection of genetic
material, the genome, which encodes the heritable genetic program of an
organism. This has opened up new views to the study of living
organisms with a profound impact on biological research.

Functional genomics is a subdiscipline of molecular biology that
investigates the functional organization of genetic information.  This
thesis develops computational strategies to investigate a key
functional layer of the genome, the transcriptome. The time- and
context-specific transcriptional activity of the genes regulates the
function of living cells through protein synthesis. Efficient
computational techniques are needed in order to extract useful
information from high-dimensional genomic observations that are
associated with high levels of complex variation. Statistical learning
and probabilistic models provide the theoretical framework for
combining statistical evidence across multiple observations and the
wealth of background information in genomic data repositories.

This thesis addresses three key challenges in transcriptome
analysis. First, new preprocessing techniques that utilize side
information in genomic sequence databases and microarray collections
are developed to improve the accuracy of high-throughput microarray
measurements.  Second, a novel exploratory approach is proposed in
order to construct a global view of cell-biological network activation
patterns and functional relatedness between tissues across normal
human body. Information in genomic interaction databases is used to
derive constraints that help to focus the modeling in those parts of
the data that are supported by known or potential interactions between
the genes, and to scale up the analysis. The third contribution is to
develop novel approaches to model dependency between co-occurring
measurement sources. The methods are used to study cancer mechanisms
and transcriptome evolution; integrative analysis of the human
transcriptome and other layers of genomic information allows the
identification of functional mechanisms and interactions that could
not be detected based on the individual measurement sources. Open
source implementations of the key methodological contributions have
been released to facilitate their further adoption by the research
community.

\newpage

\thispagestyle{empty}
\noindent{{\bf {\Large TIIVISTELMÄ}}}
\vspace{0.5cm}

\noindent Lahti, L. (2010): \newblock {\bf Ihmisen geenien
ilmentymisen ja taustatiedon tilastolli\-nen mallitus} Väitöskirja,
Aalto-yliopiston teknillinen korkeakoulu, Dissertations in Information
and Computer Science, TKK-ICS-D19, Espoo, Suomi.\\[2mm]

\noindent
\textbf{Avainsanat:} aineistojen yhdistely, data-analyysi,
toiminnallinen genomiikka, tilastollinen mallitus, geenien
ilmentyminen\\

Mittausmenetelmien kehitys ja tutkimustiedon laajentunut saatavuus
ovat mahdollistaneet ihmisen perimän eli genomin kokonaisvaltaisen
tarkastelun. Tämä on avannut uusia nä\-kö\-kul\-mi\-a biologiseen
tutkimukseen ja auttanut ymmärtämään elämän syntyä ja rakennetta uusin
tavoin. Toiminnallinen genomiikka on molekyylibiologian osa-alue, joka
tutkii perimän toiminnallisia ominaisuuksia. Perimän toimintaan
liittyvää mittausaineistoa on runsaasti saatavilla, mutta
korkeaulotteisiin mittauksiin liittyy monimutkaisia ja tuntemattomia
taus\-ta\-te\-ki\-jöi\-tä, joi\-den huomiointi mallituksessa on
haasteellista.  Tehokkaat laskennalliset menetelmät ovat avainasemassa
py\-rit\-tä\-es\-sä jalostamaan uusista havainnoista käyttökelpoista
tietoa.

Tässä väitöskirjassa on kehitetty yleiskäyttöisiä laskennallisia
menetelmiä, joilla voidaan tutkia ihmisen geenien ilmentymistä koko
perimän tasolla.  Geenien ilmentyminen viittaa lähetti-RNA-molekyylien
tuottoon solussa perimän sisältämän informaation nojalla. Tämä on
keskeinen perinnöllisen informaation säätelytaso, jonka avulla solu
säätelee proteiinien tuottoa ja solun toi\-min\-taa ajasta ja
tilanteesta riippuen. Tilastollinen oppiminen ja todennäköisyyksin
perustuva probabilistinen mallitus tarjoavat teoreettisen kehyksen,
jonka avulla rinnakkaisiin mittauksiin ja taustatietoihin sisältyvää
informaatiota voidaan käyttää kasvattamaan mallien tilastollista
voimaa. Kehitetyt menetelmät ovat yleiskäyttöisiä laskennallisen
tieteen tutkimusvälineitä, jotka tekevät vähän, mutta selkeästi
ilmaistuja mallitusoletuksia ja sietävät korkeaulotteisiin
toiminnallisen genomiikan havaintoaineistoihin sisältyviä
epävarmuuksia.

Väitöskirjassa kehitetyt menetelmät tarjoavat ratkaisuja kolmeen
keskeiseen mallitusongelmaan toiminnallisessa
genomiikassa. Luotettavien esi\-kä\-sit\-te\-ly\-me\-ne\-tel\-mi\-en
kehittäminen on työn ensimmäinen päätulos, jossa tietokantoihin
si\-säl\-ty\-vää taustatietoa käytetään perimänlaajuisten
mittausaineistojen epä\-var\-muuk\-si\-en
vä\-hen\-tä\-mi\-sek\-si. Toisena päätuloksena väitöskirjassa
kehitetään uusi ali\-ava\-ruus\-ka\-sa\-u\-tuk\-seen perustuva
me\-ne\-tel\-mä, jonka avulla voidaan tutkia ja kuvata solubiologisen
vuorovaikutusverkon käyttäytymistä kokonaisvaltaisesti ihmiskehon eri
osissa. Taustatietoa geenien vuorovaikutuksista käy\-te\-tään
ohjaamaan ja nopeuttamaan mallitusta. Me\-ne\-tel\-mäl\-lä saadaan
uutta tietoa geenien säätelystä ja kudosten toiminnallisista
yhteyksistä. Kolmanneksi väitöskirjatyös\-sä ke\-hi\-te\-tään uusia
me\-ne\-tel\-miä perimänlaajuisten mittausaineistojen yhdistelyyn.
Ihmisen geenien ilmentymisen ja muiden aineistojen riippuvuuksien
mallitus mahdollistaa sellaisten toiminnallisten yhteyksien ja
vuorovaikutusten havaitsemisen, joiden tutkimiseksi yksittäiset
havaintoaineistot ovat riit\-tä\-mät\-tö\-miä. Aineistojen yhdistelyyn
ke\-hi\-tet\-ty\-jä me\-ne\-tel\-mi\-ä sovelletaan
syö\-pä\-me\-ka\-nis\-mi\-en ja lajien välisten eroavaisuuksien
tutkimiseen. Jul\-kais\-tuil\-la avoimen läh\-de\-koo\-din
toteutuksilla on pyritty varmistamaan kehitettyjen menetelmien
saatavuus ja laajempi käyttöönotto laskennallisen biologian
tutkimuksessa.

\newpage

\setcounter{page}{3}

{\small \tableofcontents}\newpage

\pagestyle{fancy}

\chapter*{Preface}
\addcontentsline{toc}{section}{Preface}
\markboth{PREFACE}{PREFACE}

This work has been carried out at the Neural Networks Research Centre
and Adaptive Informatics Research Centre of the Laboratory of Computer
and Information Science (Department of Information and Computer
Science since 2008), Helsinki University of Technology, i.e., as of
2010 the Aalto University School of Science and Technology. Part of
the work was done at the Department of Computer Science, University of
Helsinki, when I was visiting there for a year in 2005. I am also
pleased to having had the opportunity to be a part of the Helsinki
Institute for Information Technology HIIT. The work has been supported
by the Graduate School of Computer Science and Engineering, as well as
by project funding from the Academy of Finland through the SYSBIO
program and from TEKES through the MultiBio research consortium. The
Graduate School in Computational Biology, Bioinformatics, and Biometry
(ComBi) has supported my participation to scientific conferences and
workshops abroad during the thesis work.

I wish to thank my supervisor, professor Samuel Kaski for giving me
the opportunity to work in a truly interdisciplinary research field
with the freedom and responsibilities of scientific work, and with the
necessary amount of guidance. These have been essential parts of the
learning process.

I would also like to express my gratitude to the reviewers of this
thesis, Professor Juho Rousu and Doctor Simon Rogers for their expert
feedback.

Research on computational biology has given me the excellent
opportunity to work with and learn from experts in two traditionally
distinct disciplines, computational science and genome biology. I am
particularly grateful to professor Sakari Knuutila for his enthusiasm,
curiosity, and personal example in collaboration and daily research
work. Researchers in the Laboratory of Cytomolecular Genetics at the
Haartman Institute have provided a friendly and inspiring environment
for active collaboration during the last years.

My sincere compliments belong to all of my other co-authors, in
particular to Tero Aittokallio, Laura Elo-Uhlgren, Jaakko Hollmén,
Juha Knuuttila, Samuel Myllykangas and Janne Nikkilä. It has been a
pleasure to work with you, and your contributions extend beyond what
we wrote together. I would also like to thank the former and present
members of the MI research group for working beside me through these
years, as well as for intriguing discussions about science and life in
general. I would also like to thank the personnel of the ICS
department, in particular professors Erkki Oja and Olli Simula, who
have helped to provide an excellent academic research environment, as well as
our secretaries Tarja Pihamaa and Leila Koivisto, and Markku Ranta and
Miki Sirola, who have given valuable help in so many practical matters
during the years.

Science is a community effort. Open sharing of ideas, knowledge,
publication material, data, software, code, experiences and emotions
has had a tremendous impact to this thesis. I will express my sincere
gratitude to the community by continued participation and
contributions. 

I would also like to thank my earliest scientific advisors; Reijo, who
brought me writings about the chemistry of life and helped me to grow
bacteria and prepare space dust in the 1980's, Pekka, who has
demonstrated the power of criticism and emphasized that natural
science has to be exact, Tapio, for the attitude that maths can be
just fun, and Risto, for showing how rational thinking can be applied
also in real life. Thanks also go to my science friends, Manu and
Ville; we have shared the passion for natural science, and I want to
thank you for our continuous and inspiring discussions along the way.
I am grateful to my grandfather Osmo, who shared with me the wonder
towards life, science, and humanities, and was willing to discuss it
all through days and nights when I was a child, questioning himself
the self-evident truths again and again, remaining as puzzled as I
was. And for Alli and Arja, my grandmothers, for their understanding,
and all support and love.

My Friends. With you I have explored other facets of nature, science,
and life... Thank you for staying with me through all these years and
sharing so many aspects of curiosity, exploration and mutual
understanding.


Finally, I am grateful to my parents and sister, Pipsa, Kari, and
Tuuli. You have accepted me and loved me, supported me on the paths
that I have chosen to follow, and understood that freedom can create
the strongest ties.\\[2mm]

\noindent Cambridge, November 23, 2010\\
\noindent Leo Lahti
\newpage

\section*{LIST OF PUBLICATIONS}
\addcontentsline{toc}{section}{List of publications}
\markboth{LIST OF PUBLICATIONS}{LIST OF PUBLICATIONS}
\label{sec:publications}

This thesis consists of an overview and of the following publications
which are referred to in the text by their Roman numerals.

\leftmargini=6mm {\sloppy
\begin{enumerate}
\item Laura L. Elo, Leo Lahti, Heli Skottman, Minna Kyl{\"a}niemi,
  Riitta Lahesmaa, and Tero Aittokallio.  \newblock Integrating
  probe-level expression changes across generations of Affymetrix
  arrays.  \newblock {\em Nucleic Acids Research}, 33(22):e193, 2005.
\label{PECA}

\item Leo Lahti, Laura L. Elo, Tero Aittokallio, and Samuel Kaski.
  \newblock Probabilistic analysis of probe reliability in
  differential gene expression studies with short oligonucleotide
  arrays. \newblock {\em IEEE/ACM Transactions on Computational Biology
    and Bioinformatics}, 8(1):217--225, 2011.
\label{RPA}

\item Leo Lahti, Juha E.A. Knuuttila, and Samuel Kaski.
\newblock Global modeling of transcriptional responses in interaction networks.
\newblock {\it Bioinformatics}, 26(21):2713--2720, 2010.
\label{NR}

\item Leo Lahti, Samuel Myllykangas, Sakari Knuutila, and Samuel
  Kaski.  \newblock Dependency detection with similarity constraints.
  \newblock In T{\"u}lay Adali, Jocelyn Chanussot, Christian Jutten,
  and Jan Larsen, editors, {\em Proceedings of the 2009 IEEE
    International Workshop on Machine Learning for Signal Processing
    XIX}, pages 89--94. IEEE, Piscataway, NJ, 2009.  
\label{MLSP}

\item Janne Sinkkonen, Janne Nikkil\"{a}, Leo Lahti, and Samuel Kaski.
  \newblock Associative clustering.  \newblock In Boulicaut, Esposito,
  Giannotti, and Pedreschi (editors), {\em Machine Learning: ECML2004
    (Proceedings of the ECML'04, 15th European Conference on Machine
    Learning)}, Lecture Notes in Computer Science 3201,
  396--406. Springer, Berlin, 2004.  
\label{ECML}

\item
Samuel Kaski, Janne Nikkil{\"a}, Janne Sinkkonen, Leo Lahti, Juha E.A. Knuuttila, and Cristophe Roos.
\newblock Associative clustering for exploring dependencies between functional
  genomics data sets.
\newblock {\em IEEE/ACM Transactions on Computational Biology and
  Bioinformatics: Special Issue on Machine Learning for Bioinformatics -- Part 2}, 2(3):203--216, 2005.
\label{AC}

\end{enumerate}}

\newpage


\section*{SUMMARY OF PUBLICATIONS AND THE\\ AUTHOR'S CONTRIBUTION}
\addcontentsline{toc}{section}{Summary of publications and the
  author's contribution} \markboth{THE AUTHOR'S CONTRIBUTION}{THE
  AUTHOR'S CONTRIBUTION}

The publications in this thesis have been a joint effort of all
authors; key contributions by the author of this thesis are summarized
below.

Publication~\ref{PECA} introduces a novel analysis strategy to improve
the accuracy and reproducibility of the measurements in genome-wide
transcriptional profiling studies. A central part of the approach is
the utilization of side information in external genome sequence
databases.  The author participated in the design of the study,
suggested the utilization of external sequence data, implemented this,
as well as participated in preparing the manuscript.

Publication \ref{RPA} provides a probabilistic framework for
probe-level gene expression analysis. The model combines statistical
power across multiple microarray experiments, and is shown to
outperform widely-used preprocessing methods in differential gene
expression analysis. The model provides tools to assess probe
performance, which can potentially help to improve probe and
microarray design. The author had a major role in designing the
study. The author derived the formulation, implemented the model,
performed the probe-level experiments, as well as coordinated the
manuscript preparation. The author prepared an accompanied open source
implementation which has been published in BioConductor, a reviewed
open source repository for computational biology algorithms.

Publication~\ref{NR} introduces a novel approach for organism-wide
modeling of transcriptional activity in genome-wide interaction
networks. The method provides tools to analyze large collections of
genome-wide transcriptional profiling data. The author had a major
role in designing the study. The author implemented the algorithm,
performed the experiments, as well as coordinated the manuscript
preparation. The author participated in and supervised the preparation
of an accompanied open source implementation in BioConductor.

Publication~\ref{MLSP} introduces a regularized dependency modeling
framework with particular applications in cancer genomics. The author
had a major role in formulating the biomedical modeling task, and in
designing the study. The theoretical model was jointly developed by
the author and S. Kaski. The author derived and implemented the model,
carried out the experiments, and coordinated the manuscript
preparation. The author supervised and participated in the preparation
of an accompanied open source implementation in BioConductor.

Publication \ref{ECML} introduces the associative clustering
principle, which is a novel data integration framework for dependency
detection with direct applications in functional genomics. The author
participated in implementation of the method, had the main
responsibility in designing and performing the functional genomics
experiments, as well as participated in preparing the manuscript.

Publication \ref{AC} contains the most extensive treatment of the
associative clustering principle. In addition to presenting detailed
theoretical considerations, this work introduces new sensitivity
analysis of the results, and provides a comprehensive validation in
bioinformatics case studies. The author participated in designing the
experiments, performed the comparative functional genomics experiments
and technical validation, as well as participated in preparing the
manuscript.
\newpage

\section*{LIST OF ABBREVIATIONS AND SYMBOLS}
\addcontentsline{toc}{section}{List of abbreviations and symbols}

\noindent
In this thesis boldface symbols are used to denote matrices and
vectors. Capital symbols (\(\X\)) signify matrices and lowercase
symbols ($\x$) column vectors. {\it Normal lowercase} symbols indicate
scalar variables.

\vspace{0.5cm}

\noindent
$\Real$ \hfill Real domain\\
$\X, \Y$ \hfill Data matrices (\(D \times N\)) \\
\([\X; \Y]\) \hfill Concatenated data \\ 
$\x$, $\y$ \hfill Data samples, vectors in $\Real^D$ \\
$x$, $y$ \hfill Scalars in $\Real$ \\
$\Xcal, \Ycal$ \hfill Random variables \\
$\I$ \hfill Identity matrix\\
$\bSigma, \Psib$ \hfill Covariance matrices\\
$p(\x)$ \hfill Probability or probability density of $\mathcal{X}$ \\
\(p(\X)\) \hfill Likelihood\\
$\E[\cdot]$ \hfill Expectation\\
$\|\cdot\|$ \hfill Norm of a matrix or vector \\
$Tr$ \hfill Matrix trace \\
$I(\mathcal{X};\mathcal{Y})$ \hfill Mutual information between random variables $\mathcal{X}$ and $\mathcal{Y}$ \\
$\text{Beta}(\alpha, \beta)$ \hfill Beta distribution with parameters $\alpha$ and $\beta$ \\
$\text{Dir}(\bth)$ \hfill Dirichlet distribution with parameter vector $\bth$ \\
$\text{IG}(\alpha,\beta)$ \hfill Inverse Gamma distribution with parameters $\alpha$ and $\beta$ \\
$\text{Mult}(N,\bth)$ \hfill Multinomial distribution with sample size $N$ and parameter vector $\bth$ \\
$\Nd(\bmu,\bSigma)$ \hfill Normal distribution with mean $\bmu$ and covariance matrix $\bSigma$ \\
AC			 \hfill Associative clustering \\
aCGH                     \hfill Array Comparative genomics hybridization \\
CCA                      \hfill Canonical correlation analysis \\
cDNA                     \hfill Complementary DNA\\
DNA                      \hfill Deoxyribonucleic acid \\
DP                       \hfill Dirichlet process \\
EM                       \hfill Expectation -- Maximization algorithm \\
IB                       \hfill Information bottleneck \\
KL--divergence		 \hfill Kullback-Leibler divergence \\
MAP                      \hfill Maximum a posteriori \\
MCMC                     \hfill Markov chain Monte Carlo\\
ML                       \hfill Maximum likelihood \\
mRNA                     \hfill Messenger-RNA\\
tRNA                     \hfill Transfer-RNA\\
PCA                      \hfill Principal component analysis \\
RNA                      \hfill Ribonucleic acid\\

\newpage


\pagenumbering{arabic}
\setcounter{page}{1}

\chapter{Introduction}
\label{chap:intro}

Revolutions in measurement technologies have led to revolutions in
science and society. Introduction of the microscope in the 17th
century opened a new view to the world of living organisms and enabled
the study of life processes at cellular level. Since then, new
techniques have been developed to investigate ever smaller objects.
The discovery of the molecular structure of the DNA in 1953
\citep{Watson53nature} led to the establishment of genes as
fundamental units of genetic information that is passed on between
generations. The draft sequence of the human genome, covering three
billion DNA base pairs, was published in 2001 \citep{Lander01,
  Venter01}. Modern measurement technologies provide researchers with
large volumes of data concerning the structure, function, and
interactions of genes and their products.  Rapid accumulation of
genomic data in shared community databases has accelerated biological
research \citep{Cochrane2010}, but the structural and functional
organization of genetic information is still poorly understood.  While
functional roles of individual genes have been characterized, little
is known regarding the higher-level regularities and interactions from
which the complexity and diversity of life emerges. The quest for
systems-level understanding of genome function is a major paradigm in
modern biology \citep{Collins03}.

Computational science has a key role in transforming the genomic data
collections into new biological knowledge \citep{Cohen04}. New
observations allow the formulation of new research questions, but also
bring new challenges \citep{Barbour05}.  The sheer size of
high-throughput data sets makes them incomprehensible for human mind,
and the complexity of biological phenomena and high levels of
uncontrolled variation set specific challenges for computational
analysis \citep{Tilstone03, Troyanskaya05}. Filtering relevant
information from statistically uncertain high-dimensional data is a
challenging task where new computational methods are needed to
organize and summarize the overwhelming volumes of observational data
into a comprehensible form to make new discoveries about the structure
of life; computation is a new microscope for studying massive data
sets.

This thesis develops principled exploratory methods to investigate the
{\it human transcriptome}. It is a central functional layer of the
genome and a significant source of phenotypic variation.  The
transcriptome refers to the complete collection of messenger-RNA
transcripts of an organism. The essentially static genome sequence
regulates the time- and context-specific patterns of transcriptional
activity of the genes, and subsequently the function of living cells
through protein synthesis. An average cell contains over 300,000 mRNA
molecules and the expression levels of individual genes span 4-5
orders of magnitude \citep{Carninci2009}.  A wealth of associated
genomic information resources are available in public repositories
\citep{Cochrane2010}. By combining heterogeneous information sources
and utilizing the wealth of background information in public
repositories, it is possible to solve some of the problems that are
related to the statistical uncertainties and small sample size of
individual data sets, as well as to form a holistic picture of the
genome \citep{Huttenhower2010}.

The observational data can provide the starting point to discover
novel research hypotheses of poorly characterized large-scale systems;
the analysis proceeds from general observations of the data toward
more detailed investigations and hypotheses.  This differs from
traditional hypothesis testing where the investigation proceeds from
hypotheses to measurements that target particular research questions,
in order to support or reject a given hypothesis. \emph{Exploratory
  data analysis} refers to the use of computational tools to summarize
and visualize the data in order to identify potentially interesting
structure, and to facilitate the generation of new research hypotheses
when the search space would be otherwise exhaustively large
\citep{Tukey77}.  When the system is poorly characterized, there is a
need for methods that can adapt to the data and extract features in an
automated way. This is useful since application-oriented models often
require careful preprocessing of the data and a timely model fitting
process. They may also require prior knowledge of the investigated
system, which is often not available. \emph{Statistical learning}
investigates solutions to these problems.

\section{Contributions and organization of the thesis}

This thesis introduces computational strategies for genome- and
organism-wide analysis of the human transcriptome. The thesis provides
novel tools (i) to increase the reliability of high-throughput
microarray measurements by combining statistical evidence from genome
sequence databases and across multiple microarray experiments, (ii) to
model context-specific transcriptional activation patterns of
genome-scale interaction networks across normal human body by using
background information of genetic interactions to guide the analysis,
and (iii) to integrate measurements of the human transcriptome to
other layers of genomic information with novel dependency modeling
techniques for co-occurring data sources. The three strategies address
widely recognized challenges in functional genomics
\citep{Collins03,Troyanskaya05}. 

Obtaining reliable measurements is the crucial starting point for any
data analysis task. The first contribution of this thesis is to
develop computational strategies that utilize side information in
genomic sequence and microarray data collections in order to reduce
noise and improve the quality of high-throughput
observations. Publication~\ref{PECA} introduces a probe-level strategy
for microarray preprocessing, where updated genomic sequence databases
are used in order to remove erroneously targeted probes to reduce
measurement noise. The work is extended in Publication~\ref{RPA},
which introduces a principled probabilistic framework for probe-level
analysis. A generative model for probe-level observations combines
evidence across multiple experiments, and allows the estimation of
probe performance directly from microarray measurements. The model
detects a large number of unreliable probes contaminated by known
probe-level error sources, as well as many poorly performing probes
where the source of contamination is unknown and could not be
controlled based on existing probe-level information. The model
provides a principled framework to incorporate prior information of
probe performance. The introduced algorithms outperform widely used
alternatives in differential gene expression studies.

A novel strategy for organism-wide analysis of transcriptional
activity in genome-scale interaction networks in Publication~\ref{NR}
forms the second main contribution of this thesis. The method searches
for local regions in a network exhibiting coordinated transcriptional
response in a subset of conditions.  Constraints derived from genomic
interaction databases are used to focus the modeling on those parts of
the data that are supported by known or potential interactions between
the genes. Nonparametric inference is used to detect a number of
physiologically coherent and reproducible transcriptional responses,
as well as context-specific regulation of the genes. The findings
provide a global view on transcriptional activity in cell-biological
networks and functional relatedness between tissues.

The third contribution of the thesis is to integrate measurements of
the human transcriptome to other layers of genomic information.  Novel
dependency modeling techniques for co-occurrence data are used to
reveal regularities and interactions, which could not be detected in
individual observations. The regularized dependency modeling framework
of Publication~\ref{MLSP} is used to detect associations between
chromosomal mutations and transcriptional activity. Prior biological
knowledge is used to constrain the latent variable model and shown to
improve cancer gene detection performance. The associative clustering,
introduced in Publications~\ref{ECML} and~\ref{AC}, provides tools to
investigate evolutionary divergence of transcriptional activity.

Open source implementations of the key methodological contributions of
this thesis have been released in order to guarantee wide access to
the developed algorithmic tools and to comply with the emerging
standards of transparency and reproducibility in computational
science, where an increasing proportion of research details are
embedded in code and data accompanying traditional publications
\citep{Boulesteix2010, Carey2010, Ioannidis09} and transparent sharing
of these resources can form valuable contributions to public knowledge
\citep{Sommer2010, Sonnenburg2007, Stodden2010}. 

The thesis is organized as follows: In Chapter~\ref{ch:bio}, there is
an overview of functional genomics, related measurement techniques,
and genomic data resources. General methodological background, in
particular of exploratory data analysis and the probabilistic modeling
paradigm, is provided in Chapter~\ref{ch:meth}. The methodological
contributions of the thesis are presented in
Chapters~\ref{ch:preproc}-\ref{ch:integration}.  In
Chapter~\ref{ch:preproc}, strategies to improve the reliability of
high-throughput microarray measurements are presented. In
Chapter~\ref{ch:atlases} methods for organism-wide analysis of the
transcriptome are considered. In Chapter~\ref{ch:integration}, two
general-purpose algorithms for dependency modeling are introduced and
applied in investigating functional effects of chromosomal mutations
and evolutionary divergence of transcriptional activity. The
conclusions of the thesis are summarized in
Chapter~\ref{ch:conclusion}.


\newpage

\chapter{Functional genomics}\label{ch:bio}

\begin{quotation}
\emph{From all we have learnt about the structure of living matter, we
must be prepared to find it working in a manner that cannot be reduced
to the ordinary laws of physics - - because the construction is
different from anything we have yet tested in the physical
laboratory.}
\begin{flushright}
E. Schr{\"o}dinger (1956)
\end{flushright}
\end{quotation}

Living organisms are controlled not only by natural laws but also by
inheritable {\it genetic programs} \citep{Mayr04,Schrodinger44}. Such
{\it double causation} is a unique feature of life, and in fundamental
contrast to purely physical processes of the inanimate world. Life may
have emerged on earth more than 3.4 billion years ago
\citep{Schopf2006, Tice2004}. Genetic information evolves by means of
{\it natural selection} \citep{Darwin1859}. Living organisms maintain
homeostasis, adapt to changing environments, respond to external
stimuli, and communicate.  Peculiar features of living systems include
metabolism, growth and hierarchical organization, as well as the
ability to replicate and reproduce. All known life forms share
fundamental mechanisms at molecular level, which suggests a common
evolutionary origin of the living organisms.

The complete collection of genetic material, {\it the genome}, encodes
the heritable genetic program of an organism. Advances in measurement
technology and computational science have opened up new views to the
large-scale organization of the genome
\citep{Carroll03,Lander96}. {\it Functional genomics} is a
subdiscipline of molecular biology investigating the functional
organization and properties of genetic information. In this thesis,
new computational approaches are developed for investigation of a
central functional layer of the genome of our own species, the human
{\it transcriptome}. This chapter gives an overview to the relevant
concepts in genome biology in eukaryotic organisms and associated
genomic data resources. For further background in molecular genome
biology, see \cite{Alberts02, Brown06}.

\section{Universal genetic code}

Cells are fundamental building blocks of living organisms. All known
life forms maintain a carbon-based cellular form that carries the
genetic program \citep{Alberts02}. Each cell carries a copy of the
heritable genetic code, {\it the genome}. The human genome is divided
in 23 pairs of {\it chromosomes}, located in the nucleus of the cell,
as well as in additional mitochondrial genome. Chromosomes are
macroscopic deoxyribonucleic acid (DNA) molecules in which the DNA is
wrapped around {\it histone} molecules and packed into a peculiar {\it
  chromatin} structure that will ultimately constitute
chromosomes. The {\it genetic code} in the DNA consists of four {\it
  nucleotides}: adenosine (A), thymine (T), guanine (G), and cytosine
(C). In ribonucleic acid (RNA), the thymine is replaced by uracil
(U). Ordering of the nucleotides carries genetic information. Nucleic
acid sequences have a peculiar base pairing property, where only A-T/U
and G-C pairs can hybridize with each other. This leads to the
well-known double-stranded structure of the DNA, and forms the basis
for cellular information processing.  The \emph{central dogma of
  molecular biology} \citep{Crick70} states that DNA encodes the
information to construct proteins through the irreversible process of
{\it protein synthesis}. This is a central paradigm in molecular
biology, describing the functional organization of life at the
cellular level.

\subsection{Protein synthesis}

Genes are basic units of genetic information.  The gene is a sequence
of DNA that contains the information to manufacture a protein or a set
of related proteins. Genetic variation and regulation of gene activity
has therefore major phenotypic consequences. The {\it regulatory
  region} and {\it coding sequence} are two key elements of a
gene. The regulatory region regulates gene activity, while the coding
sequence carries the instructions for protein synthesis
\citep{Alberts02}. Interestingly, the concept of a gene remains
controversial despite comprehensive identification of the
protein-coding genes in the human genome and detailed knowledge of
their structure and function \citep{Pearson2006}.

\begin{figure}[ht]\label{fig:bio}
\centerline{\includegraphics[width=.9\textwidth]{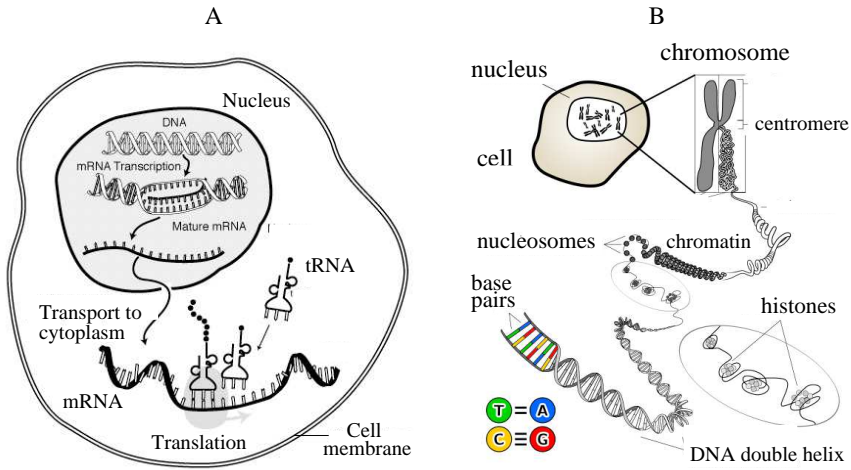}}
\caption{{\bf A} Key steps of protein synthesis. The two key processes
  in protein synthesis are called {\it transcription} and {\it
    translation}, respectively. In transcription, the DNA sequence of
  the gene is transcribed into pre-mRNA based on the base pairing
  property of nucleic acid sequences. The pre-mRNA is modified to
  produce mature messenger-RNA (mRNA), which is then transported to
  cytoplasm. Transfer-RNA (tRNA) carries the mRNA to ribosomes, where
  it is translated into an amino acid sequence based on the universal
  genetic code where each nucleotide triplet of the mRNA sequence,
  so-called \emph{codon}, corresponds to a particular amino acid. The
  amino acid sequence is subsequently modified to form the final
  functional protein product. {\bf B} Organization of the genetic
  material in an eukaryotic cell. The nucleotide base pairs form the
  double helix structure of DNA. This is wrapped around histone
  molecules to form nucleosomes, and the chromatin sequence. The
  chromatin is tightly packed to form chromosomes that carry the
  genetic material and are located in the cell nucleus. The image has
  been modified from
  http://commons.wikimedia.org/wiki/File:Chromosome\_en.svg.}
\end{figure}

Proteins, encoded by the genes, are key functional entities in the
cell. They form cellular structures, and participate in cell signaling
and functional regulation. {\it Protein synthesis} refers to the
cell-biological process that converts genetic information into final
functional protein products (Figure~\ref{fig:bio}A). Key steps in
protein synthesis include transcription, pre-mRNA splicing, and
translation. In {\it transcription}, the double-stranded DNA is opened
in a proximity of the gene sequence and the process is initiated on
the regulatory region of the gene. The DNA sequence of the gene is
then converted into a complementary pre-mRNA by a polymerase
enzyme. The pre-mRNA sequence contains both protein coding and
non-coding segments. These are called {\it exons} and {\it introns},
respectively. In {\it pre-mRNA splicing}, the introns are removed and
the exons are joined together to form mature {\it messenger-RNA
  (mRNA)}. A gene can encode multiple splice variants, corresponding
to different exon definitions and their combinations; this is called
{\it alternative splicing}. The mature mRNA is exported from nucleus
to the cell cytoplasm. In {\it translation} the mRNA is converted into
a corresponding amino acid sequence in ribosomes based on the {\it
  universal genetic code} that defines a mapping between nucleic acid
triplets, so-called {\it codons}, and amino acids. The code is common
for all known life forms. Each consecutive codon on the mRNA sequence
corresponds to an amino acid, and the corresponding sequence of amino
acids constitutes a protein. In the final stage of protein synthesis,
the amino acid sequence folds into a three-dimensional structure and
undergoes {\it post-translational modifications}. The structural
characteristics of a protein molecule will ultimately determine its
functional properties \citep{Alberts02}.

\subsection{Layers of regulation}

Phenotypic changes can rarely be attributed to changes in individual
genes; cell function is ultimately determined by coordinated
activation of genes and other biomolecular entities in response to
changes in cell-biological environment \citep{Hartwell99}. Gene
activity is regulated at all levels of protein synthesis and cellular
processes. A major portion of functional genome sequence and protein
coding-genes themselves participate in the regulatory system itself
\citep{Lauffenburger00}.

{\it Epigenetic regulation} refers to chemical and structural
modifications of chromosomal DNA, the {\it chromatin}, for instance
through methylation, acetylation, and other histone-binding molecules.
Such modifications affect the packing of the DNA molecule around {\it
  histones} in the cell nucleus. The combinatorial regulation of such
modifications regulates access to the gene sequences
\citep{Gibney2010}. Epigenetic changes are believed to be heritable
and they constitute a major source of variation at individual and
population level \citep{Johnson2010}. {\it Transcriptional regulation}
is the next major regulatory layer in protein synthesis. So-called
{\it transcription factor} proteins can regulate the transcription
rate by binding to control elements in gene regulatory region in a
combinatorial fashion. {\it Post-transcriptional modifications} will
then regulate pre-mRNA splicing. Up to 95\% of human multi-exon genes
are estimated to have alternative splice variants
\citep{Pan2008}. Consequently, a variety of related proteins can be
encoded by a single gene. This contributes to the structural and
functional diversity of cell function \citep{Stetefeld2005}. Several
mechanisms will then affect {\it mRNA degradation} rates. For
instance, micro-RNAs that are small, 21-25 basepair nucleotide
sequences can inactivate specific mRNA transcripts through
complementary base pairing, leading to mRNA degradation, or prevention
of translation. Finally, {\it post-translational modifications}, {\it
  protein degradation}, and other mechanisms will affect the
three-dimensional structure and life cycle of a protein. The proteins
will participate in further cell-biological processes. The processes
are in continuous interaction and form complex functional networks,
which regulate the life processes of an organism \citep{Alberts02}.

\section{Organization of genetic information}

The understanding of the structure and functional organization of the
genome is rapidly accumulating with the developing genome-scanning
technologies and computational methods. This section provides an
overview to key structural and functional layers of the human genome.

\subsection{Genome structure}

The genome is a dynamic structure, organized and regulated at multiple
levels of resolution from individual nucleotide base pairs to complete
chromosomes (Figure~\ref{fig:bio}B; \cite{Brown06}). A major portion
of heritable variation between individuals has been attributed to
differences in the genomic DNA sequence. Traditionally, main genetic
variation was believed to arise from small point mutations, so-called
{\it single-nucleotide polymorphisms (SNPs)}, in protein-coding
DNA. Recently, it has been increasingly recognized that {\it
  structural variation} of the genome makes a remarkable contribution to
genetic variation. Structural variation is observed at all levels of
organization from single-nucleotide polymorphisms to large chromosomal
rearrangements, including deletions, insertions, duplications,
copy-number variants, inversions and translocations of genomic regions
\citep{Feuk06, Sharp06}. Such modifications can directly and
indirectly influence transcriptional activity and contribute to human
diversity and health \citep{Collins03, Hurles08}.

The draft DNA sequence of the complete human genome was published in
2001 \citep{Lander01, Venter01}. The human genome contains three
billion base pairs and approximately 20,000-25,000 protein-coding
genes \citep{Collins04}. The protein-coding exons comprise less than
1.5\% of the human genome sequence. Approximately 5\% of the human
genome sequence has been conserved in evolution for more than 200
million years, including the majority of protein-coding genes
\citep{encode07, Waterston02}. Half of the genome consists of highly
repetitive sequences. The genome sequence contains structural elements
such as centromeres and telomeres, repetitive and mobile elements,
\citep{Prak00}, retroelements \citep{Bannert04}, and non-coding,
non-repetitive DNA \citep{Collins03}. The functional role of
intergenic DNA, which forms 75\% of the genome, is to a large extent
unknown \citep{Venter01}. Recent evidence suggests that the
three-dimensional organization of the chromosomes, which is to a large
extent regulated by the intergenic DNA is under active selection, can
have a remarkable regulatory role \citep{Lieberman-Aiden2009,
  Parker2009}. Comparison of the human genome with other organisms,
such as the mouse \citep{Waterston02} can highlight important
evolutionary differences between species. For a comprehensive review
of the structural properties of the human genome, see \cite{Brown06}.

\subsection{Genome function}

In protein synthesis, the gene sequence is transcribed into pre-mRNA,
which is then further modified into mature messenger-RNA and
transported to cytoplasm. An average cell contains over 300,000 mRNA
molecules, and the mRNA concentration, or {\it expression levels} of
individual genes, vary according to Zipf's law, a power-law
distribution where most genes are expressed at low concentrations,
perhaps only one or few copies of the mRNA per cell on average, and a
small number of genes are highly expressed, potentially with thousands
of copies per cell \citep[see][]{Carninci2009, Furusawa2003}.
Cell-biological processes are reflected at the transcriptional
level. Transcriptional activity varies by cell type, environmental
conditions and time. Different collections of genes are active in
different contexts. {\it Gene expression}, or mRNA expression, refers
to the expression level of an mRNA transcript at particular
physiological condition and time point. In addition to protein-coding
mRNA molecules that are the main target of analysis in this thesis,
the cell contains a variety of other functional and non-functional
mRNA transcripts, for instance micro-RNAs, ribosomal RNA and
transfer-RNA molecules \citep{Carninci2009, Johnson05}.

The {\it transcriptome} refers to the complete collection of mRNA
sequences of an organism. This is a central functional layer of the
genome that regulates protein production in the cells, with a
significant role in creating genetic variation
\citep{Jordan05}. According to current estimates, up to 90\% of the
eukaryotic genome can be transcribed \citep{Carninci05, Gagneur2009}.
The protein-coding mRNA transcripts are translated into proteins at
ribosomes during protein synthesis. 

{\it The proteome} refers to the collection of protein products of an
organism. The proteome is a main functional layer of the genome. Since
the final protein products carry out a main portion of the actual cell
functions, techniques for monitoring the concentrations of all
proteins and their modified forms in a cell simultaneously would
significantly help to improve the understanding of the cellular
systems \citep{Collins03}. However, sensitive, reliable and
cost-efficient genome-wide screening techniques for measuring protein
expression are currently not available. Therefore genome-wide
measurements of the mRNA expression levels are often used as an
indirect estimate of protein activity.

In addition to the DNA, RNA and proteins, the cell contains a variety
of other small molecules. The extreme functional diversity of living
organisms emerges from the complex network of interactions between the
biomolecular entities \citep{Barabasi04, Hartwell99}. Understanding of
these networks and their functional properties is crucial in
understanding cell function \citep{Collins03, Schadt2009}. However,
the systemic properties of {\it the interactome} are poorly
characterized and understood due to the complexity of biological
phenomena and incomplete information concerning the interactions.  The
cell-biological processes are inherently modular \citep{Hartwell99,
  Ihmels02, Lauffenburger00}, and they exhibit complex {\it pathway
  cross-talk} between the cell-biological processes \citep{Li08c}. In
modular systems, small changes can have significant regulatory effects
\citep{Espinosa-Soto2010}.

\section{Genomic data resources}

Systematic observations from the various functional and regulatory
layers of the genome are needed to understand cell-biological
systems. Efficient sharing and integration of genomic information
resources through digital media has enabled large-scale investigations
that no single institution could afford. The public human genome
sequencing project \citep{Lander01} is a prime example of such
project. Results from genome-wide transcriptional profiling studies
are routinely deposited to public repositories \citep{Barrett2009,
  Parkinson2009}. Sharing of original data is increasingly accepted as
the scientific norm, often following explicit data release
policies. The establishment of large-scale databases and standards for
representing biological information support the efficient use of these
resources \citep{Bammler05, Brazma06}. A continuously increasing array
of genomic information is available in these databases, concerning
aspects of genomic variability across individuals, disease states, and
species \citep{Brent2008, Church05, Cochrane2010,
  G10KCOSconsortium2009, tcga08}.

\subsection{Community databases and evolving biological knowledge}
\label{sec:biodata}

\subsubsection{Genomic sequence databases}

During the human genome project and preceding sequencing projects DNA
sequence reads were among the first sources of biological data that
were collected in large-scale public repositories, such as GenBank
\citep{Benson2010}. GenBank contains comprehensive sequence
information of genomic DNA and RNA for a number of organisms, as well
as a variety of information concerning the genes, non-coding regions,
disease associations, variation and other genomic features. Online
analysis tools, such as the Ensembl Genome browser \citep{Flicek2010},
facilitate efficient use of these annotation
resources. Next-generation sequencing technologies provide rapidly
increasing sequencing capacity to investigate sequence variation
between individuals, populations and disease states
\citep{Ledford2010, McPherson2009}. In particular, the human and mouse
transcriptome sequence collections at the Entrez Nucleotide database
of GenBank are utilized in this thesis, in Publications~\ref{PECA}
and~\ref{RPA}.

\subsubsection{Transcriptome databases}

Gene expression measurement provides a snapshot of mRNA transcript
levels in a cell population at a specific time and condition,
reflecting the activation patterns of the various cell-biological
processes. While gene expression measurements provide only an indirect
view to cellular processes, their wide availability provides a unique
resource for investigating gene co-regulation on a genome- and
organism-wide scale. Versatile collections of microarray data in
public repositories, such as the Gene Expression Omnibus (GEO;
\cite{Barrett2009}) and ArrayExpress \citep{Parkinson2009} are
available for human and model organisms, and they contain valuable
information of cell function \citep{Carninci05, DeRisi97, Russ10,
  Zhang04b}.

Several techniques are available for quantitative and highly parallel
measurements of mRNA or {\it gene expression}, allowing the
measurement of the expression levels of tens of thousands of mRNA
transcripts simultaneously \citep{Bradford2010}. Microarray techniques
are routinely used to measure the expression levels of tens of
thousands of mRNA transcripts in a given sample, and transcriptional
profiling is currently a main high-throughput technique used to
investigate gene function at genome- and organism-wide scale
\citep{Gershon05, Yauk04}. Increasing amounts of transcriptional
profiling data are being produced by sequencing-based methods
\citep{Carninci2009}. A main difference between the microarray- and
sequencing-based techniques is that gene expression arrays have been
designed to measure predefined mRNA transcripts, whereas
sequencing-based methods do not require prior information of the
measured sequences, and enable {\it de novo} discovery of expressed
transcripts \citep{Bradford2010, Hoen08}. Large-scale microarray
repositories provide currently the most mature tools for data
processing and retrieval, and form the main source of transcriptome
data in this thesis.

Microarray technology is based on the base pairing property of nucleic
acid sequences where the DNA or RNA sequences in a sample bind to the
complementary nucleotide sequences on the array. This is called {\it
  hybridization}.  The measurement process begins by the collection of
cell samples and isolation of the sample mRNA. The isolated mRNA is
converted to cDNA, {\it labeled} with specific marker molecules, and
hybridized on complementary probe sequences on the array. The array
surface may contain hundreds of thousands of spots, each containing
specific probe sequences designed to uniquely match with particular
mRNA sequences. The hybridization level reflects the target mRNA
concentration in the sample, and it is estimated by measuring the
intensity of light emitted by the label molecules with a laser
scanner. {\it Short oligonucleotide arrays} \citep{Lockhart96} are
among the most widely used microarray technologies, and they are the
main source of mRNA expression data in this thesis.  Short
oligonucleotide arrays utilize multiple, typically 10-20, probes for
each transcript target that bind to different regions of the same
transcript sequence. Use of several 25-nucleotide probes for each
target leads to more robust estimates of transcript activity. Each
probe is expected to uniquely hybridize with its intended target, and
the detected hybridization level is used as a measure of the activity
of the transcript. A short oligonucleotide array measures absolute
expression levels of the mRNA sequences; relative differences between
conditions can be investigated afterwards by comparing these
measurements. A standard whole-genome array measures typically
\(\sim\)20,000-50,000 unique transcript sequences. A single microarray
experiment can therefore produce hundreds of thousands of raw
observations.

Comparison and integration of individual microarray experiments is
often challenging due to remarkable experimental variation between the
experiments. Common standards have been developed to advance the
comparison and integration \citep{Brazma01, Brazma06}. Carefully
controlled integrative datasets, so-called {\it gene expression
  atlases}, contain thousands of genome-wide measurements of
transcriptional activity across diverse conditions in a directly
comparable format. Examples of such data collections include
GeneSapiens \citep{Kilpinen08}, the human gene expression atlas of the
European Bioinformatics Institute \citep{Lukk10}, as well as the
NCI-60 cell line panel \citep{Scherf2000}. Integrative analysis of
large and versatile transcriptome collections can provide a holistic
view of transcriptional activity of the various cell-biological
processes, and opens up possibilities to discover previously
uncharacterized cellular mechanisms that contribute to human health
and disease.

\subsubsection{Other types of microarray data}

Microarray techniques can also be used to study other functional
aspects of the genome, including epigenetics and micro-RNA regulation,
chromosomal aberrations and polymorphisms, alternative splicing, as
well as transcription factor binding \citep{Butte02a, Hoheisel06}.
For instance, chromosomal aberrations can be measured with the {\it
  array comparative genome hybridization method (aCGH;}
\citealt{Pinkel2005}), which is based on hybridization of DNA sequences
on the array surface. Copy number changes are a particular type of
chromosomal aberrations, which are a major mechanism for cancer
development and progression. Copy number alterations can cause changes
in gene- and micro-RNA expression, and ultimately cell-biological
processes \citep{Beroukhim10}.  A public repository of copy number
measurement data is provided for instance by the CanGEM database
\citep{Scheinin08}. In Publication~\ref{MLSP}, microarray measurements
of DNA copy number changes are integrated with transcriptional
profiling data to discover potential cancer genes for further
biomedical analysis.

\subsubsection{Pathway and interaction databases}

Curated information concerning cell-biological processes is valuable
in both experimental design and validation of computational studies
\citep{Blake04}. Representation of dynamic biochemical reactions in
their full richness is a challenging task beyond a mere listing of
biochemical events; a variety of proteins and other compounds interact
in a hierarchical manner through various molecular mechanisms
\citep{Hartwell99, Przytycka2010}. Standardized database formats such
as the BioPAX \citep{BioPAX05} and SBML \citep{Stromback05} advance
the accumulation of highly structured biological knowledge and
automated analysis of such data. A huge body of information concerning
cell-biological processes is available in public repositories. The
most widely used annotation resources include the Gene Ontology (GO)
database \citep{Ashburner00} and the KEGG pathway database
\citep{Kanehisa2010}. The GO database provides functional annotations
for genes and can be used for instance to detect enrichment of certain
functional categories among the key findings from computational
analysis, as in Publication~\ref{AC}, where enrichment analysis is
used for both validation and interpretation purposes.  Pathways are
more structured representations concerning cellular processes and
interactions between molecular entities. Such prior information can be
used to guide computational modeling, as in Publication~\ref{NR},
where pathway information derived from the KEGG pathway database is
used to guide organism-wide discovery and analysis of transcriptional
response patterns.

\subsubsection{Evolving biological knowledge}

The collective knowledge about genome organization and function is
constantly updated and refined by improved measurement techniques and
accumulation of data \citep{Sebat07}. This can alter the analysis and
interpretation of results from large-scale genomic screens. For
instance, evolving gene and transcript definitions are known to
significantly affect microarray interpretation. Probe design on
microarray technology relies on sequence annotations that may have
changed significantly after the original array design.
Reinterpretation of microarray data based on updated probe annotations
has been shown to improve the accuracy and comparability of microarray
results \citep{Dai05, Hwang04, Mecham04b}. Bioinformatics studies
routinely take into account updates in genome version, {\it genome
  build}, in new analyses. The constantly refined biological data
highlights the need to account for this uncertainty in computational
analyses. In Publications~\ref{PECA} and~\ref{RPA}, explicit
computational strategies that are robust against evolving transcript
definitions are developed for microarray data analysis.

\subsection{Challenges in high-throughput data analysis} 

High-throughput genetic screens are inherently noisy. Controlling all
potential sources of variation in the measurement process is
increasingly difficult when automated measurement techniques can
produce millions of data points in a single experiment, concerning
extremely complex living systems that are to a large extent poorly
understood.

Noise arises from both technical and biological sources
\citep{Butte02a}, and systematic variation between laboratories,
measurement batches and measurement platforms has to be taken into
account when combining the results across individual studies
\citep{Heber06, Shi06}. Moreover, genomic knowledge is constantly
evolving, which can potentially change the interpretation of previous
experiments \citep[see e.g.][]{Dai05}. The various sources of noise
and uncertainty in microarray studies are discussed in more detail in
Chapter~\ref{ch:preproc}.

High dimensionality of the data and small sample size form another
challenge for the analysis of high-throughput functional genomics
data. Tens of thousands of transcripts can be measured simultaneously
in a single microarray experiment, which greatly exceeds the number of
available samples in most biomedical studies. Small sample sizes leave
considerable uncertainty in the analyses; few observations contain
very limited information concerning the complex and high-dimensional
phenomena and potential interactions between different parts of the
system.  Overfitting of the models and the problem of multiple testing
forms considerable challenges in such situations.  While automated
analysis methods can generate thousands of hypotheses concerning the
system, prioritizing the findings and characterizing uncertainty in
the predictions become central issues in the analysis.  The {\it curse
  of dimensionality}, coupled with the high levels of noise in
functional genomics studies, is therefore posing particular challenges
for computational modeling \citep{Saeys2007}.

The challenges in controlling the various sources of uncertainty have
led to remarkable problems in reproducing microarray results
\citep{Ioannidis09}, but maturing technology and the development of
common standards and analytical procedures are constantly improving
the reliability of high-throughput screens \citep{Allison06,
  Reimers2010, Shi06}. The models developed in this thesis combine
statistical evidence across related experiments to improve the
reliability of the analysis and to increase modeling power.
Generative probabilistic models provide a rigorous framework for
handling noise and uncertainty in the data and models.

\section{Genomics and health}

Genomic variation between individuals has remarkable and to a large
extent unknown contribution to health and disease
susceptibility. Large-scale characterization of the variability
between individuals and populations is expected to elucidate genomic
mechanisms associated with disease, as well as to lead to the
discovery of novel medical treatments. High-throughput genomics can
provide new tools to understand disease mechanisms
\citep{Braga-Neto06, Lage08}, to 'hack the genome' \citep{Evanko06} to
treat diseases \citep{Volinia2010}, and to guide {\it personalized
  therapies} that take into account the individual variability in
sensitivity and responses to treatments \citep{Church05, Downward06,
  Foekens08, Ocana2010, vantVeer08}. Disease signatures are
potentially robust across tissues and experiments \citep{Dudley09,
  Hu06}. Genomic screens have revealed new disease subtypes
\citep{Bhattacharjee01}, and led to the discovery of various {\it
  diagnostic} \citep{Lee08c, Su09, Tibshirani02} and {\it prognostic}
\citep{Beer02} biomarkers. Diseases cause coordinated changes in gene
activity through biomolecular networks \citep{Cabusora05}. Integration
of chemical, genomic and pharmacological functional genomics data can
also help to predict new drug targets and responses \citep{Lamb06,
  Yamanishi2010}. Genomic mutations can also affect genome function
and cause diseases \citep{Taylor08}. Cancer is an example of a
prevalent genomic disease. \cite{Boveri1914} discovered that cancer
cells have chromosomal imbalances, and since then the understanding of
genomic changes associated with cancer has continuously improved
\citep{Stratton09, Wunderlich2007}.  For instance, many human
micro-RNA genes are located at cancer-associated genomic regions and
are functionally altered in cancers \citep[see][]{Calin06}. Genomic
changes also affect transcriptional activity of the genes
\citep{Myllykangas08jc}. Publication~\ref{MLSP} introduces a novel
computational approach for screening cancer-associated DNA mutations
with functional implications by genome-wide integration of chromosomal
aberrations and transcriptional activity.

This chapter has provided an overview to central modeling challenges
and research topics in functional genomics. In the following chapters,
particular methodological approaches are introduced to solve research
tasks in large-scale analysis of the human transcriptome. In
particular, methods are introduced to increase the reliability of
high-throughput measurements, to model large-scale collections of
transcriptome data and to integrate transcriptional profiling data to
other layers of genomic information. The next chapter provides general
methodological background for these studies.

\newpage

\chapter{Statistical learning and exploratory data analysis}\label{ch:meth}

\begin{quotation}
\emph{Essentially, all models are wrong, but some are useful.}
\begin{flushright}
G.E.P. Box and N.R. Draper (1987)
\end{flushright}
\end{quotation}

Models are condensed, simplified representations of observed
phenomena. Models can be used to describe observations and to predict
future events. Two key aspects in modeling are the construction and
learning of formal representations of the observed data. Complex
real-world observations contain large amounts of uncontrolled
variation, which is often called {\it noise}; all aspects of the data
cannot be described within a single model. Therefore, a {\it modeling
compromise} is needed to decide what aspects of data to describe and
what to ignore. The second step in modeling is to fill in, to {\it
learn}, details of the formal representation based on the actual
empirical observations. Various learning algorithms are typically
available that differ in efficiency and accuracy. For instance,
improvements in computation time can often be achieved by potential
decrease in accuracy. An {\it inference compromise} is needed to
decide how to balance between these and other potentially conflicting
objectives of the learning algorithm; the relative importance of each
factor depends on the particular application and available resources,
and affects the choice of the learning procedure. The modeling and
inference compromises are at the heart of data analysis. Ultimately,
the value of a model is determined by its ability to advance the
solving of practical problems. 

This chapter gives an overview of the key concepts in statistical
modeling central to the topics of this thesis. The objectives of
exploratory data analysis and statistical learning are considered in
Section~\ref{sec:tasks}. The methodological framework is introduced in
Section~\ref{sec:prob}, which contains an overview of central concepts
in probabilistic modeling and the Bayesian analysis paradigm. Key
issues in implementing and validating the models are discussed in
Section~\ref{sec:learning}.

\section{Modeling tasks}\label{sec:tasks}

Understanding requires generalization beyond particular observations.
While empirical observations contain information of the underlying
process that generated the data, a major challenge in computational
modeling is that empirical data is always finite and contains only
limited information of the system. Traditional statistical models are
based on careful hypothesis formulation and systematic collection of
data to support or reject a given hypothesis. However, successful
hypothesis formulation may require substantial prior knowledge. When
minimal knowledge of the system is available, there is a need for {\it
  exploratory methods} that can recognize complex patterns and extract
features from empirical data in an automated way \citep{Baldi99}.
This is a central challenge in computational biology, where the
investigated systems are extremely complex and contain large amounts
of poorly characterized and uncontrolled sources of
variation. Moreover, the data of genomic systems is often very limited
and incomplete. General-purpose algorithms that can learn relevant
features from the data with minimal assumptions are therefore needed,
and they provide valuable tools in functional genomics
studies. Classical examples of such exploratory methods include
clustering, classification and visualization techniques. The extracted
features can provide hypotheses for more detailed experimental testing
and reveal new, unexpected findings. In this work, general-purpose
exploratory tools are developed for central modeling tasks in
functional genomics.

\subsection{Central concepts in data analysis}

Let us start by defining some of the basic concepts and terminology.
\emph{Data set} in this thesis refers to a finite collection of
observations, or {\it samples}. In experimental studies, as in
biology, a sample typically refers to the particular object of study,
for instance a patient or a tissue sample. In computational studies,
sample refers to a numerical observation, or a subset of observations,
represented by a numerical {\it feature vector}. Each element of the
feature vector describes a particular {\it feature} of the
observation. Given \(D\) features and \(N\) samples, the data set is
presented as a matrix $\X \in \Real^{D \times N}$, where each column
vector $\x \in \Real^D$ represents a sample and each row corresponds
to a particular feature. The features can represent for instance
different experimental conditions, time points, or particular
summaries about the observations. This is the general structure of the
observations investigated in this work.

The observations are modeled in terms of probability densities; the
samples are modeled as independent instances of a random variable.  A
central modeling task is to characterize the underlying probability
density of the observations, \(p(\x)\). This defines a topology in the
sample space and provides the basis for generalization beyond
empirical observations. As explained in more detail in
Section~\ref{sec:prob}, the models are formulated in terms of
observations \(\X\), model parameters $\bth$, and {\it latent
  variables} \(\Z\) that are not directly observed, but characterize
the underlying process that generated the data.

Ultimately, all models describe relationships between objects.  {\it
Similarity} is therefore a key concept in data analysis; the basis for
characterizing the relations, for summarizing the observations, and
for predicting future events. Measures of similarity can be defined
for different classes of objects such as feature vectors, data sets,
or random variables. Similarity in general is a vague concept. {\it
Euclidean distance}, induced by the Euclidean metrics, is a common
(dis-)similarity measure for multivariate observations. {\it
Correlation} is a standard choice for univariate random variables.
{\it Mutual information} is an information-theoretic measure of
statistical dependency between two random variables, characterizing
the decrease in the uncertainty concerning the realization of one
variable, given the other one. The uncertainty of a random variable
\(\Xcal\) is measured in terms of {\it entropy}\footnote{Entropy is
defined as \(H(\Xcal) = - \int_{\x} p(\x)\log p(\x) d\x\) for a
continuous variable.} \citep{Shannon48}. The mutual information
between two random variables is then given by \(I(\Xcal, \Ycal) =
H(\Xcal) - H(\Xcal|\Ycal)\) \citep[see e.g.][]{Gelman03}. The
Kullback-Leibler divergence, or {\it KL--divergence}, is a closely
related non-symmetric dissimilarity measure for probability
distributions \(p, q\), defined as \(d_{KL} (p,q) = \int_{\x} p(\x)
\log \frac{p(\x)}{q(\x)}d\x\) \citep[see e.g.][]{Bishop06}. Mutual
information between two random variables can be alternatively
formulated as the KL--divergence between their joint density
\(p(\x,\y)\) and the product of their independent marginal densities,
\(q(\x,\y) = p_x(\x)p_y(\y)\), which gives the connection \(I(\Xcal,
\Ycal) = d_{KL}(p(\x,\y),p_x(\x)p_y(\y))\). Mutual information and
KL-divergence are central information-theoretic measures of dependency
employed in the models of this thesis.

It is important to notice that measures of similarity are inherently
coupled to the statistical representation of data and to the goals of
the analysis; different representations can reveal different
relationships between observations. For instance, the Euclidean
distance is sensitive to scaling of the features; representation in
natural or logarithmic scale, or with different units can potentially
lead to very different analysis results. Not all measures are equally
sensitive; mutual information can naturally detect non-linear
relationships, and it is invariant to the scale of the variables. On
the other hand, estimating mutual information is computationally
demanding.

{\it Feature selection} refers to computational techniques for
selecting, scaling and transforming the data into a suitable form for
further analysis. Feature selection has a central role in data
analysis, and it is implicitly present in all analysis tasks in
selecting the investigated features for the analysis. 

There are no universally optimal stand-alone feature selection
techniques, since the problem is inherently entangled with the
analysis task and multiple equally optimal feature sets may be
available for instance in classification or prediction tasks
\cite{Guyon03, Saeys2007}. Successful feature selection can reduce the
dimensionality of the data with minimal loss of relevant information,
and focus the analysis on particular features. This can reduce model
complexity, which is expected to yield more efficient, generalizable
and interpretable models. Feature selection is particularly important
in genome-wide profiling studies, where the dimensionality of the data
is large compared to the number of available samples, and only a small
number of features are relevant for the studied phenomenon. This is
also known as the {\it large p, small n} problem \citep{West03}.
Advanced feature selection techniques can take into account
dependencies between the features, consider weighted combinations of
them, and can be designed to interact with the more general modeling
task, as for instance in the nearest shrunken centroids classifier of
\cite{Tibshirani02}. The constrained subspace clustering model of
Publication~\ref{NR} can be viewed as a feature selection procedure,
where high-dimensional genomic observations are decomposed into
distinct feature subsets, each of which reveals different
relationships of the samples. In Publication~\ref{MLSP},
identification of maximally informative features between two data sets
forms a central part of a regularized dependency modeling
framework. In Publications~\ref{NR}-\ref{MLSP} the procedure and
representations are motivated by biological reasoning and analysis
goals.

\subsection{Exploratory data analysis}

{\it Exploratory data analysis} refers to the use of computational
techniques to summarize and visualize data in order to facilitate the
generation of new hypotheses for further study when the search space
would be otherwise exhaustively large \citep{Tukey77}. The analysis
strategy takes the observations as the starting point for discovering
interesting regularities and novel research hypotheses for poorly
characterized large-scale systems without prior knowledge. The
analysis can then proceed from general observations of the data toward
\emph{confirmatory data analysis}, more detailed investigations and
hypotheses that can be tested in independent data sets with standard
statistical procedures. Exploratory data analysis differs from
traditional hypothesis testing where the hypothesis is
given. Light-weight exploratory tools are particularly useful with
large data sets when prior knowledge on the system is
minimal. Standard exploratory approaches in computational biology
include for instance clustering, classification and visualization
techniques \citep{Evanko10, Polanski07}.

{\it Cluster analysis} refers to a versatile family of methods that
partition data into internally homogeneous groups of similar data
points, and often at the same time minimize the similarity between
distinct clusters. Clustering techniques enable {\it class discovery}
from the data. This differs from classification where the target is to
assign new observations into known classes. The partitions provided by
clustering can be nested, partially overlapping or mutually exclusive,
and many clustering methods generalize the partitioning to cover
previously unseen data points \citep{Jain88}.  Clustering can provide
compressed representations of the data based on a shared parametric
representation of the observations within each cluster, as for
instance in K-means or Gaussian mixture modeling \citep[see
e.g.][]{Bishop06}. Certain clustering approaches, such as the
hierarchical clustering \citep[see e.g.][]{Hastie09}, apply recursive
schemes that partition the data into internally homogeneous groups
without providing a parametric representation of the clusters.
Cluster structure can be also discovered by linear algebraic
operations on the distance matrices, as for instance in spectral
clustering. The different approaches often have close theoretical
connections. Clustering in general is an ill-defined concept that
refers to a set of related but mutually incompatible objectives
\citep{Ben-David08, Kleinberg02nips}. Cluster analysis has been
tremendously popular in computational biology, and a comprehensive
review of the different applications are beyond the scope of this
thesis. It has been observed, for instance, that genes with related
functions have often similar expression profiles and are clustered
together, suggesting that clustering can be used to formulate
hypotheses concerning the function of previously uncharacterized genes
\citep{DeRisi97, Eisen98}, or to discover novel cancer subtypes with
biomedical implications \citep{Sorlie01}.

{\it Visualization techniques} are another widely used exploratory
approach in computational biology. Visualizations can provide compact
and intuitive summaries of complex, high-dimensional observations on a
lower-dimensional display, for instance by linear projection methods
such as principal component analysis, or by explicitly optimizing a
lower-dimensional representation as in the self-organizing map
\citep[][]{Kohonen82}. Visualization can provide the first step in
investigating large data sets \citep{Evanko10}.

\subsection{Statistical learning}

\emph{Statistical learning} refers to computational models that can
learn to recognize structure and patterns from empirical data in an
automated way. Unsupervised and supervised models form two main
categories of learning algorithms.

{\it Unsupervised learning} approaches seek compact descriptions of
the data without prior knowledge. In probabilistic modeling,
unsupervised learning can be formulated as the task of finding a
probability distribution that describes the observed data and
generalizes to new observations.  This is also called {\it density
estimation}.  The parameter values of the model can be used to provide
compact representations of the data. Examples of unsupervised analysis
tasks include methods for clustering, visualization and dimensionality
reduction.  In cluster analysis, groups of similar observations are
sought from the data. Dimensionality reduction techniques provide
compact lower-dimensional representations of the original data, which
is often useful for subsequent modeling steps. Not all observations of
the data are equally valuable, and assessing the relevance of the
observed regularities is problematic in fully unsupervised analysis.

In {\it supervised learning} the task is to learn a function that maps
the inputs \(\x\) to the desired outputs \(\y\) based on a set of
training examples in a generalizable fashion, as in regression for
continuous outputs, and classification for discrete output variables.
The supervised learning tasks are inherently asymmetric; the inference
proceeds from inputs to outputs, and prior information of the modeling
task is used to supervise the analysis; the training examples also
include a desired output of the model.

The models developed in this thesis can be viewed as unsupervised
exploratory techniques. However, the distinction between supervised
and unsupervised models is not strict, and the models in this thesis
borrow ideas from both categories. The models in
Publications~\ref{RPA}-\ref{NR} are unsupervised algorithms that
utilize prior information derived from background databases to guide
the modeling by constraining the solutions.  However, since no desired
outputs are available for these models, the modeling tasks differ from
supervised analysis. The dependency modeling algorithms of
Publications~\ref{MLSP}-\ref{AC} have close theoretical connections to
the supervised learning task. In contrast to supervised learning, the
learning task in these algorithms is symmetric; modeling of the
co-occurring data sets is unsupervised, but coupled. Each data set
affects the modeling of the other data set in a symmetric fashion,
and, in analogy to supervised learning, prediction can then proceed to
either direction. Compared to supervised analysis tasks, the emphasis
in the dependency detection algorithms introduced in this thesis is in
the discovery and characterization of symmetric dependencies, rather
than in the construction of asymmetric predictive models.

\section{Probabilistic modeling paradigm}\label{sec:prob}

The main contributions of this thesis follow the generative
probabilistic modeling paradigm.  Generative probabilistic models
describe the observed data in terms of probability distributions. This
allows the calculation of expectations, variances and other standard
summaries of the model parameters, and at the same time allows to
describe the independence assumptions and relations between variables,
and uncertainty in the modeling process in an explicit manner.
Measurements are regarded as noisy observations of the general,
underlying processes; generative models are used to characterize the
processes that generated the observations.

The first task in modeling is the selection of a \emph{model family} -
a set of potential formal representations of the data. As discussed in
Section~\ref{sec:nonparametric}, the representations can also to some
extent be learned from the data. The second task is to define the
\emph{objective function}, or cost function, which is used to measure
the descriptive power of the models. The third task is to identify the
optimal model within the model family that best describes the observed
data with respect to the objective function. This is called {\it
  learning} or {\it model fitting}. The details of the modeling
process are largely determined by the exact modeling task and
particular nature of the observations.  The objectives of the modeling
task are encoded in the selected model family, the objective function
and to some extent to the model fitting procedure. The model family
determines the space of possible descriptions for the data and has
therefore a major influence on the final solution. The objective
function can be used to prefer simple models or other aspects in the
modeling process. The model fitting procedure affects the efficiency
and accuracy of the learning process.  For further information of
these and related concepts, see \cite{Bishop06}. A general overview of
the probabilistic modeling framework is given in the remainder of this
section.

\subsection{Generative modeling}\label{sec:generative}

\emph{Generative probabilistic models} view the observations as random
samples from an underlying probability distribution. The model defines
a probability distribution \(p(\x)\) over the feature space.  The
model can be parameterized by model parameters $\bth$ that specify a
particular model within the model family.  For convenience, we assume
that the model family is given, and leave it out from the notation. In
this thesis, the appropriate model families are selected based on
biological hypotheses and analysis goals. Generative models allow
efficient representation of dependencies between variables,
independence assumptions and uncertainty in the inference
\citep{Koller09}. Let us next consider central analysis tasks in
generative modeling.

\subsubsection{Finite mixture models}\label{sec:finitemixtures}

Classical probability distributions provide well-justified and
convenient tools for probabilistic modeling, but in many practical
situations the observed regularities in the data cannot be described
with a single standard distribution. However, a sufficiently rich
mixture of standard distributions can provide arbitrarily accurate
approximations of the observed data. In {\it mixture models}, a set of
distinct, latent processes, or {\it components}, is used to describe
the observations. The task is to identify and characterize the
components and their associations to the individual observations. The
standard formulation assumes independent and identically distributed
observations where each observation has been generated by exactly one
component. In a standard mixture model the overall probability density
of the data is modeled as a weighted sum of component distributions:

\begin{equation}\label{eq:mixture}
   p(\x) = \sum_{r=1}^R \pi_r p_r(\x | \bth_r),
\end{equation}

\noindent where the components are indexed by \(r\), and $\int p(\x)
d\x = 1$. Each mixture component can have a different distributional
form.  The mixing proportion, or weight, and model parameters of each
component are denoted by $\pi_r$ and \(\bth_r\), respectively, with
$\sum_r \pi_r = 1$. Many applications utilize convenient standard
distributions, such as Gaussians, or other distributions from the
exponential family. Then the mixture model can be learned for instance
with the EM algorithm described in Section~\ref{sec:em}.

In practice, the mixing proportions of the components are often
unknown. The mixing proportions can be estimated from the data by
considering them as standard model parameters to be fitted with a ML
estimate. However, the procedure is potentially prone to overfitting
and local optima, i.e., it may learn to describe the training data
well, but fails to generalize to new observations. An alternative,
probabilistic way to determine the weights is to treat the mixing
proportions as latent variables with a prior distribution
\(p(\bpi)\). A standard choice is a symmetric Dirichlet
prior\footnote{{\it Dirichlet distribution} is the probability density
  \(Dir(\bpi | \n) \sim \prod_r \pi_r^{n_r - 1}\) where the
  multivariate random variable \(\bpi\) and the positive parameter
  vector \(\n\) have their elements indexed by \(r\), \(0 < \pi_r <
  1\), and \(\sum_r \pi_r = 1\).}  \(\bpi \sim
Dir(\frac{\alp}{R})\). This gives an equal prior weight for each
component and guarantees the standard exchangeability assumption of
the mixture component labels. A label determines cluster
identity. Intuitively, exchangeability corresponds to the assumption
that the analysis is invariant to the ordering of the data samples and
mixture components. Compared to standard mixture models, probabilistic
mixture models have increased computational complexity.

Further prior knowledge can be incorporated in the model by defining
prior distributions for the other parameters of the mixture
model. This can also be used to regularize the learning process to
avoid overfitting.  A typical prior distribution for the components of
a Gaussian mixture model, parameterized by \(\bth_r = \{\bmu_r,
\bSigma_r\}\), is the normal-inverse-Gamma prior \citep[see
e.g.][]{Gelman03}.

Interpreting the mixture components as clusters provides an
alternative, probabilistic formulation of the clustering task. This
has made probabilistic mixture models a popular choice in the analysis
of functional genomics data sets that typically have high
dimensionality but small sample size.  Probabilistic analysis takes
the uncertainties into account in a rigorous manner, which is
particularly useful when the sample size is small. The number of
mixture components is often unknown in practical modeling tasks,
however, and has to be inferred based on the data. A straightforward
solution can be obtained by employing a sufficiently large number of
components in learning the mixture model, and selecting the components
having non-zero weights as a post-processing step. An alternative,
model-based treatment for learning the number of mixture components
from the data is provided by infinite mixture models considered in
Section~\ref{sec:nonparametric}.

\subsubsection{Latent variables and marginalization}

The observed variables are often affected by {\it latent variables}
that describe relevant structure in the model, but are not directly
observed. The latent variable values can be, to some extent, inferred
based on the observed variables. Combination of latent and observed
variables allows the description of complex probability spaces in
terms of simple component distributions and their relations. Use of
simple component distributions can provide an intuitive and
computationally tractable characterization of complex generative
processes underlying the observations.

A generative latent variable model specifies the distributional form
and relationships of the latent and observed variables. As a simple
example, consider the probabilistic interpretation of probabilistic
component analysis (PCA), where the observations \(\x\) are modeled
with a linear model \(\x = \W\z + \Epsilon\) where a normally
distributed latent variable \(\z \sim N(\0, \I)\) is transformed with
the parameter matrix \(\W\) and isotropic Gaussian noise
(\(\Epsilon\)) is assumed on the observations. More complex models can
be constructed by analogous reasoning. A {\it complete-data
  likelihood} \(p(\X,\Z|\bth)\) defines a joint density for the
observed and latent variables. Only a subset of variables in the model
is typically of interest for the actual analysis task. For instance,
the latent variables may be central for describing the generative
process of the data, but their actual values may be irrelevant. Such
variables are called {\it nuisance variables}. Their integration, or
{\em marginalization}, provides probabilistic averaging over the
potential realizations.  Marginalization over the latent variables in
the complete-data likelihood gives the likelihood

\begin{equation}\label{eq:marginalization}
  p(\X|\bth)=\int_{\Z} p(\X,\Z|\bth)d\Z.
\end{equation}

Marginalization over the latent variables collapses the modeling task
to finding optimal values for model parameters \(\bth\), in a way that
takes into account the uncertainty in latent variable values. This can
reduce the number of variables in the learning phase, yield more
straightforward and robust inferences, as well as speed up
computation. However, marginalization may lead to analytically
intractable integrals. As certain latent variables may be directly
relevant, marginalization depends on the overall goals of the analysis
and may cover only a subset of the latent variables. In this thesis
latent variables are utilized for instance in Publication~\ref{NR},
which treats the sample-cluster assignments as discrete latent
variables, as well as in Publication~\ref{MLSP}, where a regularized
latent variable model is introduced to model dependencies between
co-occurring observations.

\subsection{Nonparametric models}\label{sec:nonparametric}

Finite mixture models and latent variable models require the
specification of model structure prior to the analysis. This can be
problematic since for instance the number and distributional shape of
the generative processes is unknown in many practical tasks. However,
the model structure can also to some extent be learned from the data.
Non-parametric models provide principled approaches to learn the model
structure from the data.  In contrast to parametric models, the number
and use of the parameters in nonparametric models is flexible
\citep[see e.g.][]{Hjort10, Muller2004}. The infinite mixture of
Gaussians, used as a part of the modeling process in
Publication~\ref{NR}, is an example of a non-parametric model where
both the number of components, as well as mixture proportions of the
component distributions are inferred from the data. Learning of
Bayesian network structure is another example of nonparametric
inference, where relations between the model variables are learned
from the data \citep[see e.g.][]{Friedman03}. While more complex
models can describe the training data more accurately, an increasing
model complexity needs to be penalized to avoid overfitting and to
ensure generalizability of the model.

Nonparametric models provide flexible and theoretically principled
approaches for data-driven exploratory analysis. However, the
flexibility often comes with an increased computational cost, and the
models are potentially more prone to overfitting than less flexible
parametric models. Moreover, complex models can be difficult to
interpret.

Many nonparametric probabilistic models are defined by using the
theory of stochastic processes to impose priors over potential model
structures. Stochastic processes can be used to define priors over
function spaces. For instance, the {\it Dirichlet process (DP)}
defines a probability density over the function space of Dirichlet
distributions\footnote{If \(G\) is a distribution drawn from a
  Dirichlet process with the probability measure \(P\) over the sample
  space, \(G \sim \mathrm{DP}(P)\), then each finite partition
  \(\{A_k\}_k\) of the sample space is distributed as
  \((G(A_1),...,G(A_k)) \sim Dir(P(A_1),..., P(A_k))\).}. The {\it
  Chinese Restaurant Process (CRP)} provides an intuitive description
of the Dirichlet process in the cluster analysis context. The CRP
defines a prior distribution over the number of clusters and their
size distribution. The CRP is a random process in which $n$ customers
arrive in a restaurant, which has an infinite number of tables. The
process goes as follows: The first customer chooses the first
table. Each subsequent customer \(m\) will select a table based on the
state \(F_{m-1}\) of the restaurant tables after $m-1$ customers have
arrived. The new customer \(m\) will select a previously occupied
table $i$ with a probability which is proportional to the number of
customers seated at table $i$, i.e. \(p(i|F_{m-i}) \propto
n_i\). Alternatively, the new customer will select an empty table with
a probability which is proportional to a constant \(\alpha\). The
model prefers tables with a larger number of customers, and is
analogous to clustering, where the customers and tables correspond to
samples and clusters, respectively. This provides an intuitive prior
distribution for clustering tasks. The prior prefers compact models
with relatively few clusters, but the number of clusters is
potentially infinite, and ultimately determined based on the data.

\subsubsection{Infinite mixture models}

{\it Infinite mixture models} are a general class of nonparametric
methods where the number of mixture components are determined in a
data-driven manner; the number of components is potentially infinite
\citep[see e.g.][]{Muller2004, Rasmussen00}. An infinite mixture is
obtained by letting \(R \rightarrow \infty\) in the finite mixture
model of Equation~\ref{eq:mixture} and replacing the Dirichlet
distribution prior of the mixing proportions \(\bpi\) by a Dirichlet
process. The formal probability distribution of the Dirichlet process
can be intuitively derived with the so-called {\it stick-breaking
presentation}.  Consider a unit length stick and a stick-breaking
process, where the breakpoint \(\beta\) is stochastically determined,
following the beta distribution $\beta \sim Beta(1, \alpha)$, where
\(\alpha\) tunes the expected breaking point. The process can be
viewed as consecutively breaking off portions of a unit length stick
to obtain an infinite sequence of stick lengths \(\pi_1 = \beta_1\);
\(\pi_i = \beta_i \prod_{l=1}^{i-1} (1-\beta_l)\), with
$\sum_{i=1}^\infty \pi_i = 1$~\citep{Ishwaran01}.  This defines the
probability distribution \(\text{Stick}(\alpha)\) over potential
partitionings of the unit stick. A truncated stick-breaking
representation considers only the first $T$ elements. Setting the
prior \(\bpi \sim \text{Stick}(\alpha)\), defined by the
stick-breaking representation in Equation~\ref{eq:mixture} assigns a
prior on the number of mixture components and their mixing proportions
that are ultimately learned from the observed data. The prior helps to
find a compromise between increasing model complexity and likelihood
of the observations.

Traditional approaches used to determine the number mixture components
are based on objective functions that penalize increasing model
complexity, for instance in certain variants of the K-means or in
spectral clustering \citep[see e.g.][]{Hastie09}. Other model
selection criteria include cross-validation and comparison of the
models in terms of their likelihood or various information-theoretic
criteria that seek a compromise between model complexity and fit
\citep[see e.g.][]{Gelman03}. However, the sample size may be
insufficient for such approaches, and the models may lack a rigorous
framework to account for uncertainties in the observations and model
parameters. Modeling uncertainty in the parameters while learning the
model structure can lead to more robust inference in nonparametric
probabilistic models but also adds inherent computational complexity
in the learning process.

\subsection{Bayesian analysis}\label{sec:bayes}

The term 'Bayesian' refers to interpretation of model parameters as
variables. The uncertainty over the parameter values, arising from
limited empirical evidence, is described in terms of probability
distributions. This is in contrast to the traditional view where
parameters have fixed values with no distribution and the uncertainty
is ignored. The Bayesian approach leads to a learning task where the
objective is to estimate the \emph{posterior distribution} $p(\bth|\X)$
of the model parameters \(\bth\), given the observations \(\X\). The
posterior is given by the \emph{Bayes' rule} \citep{Bayes63}:

\begin{equation}\label{eq:bayesrule}
  p(\bth|\X) = \frac{p(\X|\bth)p(\bth)}{p(\X)}.
\end{equation}

\noindent The two key elements of the posterior are {\it the
likelihood} and {\it the prior}. The likelihood \(p(\X|\bth)\)
describes the probability of the observations, given the parameter
values \(\bth\). The parameters can also characterize alternative
model structures. The prior \(p(\bth)\) encodes prior beliefs about
the model and rewards solutions that match with the prior assumptions or
yield simpler models. Such regularizing properties can be particularly
useful when training data is scarce and there is considerable
uncertainty in the parameter estimates. With strong, informative
priors, new observations have little effect on the posterior. In the
limit of large sample size the posterior converges to the ordinary
likelihood \(p(\X|\bth)\). The Bayesian inference provides a robust
framework for taking the uncertainties into account when the data is
scarce, as it often is in practical modeling tasks.  Moreover, the
Bayes' rule provides a formal framework for sequential update of
beliefs based on accumulating evidence. The prior predictive density
$p(\X) = \int p(\X,\bth) d\bth$ is a normalizing constant, which is
independent of the parameters $\bth$ and can often be ignored during
model fitting. 

The involved distributions can have complex non-standard forms and
limited empirical data can only provide partial evidence regarding the
different aspects of the data-generating process.  Often only a subset
of the parameters and other variables and their interdependencies can
be directly observed. The Bayesian approach provides a framework for
making inferences on the unobserved quantities through hierarchical
models, where the probability distribution of each variable is
characterized by higher-level parameters, so-called {\it
hyperparameters}. A similar reasoning can be used to model the
uncertainty in the hyperparameters, until the uncertainties become
modeled at an appropriate detail. Prior information can help to
compensate the lack of data on certain aspects of a model, and
explicit models for the noise can characterize uncertainty in the
empirical observations. Distributions can also share parameters, which
provides a basis for pooling evidence from multiple sources, as for
instance in Publication~\ref{MLSP}. In many applications only a subset
of the parameters in the model are of interest and the modeling
process can be considerably simplified by marginalizing over the less
interesting nuisance variables to obtain an expectation over their
potential values. 

The Bayesian paradigm provides a principled framework for modeling the
uncertainty at all levels of statistical inference, including the
parameters, the observed and latent variables and the model structure;
all information of the model is incorporated in the posterior
distribution, which summarizes empirical evidence and prior knowledge,
and provides a complete description of the expected outcomes of the
data-generating process. When the data does not contain sufficient
information to decide between the alternative model structures and
parameter values, the Bayesian framework provides tools to take
expectations over all potential models, weighted by their relative
evidence.

A central challenge in the Bayesian analysis is that the models often
include analytically intractable posterior distributions, and learning
of the models can be computationally demanding. Widely-used approaches
for estimating posterior distributions include {\it Markov Chain Monte
Carlo (MCMC)} methods and variational learning. Stochastic MCMC
methods provide a widely-used family of algorithms to estimate
intractable distributions by drawing random samples from these
distributions \citep[see e.g.][]{Gelman03}; a sufficiently large pool
of random samples will converge to the underlying distribution, and
sample statistics can then be used to characterize the distribution.
However, sampling-based methods are computationally intensive and
slow. In variational learning, considered in
Section~\ref{sec:variational}, the intractable distributions are
approximated by more convenient tractable distributions, which yields
faster learning procedure, but potentially less accurate results.
While analysis of the full posterior distribution will provide a
complete description of the uncertainties regarding the parameters,
simplified summary statistics, such as the mean, variance and
quantiles of the posterior can provide a sufficient characterization
of the posterior in many practical applications. They can be obtained
for instance by summarizing the output of sampling-based or
variational methods. Moreover, when the uncertainty in the results can
be ignored, point estimates can provide simple, interpretable
summaries that are often useful in further biomedical analysis, as for
instance in Publication~\ref{RPA}. Point estimates are single optimal
values with no distribution.  However, point estimates are not
necessarily sufficient for instance in biomedical diagnostics and
other prediction tasks, where different outcomes are associated with
different costs and it may be crucial to assess the probabilities of
the alternative outcomes. For further discussion on learning the
Bayesian models, see Section~\ref{sec:fitting}.

In this thesis the Bayesian approach provides a formal framework to
perform robust inference based on incomplete functional genomics data
sets and to incorporate prior information of the models in the
analysis. The Bayesian paradigm can alternatively be interpreted as a
philosophical position, where probability is viewed as a subjective
concept \citep{Cox46}, or considered a direct consequence of making
rational decisions under uncertainty \citep{Bernardo00}. For further
concepts in model selection, comparison and averaging in the Bayesian
analysis, see \cite{Gelman03}. For applications in computational
biology, see \cite{Wilkinson2007}.

\section{Learning and inference}\label{sec:learning}

The final stage in probabilistic modeling is to learn the optimal
statistical presentation for the data, given the model family and the
objective function. This section highlights central challenges and
methodological issues in statistical learning. 

\subsection{Model fitting}\label{sec:fitting}

{\it Learning} in probabilistic models often focuses on optimizing the
model parameters \(\bth\). In addition, posterior distribution of the
latent variables, \(p(\z|\x, \bth)\), can be calculated. Estimating
the latent variable values is called statistical {\it inference}. In
the Bayesian analysis, the model parameters can also be treated as
latent variables with a prior probability density, in which case the
distinction between model parameters and latent variables will
disappear.  A comprehensive characterization of the variables and
their uncertainty would be achieved by estimating the full posterior
distribution. However, this can be computationally very demanding, in
particular when the posterior is not analytically tractable.  The
posterior is often approximated with stochastic or analytical
procedures, such as stochastic MCMC sampling methods or variational
approximations, and appropriate summary statistics. In many practical
settings, it is sufficient to summarize the full posterior
distribution with a point estimate. Point estimates do not
characterize the uncertainties in the analysis result, but are often
more convenient to interpret than full posterior distributions.

Various optimization algorithms are available to learn statistical
models, given the learning procedure. The potential challenges in the
optimization include {\it computational complexity} and the presence
of {\it local optima} on complex probability density topologies, as
well as {\it unidentifiability} of the models. Finding a global
optimum of a complex model can be computationally exhaustive, and it
can become intractable with increasing sample size. In unidentifiable
models, the data does not contain sufficient information to choose
between alternative models with equal statistical
evidence. Ultimately, the uncertainty in inference arises from limited
sample size and the lack of computational resources.

In the remainder of this section, let us consider more closely the
particular learning procedures central to this thesis: point estimates
and variational approximation, and the standard optimization
algorithms used to learn such representations.

\subsubsection{Point estimates}\label{sec:point}

Assuming independent and identically distributed observations, the
{\it likelihood} of the data, given model parameters, is \(p(\X|\bth)
= \prod_i p(\x_i|\bth)\). This provides a probabilistic measure of
model fit and the objective function to maximize.  Maximization of the
likelihood \(p(\X|\bth)\) with respect to \(\bth\) yields a {\it
  maximum likelihood (ML)} estimate of the model parameters, and
specifies an optimal model that best describes the data. This is a
standard point estimate used in probabilistic modeling. Practical
implementations typically operate on {\it log-likelihood}, the
logarithm of the likelihood function. As a monotone function, this
yields the same optima, but has additional desirable properties: it
factorizes the product into a sum and is less prone to numerical
overflows during optimization.

The {\it maximum a posteriori (MAP)} estimate additionally takes prior
information of the model parameters into account. While the ML
estimate maximizes the likelihood \(p(\X|\bth)\) of the observations,
the MAP estimate maximizes the posterior \(p(\bth|\X) \sim
p(\X|\bth)p(\bth)\) of the model parameters. The objective function to
maximize is the log-likelihood

\begin{equation}\label{eq:map}
  log p(\bth|\X) \sim log p(\X|\bth) + log p(\bth).
\end{equation}

The prior is explicit in MAP estimation and the model contains the ML
estimate as a special case; assuming large sample size, or
non-informative, uniform prior $p(\bth) \sim 1$, the likelihood of the
data \(p(\X|\bth)\) will dominate and the MAP estimation becomes
equivalent to optimizing \(p(\X|\bth)\), yielding the traditional ML
estimate. The ML and MAP estimates are asymptotically consistent
approximations of the posterior distribution, since the posterior will
converge a point distribution with a large sample size. The
computation and interpretation of point estimates is straightforward
compared to the use of posterior distributions in the full Bayesian
treatment. The differences between ML and MAP estimates highlight the
role of prior information in the modeling when training data is
limited.

\subsubsection{Variational inference}\label{sec:variational}

In certain modeling tasks the uncertainty in the model parameters
needs to be taken into account. Then point estimates are not
sufficient. The uncertainty is characterized by the posterior
distribution \(p(\bth|\X)\). However, the posterior distributions are
often intractable and need to be estimated by approximative methods.
{\it Variational approximations} provide a fast and principled
optimization scheme \citep[see e.g.][]{Bishop06} that yields only
approximative solutions, but can accelerate posterior inference by
orders of magnitude compared to stochastic, sampling-based MCMC
methods that can in principle provide exact solutions, assuming that
infinite computational resources are available. The potential decrease
in accuracy in variational approximations is often acceptable, given
the gains in efficiency. Variational approximation characterizes the
uncertainty in \(\bth\) with a tractable distribution \(q(\bth)\) that
approximates the full, potentially intractable posterior
\(p(\bth|\X)\),

Variational inference is formulated as an optimization problem where
an intractable posterior distribution \(p(\Z, \bth|\X)\) is
approximated by a more easily tract-able distribution \(q(\Z, \bth)\)
by minimizing the KL--divergence between the two distributions. This
is also shown to maximize a lower bound of the marginal likelihood
$p(\X)$, and subsequently the likelihood of the data, yielding an
approximation of the overall model.  The log-likelihood of the data
can be decomposed into a sum of the lower bound \(\L(q)\) of the
observed data and the KL--divergence \(d_{KL}(q, p)\) between the
approximative and the exact posterior distributions:

\begin{equation}\label{eq:variational}
            log p(\X) = \L(q) + d_{KL}(q,p),
\end{equation}

where 

\begin{flalign}
   \begin{array}{cll}
     \L(q) &=& \int_{\z} q(\Z, \bth) log \frac{p(\Z, \bth, \X)}{q(\Z,
     \bth)};\\
     d_{KL}(q,p) &=&-\int_{\z} q(\Z, \bth) log \frac{p(\Z,  \bth|\X)}{q(\Z, \bth)}.
  \end{array}
\end{flalign}

The KL-divergence is non-negative, and equals to zero if and only if
the approximation and the exact distribution are identical. Therefore
\(\L(q)\) gives a lower bound for the log-likelihood \(log p(\X)\) in
Equation~\ref{eq:variational}.  Minimization of \(d_{KL}\) with
respect to \(q\) will provide an analytically tractable approximation
$q(\Z, \bth)$ of $p(\Z, \bth | \X)$.  Minimization of \(d_{KL}\) will
also maximize the lower bound \(\L(q)\) since the log-likelihood \(log
p(\X)\) is independent of \(q\). The approximation typically assumes
independent parameters and latent variables, yielding a {\it
  factorized} approximation \(q(\Z, \bth) = q_{\z}(\Z)q_{\bth}(\bth)\)
based on tractable standard distributions. It is also possible to
factorize \(q_{\z}\) and \(q_{\bth}\) into further
components. Variational approximations are used for efficient learning
of infinite multivariate Gaussian mixture models in
Publication~\ref{NR}. 

\subsubsection{Expectation--Maximization (EM)}\label{sec:em}

The {\it EM algorithm} is a general procedure for learning
probabilistic latent variable models \citep{Dempster77}, and a special
case of variational inference. The algorithm provides an efficient
algorithm for finding point estimates for model parameters in latent
variable models. The objective of the EM algorithm is to maximize the
marginal likelihood

\begin{equation}\label{eq:emlikelihood}
           p(\X|\bth) = \int_{\z} p(\X,\Z|\bth) d\Z
\end{equation}
of the observations \(\X\) with respect to the model parameters
\(\bth\). Marginalization over the probability density of the latent
variables provides an inference procedure that is robust to
uncertainty in the latent variable values. The algorithm iterates
between estimating the posterior of the latent variables, and
optimizing the model parameters \citep[see e.g.][]{Bishop06}.  Given
initial values \(\bth_0\) of the model parameters, the {\it
  expectation step} evaluates the posterior density of the latent
variables, \(p(\z|\x,\bth_t)\), keeping \(\bth_t\) fixed. If the
posterior is not analytically tractable, variational approximation
\(q(\z)\) can be used to obtain a lower bound for the likelihood in
Equation~\ref{eq:emlikelihood}. The {\it maximization step} optimizes
the model parameters \(\bth\) with respect to the following objective
function:
\begin{equation}\label{EMQ}
  Q(\bth, \bth_t) = \int_{\z} p(\Z|\X, \bth_t) log p(\X, \Z| \bth) d\Z.
\end{equation}
This is the expectation of the {\it complete-data log-likelihood}
\(log p(\X, \Z| \bth)\) over the latent variable density \(p(\Z|\X,
\bth_t)\), obtained from the previous expectation step. The new
parameter estimate is then

\[ 
   \bth_{t+1} = argmax_{\bth} Q(\bth, \bth_t).
\] 

The expectation and maximization steps determine an iterative learning
procedure where the latent variable density and model parameters are
iteratively updated until convergence.  The maximization step will
also increase the target likelihood of Equation~\ref{eq:emlikelihood},
but potentially with a remarkably smaller computational cost
\citep{Dempster77}.  In contrast to the marginal likelihood in
Equation~\ref{eq:emlikelihood}, the complete-data likelihood in
Equation~\ref{EMQ} is logarithmized before integration in the
maximization step. When the joint distribution \(p(\x,\z|\bth)\)
belongs to the exponential family, the logarithm will cancel the
exponential in algebraic manipulations. This can considerably simplify
the maximization step. When the likelihoods in the optimization are of
suitable form, the iteration steps can be solved analytically, which
can considerably reduce required evaluations of the objective
function. Convergence is guaranteed, if the optimization can increase
the likelihood at each iteration. However, the identification of a
global optimum is not guaranteed in the EM algorithm.

Incorporating prior information of the parameter values through
Bayesian priors can be used to avoid overfitting and focus the
modeling on particular features in the data, as in the regularized
dependency modeling framework of Publication~\ref{MLSP}, where the EM
algorithm is used to learn Gaussian latent variable models.  

\subsubsection{Standard optimization methods}\label{sec:optim}

Optimization methods provide standard tools to implement selected
learning procedures. Optimization algorithms are used to identify
parameter values that minimize or maximize the objective function,
either globally, or in local surroundings of the optimized
value. Selection of optimization method depends on smoothness and
continuity properties of the objective function, required accuracy,
and available resources.  

{\it Gradient-based approaches} optimize the objective function by
assuming smooth, continuous topology over the probability density
where setting the derivatives to zero will yield local optima. If a
closed form solution is not available, it is often possible to
estimate gradient directions in a given point. Optimization can then
proceed by updating the parameters towards the desired direction along
the gradient, gradually improving the objective function value in
subsequent gradient ascent steps. So-called {\it quasi-Newton methods}
use function values and gradients to characterize the optimized
manifold, and to optimize the parameters along the approximated
gradients. An appropriate step length is identified automatically
based on the curvature of the objection function surface. The
Broyden-Fletcher-Goldfarb-Shanno (BFGS) \citep{Broyden70, Fletcher70,
  Goldfarb70, Shanno70} method is a quasi-Newton approach used for
standard optimization tasks in this thesis.

\subsection{Generalizability and overlearning}\label{sec:generalizability}

Probabilistic models are formulated in terms of probability
distributions over the sample space and parameter values. This forms
the basis for generalization to new, unobserved events. A
generalizable model can describe essential characteristics of the
underlying process that generated the observations; a generalizable
model is also able to characterize future observations. {\it
  Overlearning}, or {\it overfitting} refers to models that describe
the training data well, but do not generalize to new
observations. Such models describe not only the general processes
underlying the observations, but also noise in the particular
observations. Avoiding overfitting is a central aspect in modeling.
Overlearning is particularly likely when training data is scarce. While
overfitting could in principle be avoided by collecting more data,
this is often not feasible since the cost of data collection can be
prohibitively large.

Generalizability can be measured by investigating how accurately the
model describes new observations.  A standard approach is to split the
data into a {\it training set}, used to learn the model, and a {\it
  test set}, used to measure model performance on unseen observations
that were not used for training. In \emph{cross-validation} the test
is repeated with several different learning and test sets to assess
the variability in the testing procedure.  Cross-validation is used
for instance in Publication~\ref{ECML} of this thesis. \emph{Bootstrap
  analysis} \citep[see, for instance,][]{Efron94} is another widely
used approach to measure model performance. The observed data is
viewed as a finite realization of an underlying probability
density. New samples from the underlying density are obtained by
re-sampling the observed data points with replacement to simulate
variability in the original data; observations from the more dense
regions of the probability space become re-sampled more often than
rare events. Each bootstrap sample resembles the probability density
of the original data. Modeling multiple data sets obtained with the
bootstrap helps to estimate the sensitivity of the model to variations
in the data. Bootstrap is used to assess model performance in
Publication~\ref{AC}.

\subsection{Regularization and model selection}

In general, increasing model complexity will yield more flexible
models, which have higher descriptive power but are, on the other hand, more likely
to overfit. Therefore relatively simple models can often outperform
more complex models in terms of generalizability. A compromise between
simplicity and descriptive power can be obtained by imposing
additional constraints or soft penalties in the modeling to prefer
compact solutions, but at the same time retain the descriptive power
of the original, flexible model family.  This is called {\it
  regularization}. Regularization is particularly important when the
sample size is small, as demonstrated for instance in
Publication~\ref{MLSP}, where explicit and theoretically principled
regularization is achieved by setting appropriate priors on the model
structure and parameter values. The priors will then affect the MAP
estimate of the model parameters. One commonly used approach is to
prefer {\it sparse} solutions that allow only a small number of the
potential parameters to be employed at the same time to model the data
\citep[see e.g.][]{Archambeau08}. A family of probabilistic approaches
to balance between model fit and model complexity is provided by
information-theoretic criteria \citep[see e.g.][]{Gelman03}. The {\it
  Bayesian Information Criterion (BIC)} is a widely used information
criterion that introduces a penalty term on the number of model
parameters to prefer simpler models. The log-likelihood \(\L\) of the
data, given the model, is balanced by a measure of model complexity,
\(q log(N)\), in the final objective function \(- 2\L + q
log(N)\). Here \(q\) denotes the number of model parameters and \(N\)
is the constant sample size of the investigated data set. The BIC has
been criticized since it does not address changes in prior
distributions, and its derivation is based on asymptotic
considerations that hold only approximately with finite sample size
\citep[see e.g.][]{Bishop06}. On the other hand, BIC provides a
principled regularization procedure that is easy to implement. In this
thesis, the BIC has been used to regularize the algorithms in
Publication~\ref{NR}.

\subsection{Validation}

After learning a probabilistic model, it is necessary to confirm the
quality of the model and verify potential findings in further,
independent experiments. {\it Validation} refers to a versatile set of
approaches used to investigate model performance, as well as in model
criticism, comparison and selection.  Internal and external approaches
provide two complementary categories for model validation. {\it
  Internal validation} refers to procedures to assess model
performance based on training data alone. For instance, it is possible
to estimate the sensitivity of the model to initialization,
parameterization, and variations in the data, or convergence of the
learning process. Internal analysis can help to estimate the
weaknesses and generalizability of the model, and to compare
alternative models. Bootstrap and cross-validation are widely used
approaches for internal validation and the analysis of model
performance \citep[see e.g.][]{Bishop06}. Bootstrap can provide
information about the sensitivity of the results to sampling effects
in the data. Cross-validation provides information about the model
generalization performance and robustness by comparing predictions of
the model to real outcomes. {\it External validation} approaches
investigate model predictions and fit on new, independent data sets
and experiments. Exploratory analysis of high-throughput data sets
often includes massive multiple testing, and provides potentially
thousands of automatically generated hypotheses. Only a small set of
the initial findings can be investigated more closely by human
intervention and costly laboratory experiments. This highlights the
need to prioritize the results and assess the uncertainty in the
models.
\newpage

\chapter{Reducing uncertainty in high-throughput microarray studies}
\label{ch:preproc} 

\begin{quotation}
  \emph{As far as the laws of mathematics refer to reality, they are
    not certain, as far as they are certain, they do not refer to
    reality.}
\begin{flushright}
A. Einstein (1956)
\end{flushright}
\end{quotation}

Gene expression microarrays are currently the most widely used
technology for genome-wide transcriptional profiling, and they
constitute the main source of data in this thesis. An overview of
microarray technology is provided in
Section~\ref{sec:biodata}. Microarray measurements are associated with
high levels of noise from technical and biological
sources. Appropriate preprocessing techniques can help to reduce noise
and obtain reliable measurements, which is the crucial starting point
for any data analysis task. This chapter presents the first main
contribution of the thesis, preprocessing techniques that utilize side
information in genomic sequence databases and microarray data collections in order to improve the accuracy
of high-throughput gene expression data. The chapter is organized as follows: Section~\ref{sec:noise} gives an overview of the various sources of noise in high-throughput microarray studies. Section~\ref{sec:preprocessing} introduces a strategy for noise reduction based on side information in external genomic sequence databases. Section~\ref{sec:modelnoise} extends this model by describing a model-based approach that additionally combines statistical evidence across multiple microarray experiments in order to provide quantitative information of probe performance and utilizes this information to improve the reliability of high-throughput observations. The results are summarized in Section~\ref{sec:preconclusion}.

\section{Sources of uncertainty}\label{sec:noise}

Measurement data obtained with novel high-throughput technologies
comes with high levels of uncontrolled biological and technical
variation. This is often called {\it noise} as it obscures the
measurements, and adds potential bias and variance on the {\it signal}
of interest. Biological noise is associated with natural biological
variation between cell populations, cellular processes and
individuals.  Single-nucleotide polymorphisms, alternative splicing
and non-specific hybridization add biological variation in the data
\citep{Dai05,Zhang05}. More technical sources of noise in the
measurement process include RNA extraction and amplification,
experiment-specific variation, as well as platform- and
laboratory-specific effects \citep{Choi03, Shi06,Tu02}.

\begin{figure}[t]
\label{fig:badprobe}
\begin{center}
\includegraphics[width=\textwidth]{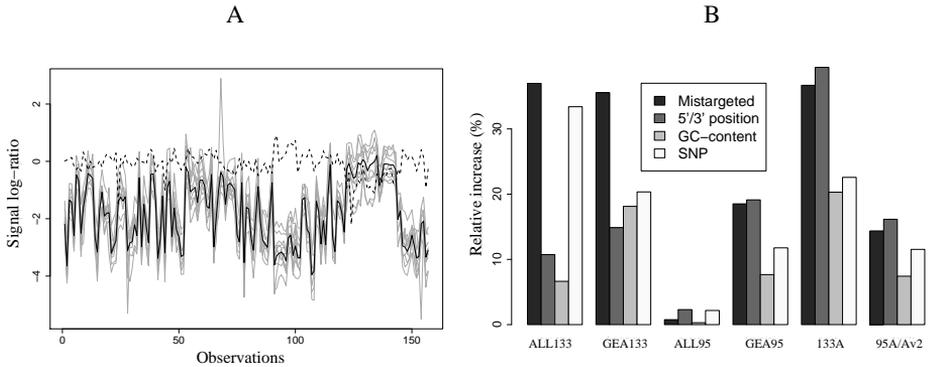}
\end{center}
\caption{{\bf A} Example of a probe set that contains a probe with
  high contamination levels (dashed line) detected by the
  probabilistic RPA model. The probe-level observations of
  differential gene expression for the different probes that measure
  the same target transcript are indicated by gray lines. The black
  line shows the estimated signal of the target transcript across a
  number of conditions. {\bf B} Increase in the average variance of
  the probes associated with the investigated noise sources:
  mistargeted probes having errors in the genomic alignment, most
  5'/3' probes of each probe set, GC-rich, and SNP-associated probes.
  The variances were estimated by RPA and describe the noise level of
  the probes. The results are shown for the individual ALL and GEA
  data sets, and for their combined results on both platforms (133A
  and 95A/Av2). \copyright IEEE. Reprinted with permission from
  Publication~\ref{RPA}.} 
\end{figure}

A significant source of noise on gene expression arrays comes from
individual probes that are designed to measure the activity of a given
transcript in a biological sample. Figure~\ref{fig:badprobe}A
shows probe-level observations of differential gene expression for a
collection of probes designed to target the same mRNA transcript. One
of the probes is highly contaminated and likely to add unrelated
variation to the analysis. A number of factors affect probe
performance.  For instance, it has been reported in
Publication~\ref{PECA} and elsewhere \citep{Hwang04,Mecham04b} that a
large portion of microarray probes may target unintended mRNA
sequences. Moreover, although the probes have been designed to
uniquely hybridize with their intended mRNA target, remarkable
cross-hybridization with the probes by single-nucleotide polymorphisms
\citep{Dai05, Sliwerska07} and other mRNAs with closely similar
sequences \citep{Zhang05} have been reported; high-affinity probes
with high GC-content may have higher likelihood of cross-hybridization
with nonspecific targets \citep{Mei03}.  Alternative splicing
\citep{Shi06} and mRNA degradation \citep{Auer03} may cause
differences between probes targeting different positions of the gene
sequence. Such effects will contribute to probe-level contamination in
a probe- and condition-specific manner. However, sources of
probe-level noise are still poorly understood \citep{Irizarry05,Li05}
despite their importance for expression analysis and probe design.

High levels of noise set specific challenges for analysis. Better
understanding of the technical aspects of the measurement process will
lead to improved analytical procedures and ultimately to more
accurate biological results \citep{Reimers2010}. Publication~\ref{RPA}
provides computational tools to investigate probe performance and the
relative contributions of the various sources of probe-level
contamination on short oligonucleotide arrays.

\section{Preprocessing microarray data with side
  information}\label{sec:preprocessing}

{\it Preprocessing} of the {\it raw data} obtained from the original
measurements can help to reduce noise and improve comparability
between microarray experiments. Preprocessing can be defined in terms
of statistical transformations on the raw data, and this is a central
part of data analysis in high-throughput studies. This section
outlines the standard preprocessing steps for short oligonucleotide
arrays, the main source of transcriptional profiling data in this
thesis. However, the general concepts also apply to other microarray
platforms \citep{Reimers2010}.

\subsubsection{Standard preprocessing steps}

A number of preprocessing techniques for short oligonucleotide arrays
have been introduced \citep{Irizarry06, Reimers2010}. The standard
preprocessing steps in microarray analysis include quality control,
background correction, normalization and summarization.

{\it Microarray quality control} is used to identify arrays with
remarkable experimental defects, and to remove them from subsequent
analysis. The typical tests consider RNA degradation levels and a
number of other summary statistics to guarantee that the array data is
of reasonable quality. The arrays that pass the microarray quality
control are preprocessed further. Each array typically has spatial
biases that vary smoothly across the array, arising from technical
factors in the experiment. {\it Background correction} is used to
detect and remove such spatial effects from the array data, and to
provide a uniform background signal, enhancing the comparability of
the probe-level observations between different parts of the
array. Moreover, background correction can estimate the general noise
level on the array; this helps to detect probes whose signal differs
significantly from the background noise. Robust multi-array averaging
(RMA) is one of the most widely used approaches for preprocessing
short oligonucleotide array data \citep{Irizarry03rma}.  The
background correction in RMA is based on a global model for probe
intensities. The observed intensity, \(Y\), is modeled as a sum of an
exponential signal component, \(S\) and Gaussian noise
\(B\). Background corrected data is then obtained as the expectation
\(\E_B(S|Y)\). While background correction makes the observations
comparable within array, {\it normalization} is used to improve the
comparability between arrays. Quantile normalization is a widely used
method that forces all arrays to follow the same empirical intensity
distribution \citep[see e.g.][]{Bolstad03}.  Quantile normalization
makes the measurements across different arrays comparable, assuming
that the overall distribution of mRNA concentration is approximately
the same in all cell populations. This has proven to be a feasible
assumption in transcriptional profiling studies. As always, there are
exceptions. For instance, human brain tissues have systematic
differences in gene expression compared to other organs. On short
oligonucleotide arrays, a number of probes target the same
transcript. In the final {\it summarization step}, the individual
probe-level observations of each target transcript are summarized into
a single summary estimate of transcript activity. Standard algorithmic
implementations are available for each preprocessing step.

\subsubsection{Probe-level preprocessing methods}

Differences in probe characteristics cause systematic differences in
probe performance.  The use of several probes for each target leads to
more robust estimates on transcript activity but it is clear that
probe quality may significantly affect the results of a microarray
study \citep{Irizarry03}. Widely used preprocessing algorithms utilize
probe-specific parameters to model probe-specific effects in the probe
summarization step. Some of the first and most well-known probe-level 
preprocessing algorithms include dChip/MBEI \citep{Li01mbei}, RMA \citep{Irizarry03rma}, and
gMOS \citep{Milo03}. Taking probe-level effects into account can
considerably improve the quality of a microarray study
\citep{Reimers2010}. Publications~\ref{PECA} and~\ref{RPA} incorporate
side information of the probes to preprocessing, and introduce
improved probe-level analysis methods for differential gene expression
studies.

In order to introduce probe-level preprocessing methods in more
detail, let us consider the probe summarization step of the RMA
algorithm \citep{Irizarry03rma}. RMA has a Gaussian model for probe
effects with probe-specific mean parameters and a shared variance
parameter for the probes.  The mean parameters characterize
probe-specific binding affinities that cause systematic differences in
the signal levels captured by each probe. Estimating the
probe-specific effects helps to remove this effect in the final
probeset-level summary of the probe-level observations. To briefly
outline the algorithm, let us consider a collection of probes (a {\it
  probeset}) that measure the expression level of the same target
transcript \(g\) in condition \(i\). The probe-level observations are
modeled as a sum of the true, underlying expression signal \(\gi\),
which is common to all probes, probe-specific binding affinity
\(\muj\), and Gaussian noise \(\epsilon\). A probe-level observation
for probe \(j\) in condition \(i\) is then modeled in RMA as

\begin{equation}\label{eq:rmamodel}
  \sij = \gi + \muj + \epsilon.
\end{equation} 

Measurements from multiple conditions are needed to estimate the
probe-specific effects \(\muj\). RMA and other models that measure
absolute gene expression have an important drawback: the probe
affinity effects \(\{\muj\}\) are unidentifiable. In order to obtain
an identifiable model, the RMA algorithm includes an additional
constraint that the probe affinity effects are zero on average:
\(\Sigma_j\muj = 0\). This yields a well-defined algorithm that has
been shown to produce accurate measurements of gene expression in
practical settings. Further extensions of the RMA algorithm include
gcRMA, which has a more detailed chemical model for the probe effects
\citep{Wu04}, refRMA \citep{Katz06}, which utilizes probe-specific
effects derived from background data collections, and fRMA
\citep{McCall2010}, which also models batch-specific effects in
microarray studies. The estimation of unidentifiable probe affinities
is a main challenge for most probe-level preprocessing models.

RMA and other probe-level models for short oligonucleotide arrays have
been designed to estimate absolute expression levels of the
genes. However, gene expression studies are often ultimately targeted
at investigating {\it differential expression levels}, that is,
differences in gene expression between experimental conditions.
Measurements of differential expression is obtained for instance by
comparing the expression levels, obtained through the RMA algorithm or
other methods, between different conditions.  However, the
summarization of the probe-level values is then performed prior to the
actual comparison. Due to the unidentifiability of the probe affinity
parameters in the RMA and other probe-level models, this is
potentially suboptimal. Publication~\ref{PECA} demonstrates that
reversing the order, i.e., calculating differential gene expression
already at the probe level before probeset-level summarization, leads
to improved estimates of differential gene expression. The explanation
is that the procedure circumvents the need to estimate the
unidentifiable probe affinity parameters. This is formally described
in Publication~\ref{RPA}, which provides a probabilistic extension of
the Probe-level Expression Change Averaging (PECA) procedure of Publication~\ref{PECA}. In PECA, a standard
weighted average statistics summarizes the probe level observations of
differential gene expression. PECA does not model probe-specific
effects, but it is shown to outperform widely used probe-level
preprocessing methods, such as the RMA, in estimating differential
expression. Publication~\ref{RPA}, considered in more detail in
Section~\ref{sec:modelnoise}, provides an extended probabilistic
framework that also models probe-specific effects.

\subsubsection{Utilizing side information in transcriptome databases}

Probe-level preprocessing models and microarray analysis can be
further improved by utilizing external information of the probes
\citep{Eisenstein06, Hwang04, Katz06}.  Although any given microarray
is designed on most up-to-date sequence information available, rapidly
evolving genomic sequence data can reveal inaccuracies in probe
annotations when the body of knowledge grows.  In recent studies,
including Publication~\ref{PECA}, a remarkable number of probes on
various oligonucleotide arrays have been detected not to uniquely
match their intended target \citep{Hwang04, Mecham04a}.  A remarkable
portion of probes on several popular microarray platforms in human and
mouse did not match with their intended mRNA target, or were found to
target unintended mRNA transcripts in the Entrez Nucleotide
\citep{Wheeler05} sequence database in Publication~\ref{PECA}
(Table~\ref{tab:verified}). The observations are in general concordant
with other studies, although the exact figures vary according to the
utilized database and comparison details \citep{Gautier04b,
  Mecham04b}. In this thesis, strategies are developed to improve
microarray analysis with background information from genomic sequence
databases, and with model-based analysis of microarray collections.

Probe verification is increasingly used in standard preprocessing, and
to confirm the results of a microarray study.  Matching the probe
sequences of a given array to updated genomic sequence databases and
constructing an alternative interpretation of the array data based on
the most up-to-date genomic annotations has been shown to increase the
accuracy and cross-platform consistency of microarray analyses in
Publication~\ref{PECA} and elsewhere \citep{Dai05, Gautier04b}.

\begin{figure}[ht]
\label{fig:rpaReproducibility}
\centering
\begin{tabular}{cc}
{\bf A}&{\bf B}\\
\rotatebox{0}{\includegraphics[width=.48\textwidth]{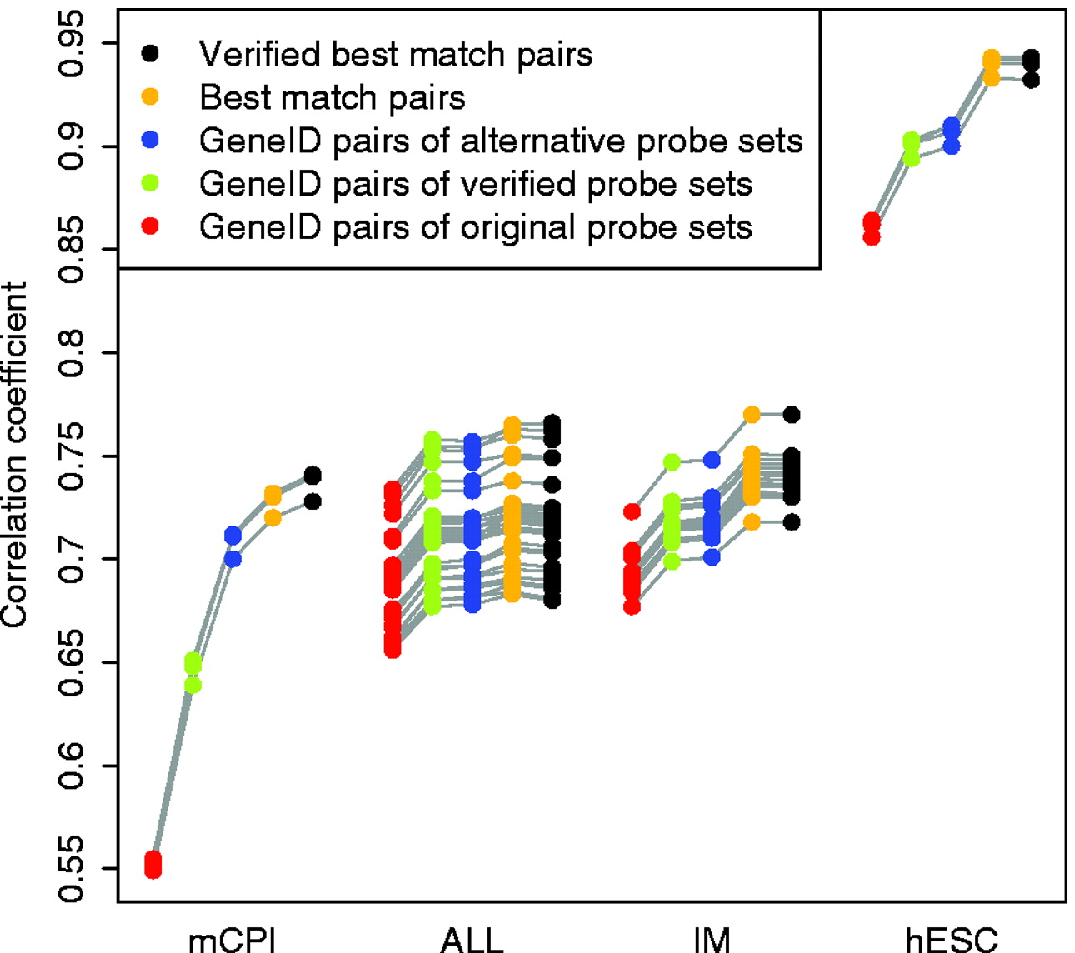}}&
\includegraphics[width=.48\textwidth]{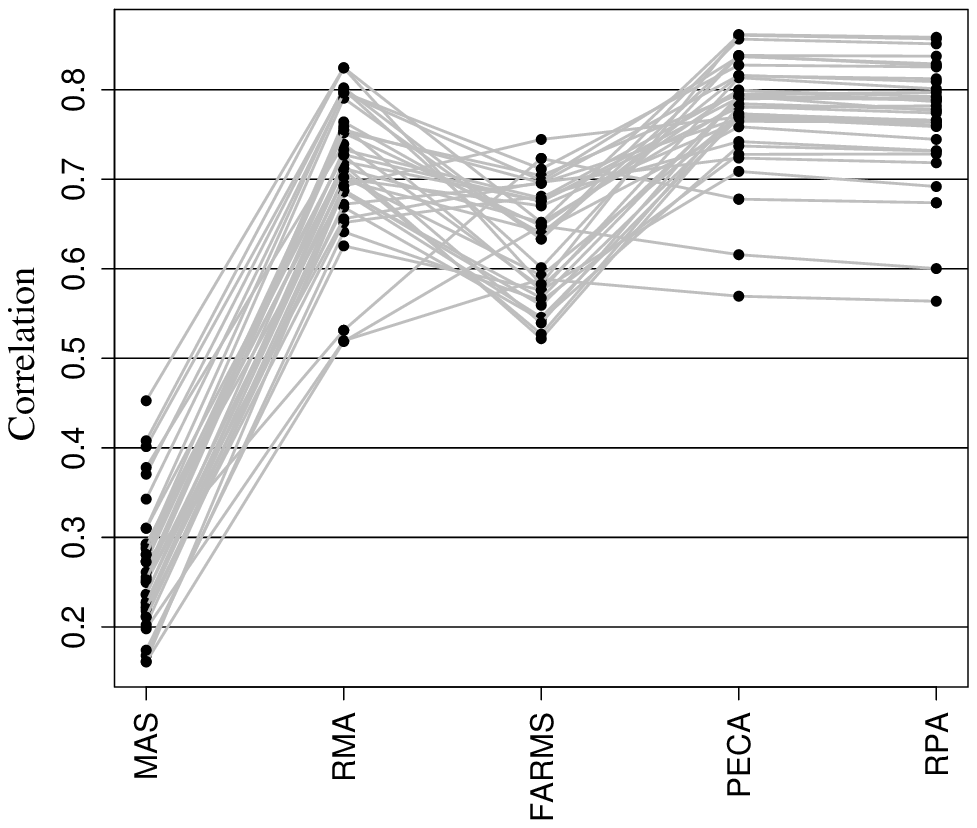}
\end{tabular}
\caption{{\bf A} Effect of sequence verification on comparability
  between microarray platforms. Correlations between RMA-preprocessed
  technical replicates on two array platforms where the same samples
  have been hybridized on the two array types. The Pearson
  correlations were calculated for each pair of arrays measuring the
  same biological sample. The gray lines show correlations obtained
  with the different probe matching criteria. In the hESC array
  comparison, the best match probe sets contained exactly the same
  probes on both array generations, which resulted in very high
  correlations. The advantages of probe verification and alternative
  mappings were largest when arrays with different probe collections
  were compared in the mCPI, ALL and IM array comparisons. {\bf B}
  Reproducibility of signal estimates in real data sets between the
  technical replicates, i.e., the 'best match' probe sets between the
  HG-U95Av2 and HG-U133A platforms.  The consistency was measured by
  the Pearson correlation between the pairs of arrays, to which the
  same sample was hybridized. \copyright Published by Oxford
  University Press. Reprinted with permission from
  Publication~\ref{PECA}.}
\end{figure}

Publication~\ref{PECA} combines probe verification with a novel
probe-level preprocessing method, PECA, to suggest a novel framework
for comparing and combining results across different microarray
platforms. While huge repositories of microarray data are available,
the data for any particular experimental condition is typically
scarce, and coming from a number of different microarray
platforms. Therefore reliable approaches for integrating microarray
data are valuable. Integration of results across platforms has proven
problematic due to various sources of technical variation between
array technologies.  Matching of probe sequences between microarray
platforms has been shown to increase the consistency of microarray
measurements \citep{Hwang04,Mecham04b}. However, probe matching
between array platforms guarantees only technical comparability
\citep{Irizarry05}.  Probe verification against external sequence
databases is needed to confirm that the probes are also biologically
accurate. This can also improve the comparability across array
platforms, as confirmed by the validation studies in
Publication~\ref{PECA} (Figure~\ref{fig:rpaReproducibility}A).

The PECA method of Publication~\ref{PECA} utilizes genomic sequence
databases to reduce probe-level noise by removing erroneous probes
based on updated genomic knowledge. The strategy relies on external
information in the databases and can therefore only remove known
sources of probe-level contamination.  Publication~\ref{RPA}
introduces a probabilistic framework to measure probe reliability
directly based on microarray data collections. The analysis can reveal
both well-characterized and unknown sources of probe-level
contamination, and leads to improved estimates of gene expression.
This model, coined Robust Probabilistic Averaging (RPA), also provides
a theoretically justified framework for incorporating prior knowledge
of the probes into the analysis.

\begin{center}
\begin{table}[htb]
\label{tab:verified}
\begin{tabular}{lcc}
Array type  	&Number of probes  	&Verified probes (\%)\\\hline
HG-U133 Plus2.0 &604,258 	&58.2\\
HG-U133A 	&247,965 	&82.5\\	
HG-U95Av2 	&199,084 	&82.6\\	
MOE430 2.0 	&496,468 	&68.2\\	
MG-U74Av2 	&197,993 	&73.1 	
\end{tabular}
\caption{The proportion of sequence-verified probes on three popular human
  microarray platforms and two mouse platforms, as observed in
  Publication~\ref{PECA}. Probes that matched to mRNA sequences corresponding to unique genes (defined by a GeneID identifier) in the Entrez database are considered verified. A remarkable portion of the probes on the investigated arrays did not match the Entrez transcript sequences, or had ambiguous targets.}
\end{table}
\end{center}

\section{Model-based noise reduction}\label{sec:modelnoise}

Standard approaches for investigating probe performance typically rely
on external information, such as genomic sequence data (see
\citealt{Mecham04b, Zhang05} and Publication~\ref{PECA}) or physical
models \citep{Naef03,Wu05}. However, such models cannot reveal probes
with uncharacterized sources of contamination, such as
cross-hybridization with alternatively spliced transcripts or closely
related mRNA sequences.  Vast collections of microarray data are
available in public repositories. These large-scale data sets contain
valuable information of both biological and technical aspects of gene
expression studies. Publication~\ref{RPA} introduces a data-driven
strategy to extract and utilize probe-level information in microarray
data collections.

The model, {\it Robust Probabilistic Averaging (RPA)}, is a
probabilistic preprocessing procedure that is based on explicit
modeling assumptions to analyze probe reliability and quantify the
uncertainty in measurement data based on gene expression data
collections, independently of external information of the probes. The
model can be viewed as a probabilistic extension of the probe-level
preprocessing approach for differential gene expression studies
presented in Publication~\ref{PECA}.  The explicit Bayesian
formulation quantifies the uncertainty in the model parameters, and
allows the incorporation of prior information concerning probe
reliability into the analysis. RPA provides estimates of probe
reliability, and a probeset-level estimate of differential gene
expression directly from expression data and independently of the
noise source. The RPA model is independent of physical models or
external and constantly updated information such as genomic sequence
data, but provides a framework for incorporating such prior
information of the probes in gene expression analysis.

Other probabilistic methods for microarray preprocessing include BGX
\citep{Hein05}, gMOS \citep{Milo03} and its extensions
\citep{Liu05}. The key difference to the RPA procedure of
Publication~\ref{RPA} is that these methods are designed to provide
probeset-level summaries of absolute gene expression levels, and
suffer from the same unidentifiability problem of probe affinity
parameters as the RMA algorithm \citep{Irizarry03rma}. In contrast,
RPA models probe-level estimates of differential gene expression. This
removes the unidentifiability issue, which is advantageous when the
objective is to compare gene expression levels between experimental
conditions. Another important difference is that the other
preprocessing methods do not provide explicit estimates of
probe-specific parameters, or tools to investigate probe performance.
Publication~\ref{RPA} assigns an explicit probabilistic measure of
reliability to each probe. This gives tools to analyze probe
performance and to guide probe design.

\subsubsection{Robust Probabilistic Averaging}

Let us now consider in more detail the probabilistic preprocessing
framework, RPA, introduced in Publication~\ref{RPA}. Probe performance
is ultimately determined by its ability to accurately measure the
expression level of the target transcript, which is unknown in
practical situations.  Although the performance of individual probes
varies, the collection of probes designed to measure the same
transcript will provide ground truth for assessing probe performance
(Figure~\ref{fig:badprobe}A). RPA captures the shared signal of the
probes within a probeset, and assumes that the shared signal
characterizes the expression of the common target transcript of the
probes. The reliability of individual probes is estimated with respect
to the strongest shared signal of the probes. RPA assumes normally
distributed probe effects, and quantifies probe reliability based on
probe variance around the probeset-level signal across a large number
of arrays. This extends the formulation of the RMA model in
Equation~\ref{eq:rmamodel} by introducing an additional probe-specific
Gaussian noise component:

\begin{equation}\label{eq:rpa}
  \sij = \gi + \muj + \epsilonij.
\end{equation}

\noindent In contrast to RMA, the variance is probe-specific in this
model, and distributed as \(\epsilonij \sim N(0,\taujSq)\). The
variance parameters \(\taus\) are of interest in probe reliability
analysis; they reflect the noise level of the probe, in contrast to
probe-level preprocessing methods that focus on estimating the
unidentifiable mean parameter of the Gaussian noise model,
corresponding to probe affinity \citep[see e.g.][]{Irizarry03rma,
  Li01mbei}. In Publication~\ref{RPA}, probe-level calculation of
differential expression avoids the need to model unidentifiable probe
affinities, the key probe-specific parameter in other probe-level
preprocessing methods. More formally, the unidentifiable probe
affinity parameters \(\mu_.\) cancel out in RPA when the signal
log-ratio between a user-specified 'reference' array and the remaining
arrays is computed for each probe: the differential expression signal
between arrays \(t = \{1, \dots, T\} \) and the reference array \(c\)
for probe \(j\) is obtained by \(\mtj = \stj - \scj = \gt - \gc +
\epsilontj - \epsiloncj = \dt + \epsilontj - \epsiloncj\). In vector
notation, the differential expression profile of probe \(j\) across
the \(T\) arrays is then written as \(\mj = \d + \Epsilonj\), i.e.,
a noisy observation of the true underlying differential expression
signal \(\d\) and probe-specific noise \(\Epsilonj\).

The unidentifiable probe affinity parameters cancel out in the RPA
model of Publication~\ref{RPA}. This can partly explain the previous
empirical observations that calculating differential expression
already at probe-level improves the analysis of differential gene
expression \citep{Zhang02, Elo05}. However, the previous models are
non-probabilistic preprocessing methods that do not aim at quantifying
the uncertainty in the probes. Use of a single parameter for probe
effects in RPA also gives more straightforward interpretations of
probe reliability.

Posterior estimates of the model parameters are derived to estimate
probe reliability and differential gene expression. The differential
expression vector \(\d = \{d_t\}\) and the probe-specific variances
\(\TauSq = \{\taujSq\}\) are estimated simultaneously.  The posterior
density of the model parameters is obtained from the likelihood of the
data and the prior according to Bayes' rule
(Equation~\ref{eq:bayesrule}) as

\begin{equation}\label{eq:rpapost1}
    p(\d, \TauSq | \m) \sim p(\m | \d, \TauSq) p(\d, \TauSq).
\end{equation}

\noindent To obtain this posterior, let us consider the likelihood
\(p(\m | \d, \TauSq)\) of the data and the prior \(p(\d, \TauSq)\) of the
model parameters. The noise on the selected control array
\(\epsiloncj\) is a latent variable, and marginalized out in the model
to obtain the likelihood:

\begin{equation}
\begin{split}
p(\m | \d, \TauSq) 
= \prod_{tj} \int N(\mtj | \dt - \epsiloncj, \taujSq) N(\epsiloncj | 0, \taujSq) d\epsiloncj \\
\sim \prod_j (2 \pi \taujSq)^{-\frac{T}{2}} exp(- \frac{\sum_t (\mtj -\dt)^2 - \frac{ [\sum_t (\mtj - \dt)]^2} {T + 1}}{2 \taujSq}).
\end{split}
\label{eq:datalikelihood}
\end{equation}

\noindent Let us assume independent priors, \(p(\d,\TauSq) = p(\d)
p(\TauSq)\), flat non-informative prior \(p(\d) \sim 1\) and
conjugate priors for the variance parameters in \(\TauSq\) (inverse
Gamma function, see \citealt{Gelman03}). With these standard
assumptions, the prior takes the form

\begin{equation}\label{eq:rpaprior}
  p(\d,\TauSq) \sim \prod_j IG(\taujSq; \alphaj, \betaj),
\end{equation}

\noindent where \(\alphaj\) and \(\betaj\) are the shape and scale parameters of
the inverse Gamma distribution. Prior information of the probes can be
incorporated in the analysis through these parameters. Probe-level
differential expression is then described by two sets of parameters;
the differential gene expression vector \(\d = [d_1 \dots d_T]\), and
the probe-specific variances \(\TauSq = [\tau^2_1 \dots \tau^2_J]\).
High variance \(\taujSq\) indicates that the probe-level observation
\(\mj\) is strongly deviated from the estimated true signal
\(\d\). Denoting \(\alphahatj = \alphaj + \frac{T}{2}\) and
\(\betahatj = \betaj + \frac{1}{2}\sum_t (\mtj - \dt)^2 -
\frac{1}{2}\frac{(\sum_t (\mtj - \dt))^2}{T+1}\), the posterior of the
model parameters in Equation~\ref{eq:rpapost1} takes the form

\begin{equation}
\label{eq:fullpost}
p(\d, \TauSq | \m) \sim \prod_j (\taujSq)^{-(\alphahatj + 1)} exp(-\frac{\betahatj}{\taujSq}).
\end{equation}

\noindent The formulation allows estimating the uncertainty in the
expression estimates and probe-level parameters.  In practice, a MAP
point estimate of the parameters, obtained by maximizing the
posterior, is often sufficient. In the limit of a large sample size
(\(T \rightarrow \infty\)), the model will converge to estimating
ordinary mean and variance parameters. With limited sample sizes that
are typical in microarray studies the prior parameters provide
regularization that makes the probabilistic formulation more robust to
overfitting and local optima, compared to direct estimation of the
mean and variance parameters. Moreover, the probabilistic analysis
takes the uncertainty in the data and model parameters into account in
an explicit manner.

The model also provides a principled framework for incorporating prior
knowledge probe reliability in microarray preprocessing through the
probe-specific hyperparameters \(\alpha, \beta\). Estimation and use
of probe-specific effects from external microarray data collections
has been previously suggested in the context of the refRMA method by
\cite{Katz06}, where such side information was shown to improve gene
expression estimates. The RPA method of Publication~\ref{RPA} provides
an alternative probabilistic treatment.

\subsubsection{Model validation}

The probabilistic RPA model introduced in Publication~\ref{RPA} was
validated by comparing the preprocessing performance to other
preprocessing methods, and additionally by comparing the estimates of
probe-level noise to known sources of probe-level contamination.  The
comparison methods include the FARMS \citep{Hochreiter06}, MAS5
\citep{Hubbell2002}, PECA (Publication~\ref{PECA}), and RMA
\citep{Irizarry03rma} preprocessing algorithms. FARMS has a more
detailed model for probe effects than the other methods, and it
contains implicitly a similar probe-specific variance parameter than
our RPA model. FARMS is based on a factor analysis model, and is
defined as \(\sij = z_i \lambdaj + \muj + \epsilonij\), where \(z_i\)
captures the underlying gene expression. In contrast to RMA and RPA
that have a single probe-specific parameter, FARMS has three
probe-specific parameters \(\{\lambdaj, \muj, \epsilonij\}\). MAS5 is
a standard preprocessing algorithm provided by the array
manufacturer. The algorithm performs local background correction,
utilizes so-called mismatch probes to control for non-specific
hybridization, and scales the data from each array to the same average
intensity level to improve comparability across arrays. MAS5
summarizes probe-level observations of absolute gene expression levels
using robust summary statistics, Tukey biweight estimate, but unlike
FARMS, RMA and RPA, MAS5 does not model probe-specific effects.

The preprocessing performance of these methods was investigated in
spike-in experiments where certain target transcripts measured by the
array have been spiked in at known concentrations, as well as on real
data sets.  The results from the spike-in experiments were compared in
terms of receiver operating characteristics (ROC).  The standard RMA,
PECA (Publication~\ref{PECA}) and RPA (Publication~\ref{RPA}) had
comparable performance in spike-in data, and they outperformed the
MAS5 \citep{Hubbell2002} and FARMS \citep{Hochreiter06} preprocessing
algorithms in estimating differential gene expression. On real data
sets, PECA and RPA outperformed the other methods, providing higher
reproducibility between technical replicates measured on different
microarray platforms (Figure~\ref{fig:rpaReproducibility}B).

In contrast to standard preprocessing algorithms, RPA provides
explicit quantitative estimates of probe performance. The model has
been validated on widely used human whole-genome arrays by comparing
the estimates of probe reliability with known probe-level error
sources: errors in probe-genome alignment, interrogation position of a
probe on the target sequence, GC-content, and the presence of SNPs in
the probe target sequences; a good model for assessing probe
reliability should detect probes contaminated by the known error
sources.  The results from our analysis can be used to characterize
the relative contribution of different sources of probe-level noise
(Figure~\ref{fig:badprobe}B). In general, the probes with known sources
of contamination were more noisy than the other probes, with 7-39\%
increase in the average variance, as detected by RPA.  Any single
source of error seems to explain only a fraction of the most highly
contaminated probes. A large portion (35-60\%) of the detected least
reliable probes were not associated with the investigated known noise
sources. This suggests that previous methods that remove probe-level
noise based on external information, such as genomic alignments will
fail to detect a significant portion of poorly performing probes.  The
RPA model of Publication~\ref{RPA} provides rigorous algorithmic tools
to investigate the various probe-level error sources. Better
understanding of the factors affecting probe performance can advance
probe design and contribute to reducing probe-related noise in future
generations of gene expression arrays.

\section{Conclusion}\label{sec:preconclusion}

The contributions presented in this Chapter provide improved
preprocessing strategies for differential gene expression studies. The
introduced techniques utilize probe-level analysis, as well as side
information in sequence and microarray databases.  Probe-level studies
have led to the establishment of probe verification and alternative
microarray interpretations as a standard step in microarray
preprocessing and analysis. The alternative interpretations for
microarray data based on updated genomic sequence data
\citep{Gautier04b, Dai05} are now implemented as routine tools in
popular preprocessing algorithms such as the RMA, or the RPA method of
Publication~\ref{RPA}. The probe-level analysis strategy has been
recently extended to exon array context, where expression levels of
alternative splice variants of the same genes are compared under
particular experimental conditions. The probe-level approach has shown
superior preprocessing performance also with exon arrays
\citep{Laajala2009}.  A convenient access to the algorithmic tools
developed in Publications~\ref{PECA} and~\ref{RPA} for microarray
preprocessing and probe-level analysis is provided by the accompanied
open source implementation in
BioConductor.\footnote{http://www.bioconductor.org/packages/release/bioc/html/RPA.html}

\newpage

\chapter{Global analysis of the human transcriptome}\label{ch:atlases}

\begin{quotation}
\emph{When we try to pick out anything by itself, we find that it is bound fast by a thousand invisible cords that cannot be broken, to everything in the universe.}
\begin{flushright}
J. Muir (1869)
\end{flushright}
\end{quotation}

Measurements of transcriptional activity provide only a partial view
to physiological processes, but their wide availability provides a
unique resource for investigating gene activity at a genome- and
organism-wide scale. Versatile and carefully controlled {\it gene
expression atlases} have become available for normal human tissues,
cancer as well as for other diseases \citep[see, for
instance,][]{Kilpinen08, Lukk10, Roth06, Su04}. These data sources
contain valuable information about shared and unique mechanisms
between disparate conditions, which is not available in smaller and
more specific experiments \citep{Lage08, Scherf2000}.  While standard
methods for gene expression analysis have focused on comparisons
between particular conditions, versatile transcriptome atlases allow
for global organism-wide characterization of transcriptional
activation patterns \citep{Levine06}. Novel methodological approaches
are needed in order to realize the full potential of these information
sources, as many traditional methods for expression analysis are not
applicable to versatile large-scale collections. This chapter provides
an overview to current approaches for global transcriptome analysis in
Section~\ref{sec:standard} and introduces the second main contribution
of the thesis, a novel exploratory approach that can be used to
investigate context-specific responses in genome-scale interaction
networks across organism-wide collections of measurement data in
Section~\ref{sec:atlas}. The conclusions are summarized in
Section~\ref{sec:atlasconclusion}.

\section{Standard approaches}\label{sec:standard}

Global observations of transcriptional activity reflect known and
previously uncharacterized cell-biological processes. Exploratory
analysis of the transcriptome can provide research hypotheses and
material for more detailed investigations. Widely-used standard
approaches for global transcriptome analysis include various
clustering, dimensionality reduction and visualization techniques
\citep[see e.g.][]{Huttenhower2010, Polanski07, Quackenbush01}.  The
large data collections open up new possibilities to investigate
functional relatedness between physiological conditions, disease
states, as well as cellular processes, and to discover previously
uncharacterized connections and functional mechanisms
\citep{Bergmann04, Kilpinen08, Lukk10}.

Gene expression studies have traditionally focused on the analysis of
relatively small and targeted data sets, such as particular diseases
or cell types. A typical objective is to detect genes, or gene groups,
that are differentially expressed between particular conditions, for
instance to predict disease outcomes, or to identify potentially
unknown disease subtypes.  The increasing availability of large and
versatile transcriptome collections that may cover thousands of
experimental conditions allows global, data-driven analysis, and the
formulation of novel research questions where the traditional analysis
methods are often insufficient \citep{Huttenhower2010}.  

A variety of approaches have been proposed and investigated in the
recent years in the global transcriptome analysis context. An actively
studied modeling problem in transcriptome analysis is the {\it
  discovery of transcriptional modules}, i.e., identification of
coherent gene groups that show coordinated transcriptional responses
under particular conditions \citep{Segal03psb, Segal04nature,
  Stuart03}. Models have also been proposed to predict gene regulators
\citep{Segal03nature}, and to infer cellular processes and networks
based on transcriptional activation patterns \citep{Friedman04,
  Segal03a}. An increasing number of models are being developed to
integrate transcriptome measurements to other sources of genomic
information, such as regulation and interactions between the genes to
detect and characterize cellular processes and disease mechanisms
\citep{Barash02, Chari2010a, Vaske2010}.  Findings from transcriptome
analysis have potential biomedical implications, as in \cite{Lamb06},
where chemically perturbed cancer cell lines were screened to enhance
the detection of drug targets based on shared functional mechanisms
between disparate conditions, or in \cite{Sorlie01}, where cluster
analysis of cancer patients based on genome-wide transcriptional
profiling experiments led to the discovery of a novel breast cancer
subtype. In the remainder of this section, the modeling approaches
that are particularly closely related to the contributions of this
thesis are considered in more detail.

\subsubsection{Investigating known processes}

A popular strategy for genome-wide gene expression analysis is to
consider known biological processes and their activation patterns
across diverse collections measurement data from various experimental
conditions.  Biomedical databases contain a variety of information
concerning genes and their interactions. For instance, the Gene
Ontology database \citep{Ashburner00} provides functional and
molecular classifications for the genes in human and a number of other
organisms. Other categories are based on micro-RNA regulation,
chromosomal locations, chemical perturbations and other features
\citep{Subramanian05}. Joint analysis of functionally related genes
can increase the statistical power of the analysis.  So-called {\it
  gene set-based approaches} are typically designed to test
differential expression between two particular conditions
\citep{Goeman07, Nam08}, but they can also be used to build global
maps of transcriptional activity of the known processes
\citep{Levine06}. However, gene set-based approaches typically ignore
more detailed information of the interactions between individual
genes. Pathway and interaction databases contain more detailed
information concerning molecular interactions and cell-biological
processes \citep{Kanehisa08, Vastrik07}.  {\it Network-based methods}
utilize relational information of the genes to guide expression
analysis. For instance, \cite{Draghici07} demonstrated that taking
into account aspects of pathway topology, such as gene and interaction
types, can improve the estimation of pathway activity between two
predefined conditions. Another recent approach which utilizes pathway
topology in inferring pathway activity is PARADIGM \citep{Vaske2010},
which also integrates other sources of genomic information in pathway
analysis. However, these methods have been designed for the analysis
of particular experimental conditions, rather than comprehensive
expression atlases. MATISSE \citep{Ulitsky07} is a network-based
approach that searches for functionally related genes that are
connected in the network, and have correlated expression profiles
across many conditions. The potential shortcoming of this approach is
that it assumes global correlation across all conditions between the
interacting genes, while many genes can have multiple,
context-sensitive functional roles.  Different conditions induce
different responses in the same genes, and the definition of 'gene
set' is vague \citep{Montaner09, Nacu07}. Therefore methods have been
suggested to identify 'key condition-responsive genes' of predefined
gene sets \citep{Lee08c}, or to decompose predefined pathways into
smaller and more specific functional modules \citep{Chang09}. These
approaches rely on predefined functional classifications for the
genes. The data-driven analysis in Publication~\ref{NR} provides a
complementary approach where the gene sets are learned directly from
the data, guided by prior knowledge of genetic interactions. This
avoids the need to refine suboptimal annotations, and enables the
discovery of new processes. The findings demonstrate that simply
measuring whether a gene set, or a network, is differentially
expressed between particular conditions is often not sufficient for
measuring the activity of cell-biological processes. Since gene
function and interactions are regulated in a context-specific manner,
it is important to additionally characterize how, and in which
conditions the expression changes. Global analysis of transcriptional
activation patterns interaction networks, introduced in
Publication~\ref{NR}, can address such questions.

\subsubsection{Biclustering and subspace clustering}

Approaches that are based on previously characterized genes and
processes are biased towards well-characterized phenomena. This limits
their value in {\it de novo} discovery of functional patterns.
Unsupervised methods provide tools for such analysis, but often with
an increased computational cost and a higher proportion of false
positive findings.

{\it Cluster analysis} is widely used for unsupervised analysis of
gene expression data, providing tools for class discovery, gene
function prediction and for visualization purposes. Examples of widely
used clustering approaches include hierarchical clustering and K-means
\citep[see e.g.][]{Polanski07}. Clustering of patient samples with
similar expression profiles has led to the discovery of novel cancer
subtypes with biomedical implications \citep{Sorlie01}; clustering of
genes with coordinated activation patterns can be used, for instance,
to predict novel functional associations for poorly characterized
genes \citep{Allocco04}. The self-organizing map \citep{Kohonen82,
  Kohonen01} is a related approach that provides efficient tools to
{\it visualize} high-dimensional data on lower-dimensional displays,
with particular applications in transcriptional profiling studies
\citep{Tamayo99, Toronen99}.  The standard clustering methods are
based on comparison of global expression patterns, and therefore are
relatively coarse tools for analyzing large transcriptome collections.
Different genes respond in different ways, as well as in different
conditions.  Therefore it is problematic to find clusters in
high-dimensional data spaces, such as in whole-genome expression
profiling studies; different gene groups can reveal different
relationships between the samples. Detection of smaller, coherent
subspaces with a particular structure can be useful in biomedical applications, 
where the objective is to identify sets of interesting genes for further analysis. 
Both genes and the associated conditions may be unknown, and
the learning task is to detect them from the data. This can help, for
instance, in identifying responses to drug treatments in particular genes
\citep{Ihmels02, Tanay02bioinf}, or in identifying functionally coherent
transcriptional modules in gene expression databases
\citep{Segal04nature, Tanay05msb}.

{\it Subspace clustering} methods \citep{Parsons04} provide a family
of algorithms that can be used to identify subsets of dependent
features revealing coherent clustering for the samples; this defines a
subspace in the original feature space. Subspace clustering models are
a special case of a more general family of {\it biclustering}
algorithms \citep{Madeira04}. Closely related models are also called
co-clustering \citep{Cho04}, two-way clustering \citet{Getz00}, and
plaid models \citep{Lazzeroni02}. Biclustering methods provide general
tools to detect co-regulated gene groups and associated conditions
from the data, to provide compact summaries and to aid interpretation
of transcriptome data collections. Biclustering models enable the
discovery of {\it gene expression signatures} \citep{Hu06} that have
emerged as a central concept in global expression analysis context. A
signature describes a co-expression state of the genes, associated
with particular conditions. Established signatures have been found to
be reliable indicators of the physiological state of a cell, and
commercial signatures have become available for routine clinical
practice \citep{Nuyten08}. However, the established signatures are
typically designed to provide optimal classification performance
between two particular conditions. The problem with the
classification-based signatures is that their associations to the
underlying physiological processes are not well understood
\citep{Lucas09}. In Publication~\ref{NR} the understanding is enhanced
by deriving transcriptional signatures that are explicitly connected
to well-characterized processes through the network.

\subsubsection{Role of side information}

Standard clustering models ignore prior information of the data, which
could be used to supervise the analysis, to connect the findings to
known processes, as well as to improve scalability. For instance,
standard model-based feature selection, or subspace clustering
techniques would consider all potential connections between the genes
or features \citep{Law04, Roth04}. Without additional constraints on
the solution space they can typically handle at most tens or hundreds
of features, which is often insufficient in high-throughput genomics
applications.  Use of side information in clustering can help to guide
unsupervised analysis, for instance based on known or potential
interactions between the genes. This has been shown to improve the
detection of functionally coherent gene groups \citep{Hanisch02,
Shiga07, Ulitsky07, Zhu05}. However, while these methods provide tools
to cluster the genes, they do not model differences between
conditions. Extensions of biclustering models that can utilize
relational information of the genes include cMonkey \citep{Reiss06}
and a modified version of SAMBA biclustering \citep{Tanay04}. However,
cMonkey and SAMBA are application-oriented tools that rely on
additional, organism-specific information, and their implementation is
currently not available for most organisms, including that of the
human.  Further application-oriented models for utilizing side
information in the discovery of transcriptional modules have recently
been proposed for instance by \cite{Savage2010} and \cite{Suthram10}.
Publication~\ref{NR} introduces a complementary method where the
exhaustively large search space is limited with side information
concerning known relations between the genes, derived from genomic
interaction databases. This is a general algorithmic approach whose
applicability is not limited to particular organisms.

\subsubsection{Other approaches}

Prior information on the cellular networks, regulatory mechanisms, and
gene function is often available, and can help to construct more
detailed models of gene function and network analysis, as well as to
summarize functional aspects of genomic data collections
\citep{Huttenhower09, Segal03nature, Troyanskaya05}. Versatile
transcriptome collections also enable {\it network reconstruction},
i.e., {\it de novo} discovery \citep{Lezon06, Myers05} and
augmentation \citep{Novak06} of genetic interaction networks. Other
methodological approaches for global transcriptome analysis are
provided by probabilistic latent variable models \citep{Rogers05,
  Segal03psb}, hierarchical Dirichlet process algorithms
\citep{Gerber07}, as well as matrix and tensor computations
\citep{Alter05}. These methods provide further model-based tools to
identify and characterize transcriptional programs by decomposing gene
expression data sets into smaller, functionally coherent components.

\section{Global modeling of transcriptional activity in interaction
  networks}\label{sec:atlas}

Molecular interaction networks cover thousands of genes, proteins and
small molecules. Coordinated regulation of gene function through
molecular interactions determines cell function, and is reflected in
transcriptional activity of the genes.  Since individual processes and
their transcriptional responses are in general unknown \citep{Lee08c,
  Montaner09}, data-driven detection of condition-specific responses
can provide an efficient proxy for identifying distinct
transcriptional states of the network with potentially distinct
functional roles. While a number of methods have been proposed to
compare network activation patterns between particular conditions
\citep{Draghici07, Ideker02, Cabusora05, Noirel08}, or to use network
information to detect functionally related gene groups
\citep{Segal03b, Shiga07, Ulitsky07}, general-purpose algorithms for a
global analysis of context-specific network activation patterns in a
genome- and organism-wide scale have been missing.

Publication~\ref{NR} introduces and validates two general-purpose
algorithms that provide tools for global modeling of transcriptional
responses in interaction networks. The motivation is similar to
biclustering approaches that detect functionally coherent gene groups
that show coordinated response in a subset of conditions
\citep{Madeira04}. The network ties the findings more tightly to
cell-biological processes, focusing the analysis and improving
interpretability. In contrast to previous network-based biclustering
models for global transcriptome analysis, such as cMonkey
\citep{Reiss06} or SAMBA \citep{Tanay04}, the algorithms introduced in
Publication~\ref{NR} are general-purpose tools, and do not depend on
organism-specific annotations.

\subsubsection{A two-step approach}

The first approach in Publication~\ref{NR} is a straightforward
extension of network-based gene clustering methods. In this two-step
approach, the functionally coherent subnetworks, and their
condition-specific responses are detected in separate steps. In the
first step, a network-based clustering method is used to detect
functionally coherent subnetworks.  In Publication~\ref{NR}, MATISSE,
a state-of-the-art algorithm described in \cite{Ulitsky07}, is used to
detect the subnetworks. MATISSE finds connected subgraphs in the
network that have high internal correlations between the genes.  In
the second step, condition-specific responses of each identified
subnetwork are searched for by a nonparametric Gaussian mixture model,
which allows a data-driven detection of the responses. However, the
two-step approach, coined MATISSE+, can be suboptimal for detecting
subnetworks with particular condition-specific responses. The main
contribution of Publication~\ref{NR} is to introduce a second
general-purpose algorithm, coined NetResponse, where the detection of
condition-specific responses is used as the explicit key criterion for
subnetwork search.

\begin{figure}[t]
\begin{center}
\centering{\includegraphics[width=\textwidth]{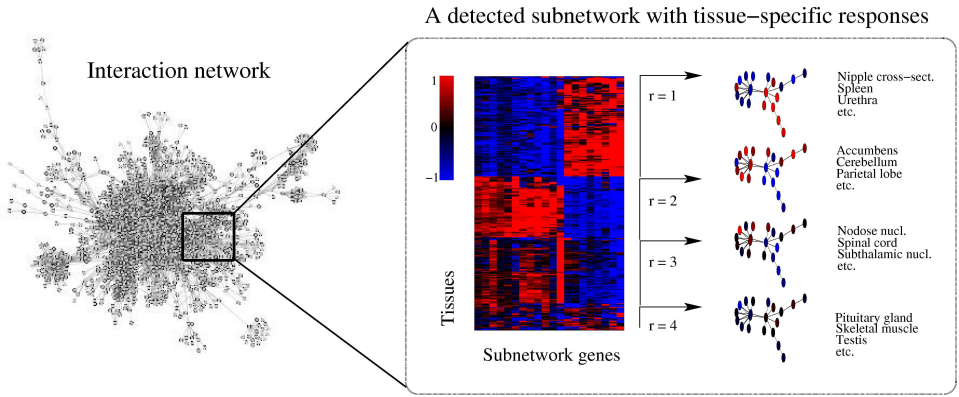}}
\end{center}
\caption{Organism-wide analysis of transcriptional responses in a
  human pathway interaction network reveals physiologically coherent
  activation patterns and condition-specific regulation. One of the
  subnetworks and its condition-specific responses, as detected by the
  NetResponse algorithm is shown in the Figure. The expression of each
  gene is visualized with respect to its mean level of expression
  across all samples. \copyright The Author 2010. Published by Oxford
  University Press. Reprinted with permission from
  Publication~\ref{NR}.}
\label{fig:netresponse}
\end{figure}

\subsubsection{The NetResponse algorithm}\label{sec:netresponse}

The network-based search procedure introduced in Publication~\ref{NR}
searches for local {\it subnetworks}, i.e., functionally coherent
network modules where the interacting genes show coordinated responses
in a subset of conditions (Figure~\ref{fig:netresponse}).  Side
information of the gene interactions is used to guide modeling, but
the algorithm is independent of predefined classifications for genes
or measurement conditions. Transcriptional responses of the network
are described in terms of subnetwork activation. Regulation of the
subnetwork genes can involve simultaneous activation and repression of
the genes: sufficient amounts of mRNA for key proteins has to be
available while interfering genes may need to be silenced. The model
assumes that a given subnetwork \(n\) can have multiple
transcriptional states, associated with different physiological
contexts. A transcriptional state is reflected in a unique expression
signature \(\s^{(n)}\), a vector that describes the expression levels
of the subnetwork genes, associated with the particular
transcriptional state. Expression of some genes is regulated at
precise levels, whereas other genes fluctuate more freely. Given the
state, expression of the subnetwork genes is modeled as a noisy
observation of the transcriptional state. With a Gaussian noise model
with covariance \(\Sigma^{(n)}\), the observation is described by
\(\xn \sim N(\s^{(n)}, \Sigma^{(n)})\). A given subnetwork can have
\(\Rn\) latent transcriptional states indexed by \(r\).  In practice,
the states, including their number \(\Rn\), are unknown, and they have
to be estimated from the data.  In a specific measurement condition,
the subnetwork $n$ can be in any one of the latent physiological
states indexed by $r$. Associations between the observations and the
underlying transcriptional states are unknown and they are treated as
latent variables. Gene expression in subnetwork \(n\) is then modeled
with a Gaussian mixture model:

\begin{equation}
  \xn \sim \sum_{r=1}^{\Rn} \wrn p(\xn | \bth_r ),
  \label{eq:gm}
\end{equation}

\noindent where each component distribution $p$ is assumed to be
Gaussian with parameters \(\bth_r = \{\srn, \Sigrn\}\).  In practice,
we assume a diagonal covariance matrix \(\Sigrn\), leaving the
dependencies between the genes unmodeled within each transcriptional
state. Use of diagonal covariances is justified by considerable gains
in computational efficiency when the detection of distinct responses
is of primary interest. It is possible, however, that such simplified
model will fail to detect certain subnetworks where the
transcriptional levels of the genes have strong linear dependencies
within the individual transcriptional states; signaling cascades
could be expected to manifest such activation patterns, for
instance. More detailed models of transcriptional activity could help
to distinguish the individual states in particular when the
transcriptional states are partially overlapping, but with increased
computational cost. A particular transcriptional response is then
characterized with the triple \(\{\srn, \Sigrn, \wrn\}\). This defines
the shape, fluctuations and frequency of the associated
transcriptional state of subnetwork \(n\).  A posterior probability of
each latent state can be calculated for each measurement sample from
the Bayes' rule (Equation~\ref{eq:bayesrule}). The posterior
probabilities can be interpreted as soft component memberships for the
samples. A hard, deterministic assignment is obtained by selecting for
each sample the component with the highest posterior probability.

The remaining task is to identify the subnetworks having such distinct
transcriptional states. Detection of the distinct states is now used
as a search criterion for the subnetworks. In order to achieve fast
computation, an agglomerative procedure is used where interacting
genes are gradually merged into larger subnetworks.  Initially, each
gene is assigned in its own singleton subnetwork. Agglomeration
proceeds by at each step merging the two neighboring subnetworks where
joint modeling of the genes leads to the highest improvement in the
objective function value. Joint modeling of dependent genes reveals
coordinated responses and improves the likelihood of the data in
comparison with independent models, giving the first criterion for
merging the subnetworks. However, increasing subnetwork size tends to
increase model complexity and the possibility of overfitting, since
the number of samples remains constant while the dimensionality
(subnetwork size) increases. To compensate for this effect, the
Bayesian information criterion \citep[see][]{Gelman03} is used to
penalize increasing model complexity and to determine optimal
subnetwork size. The final cost function for a subnetwork \(G\) is
\(C(G) = - 2\L + q log(N)\), where \(\L\) is the (marginal)
log-likelihood of the data, given the mixture model in
Equation~\ref{eq:gm}, \(q\) is the number of parameters and \(N\) denotes
sample size. The algorithm then compares independent and joint models
for each subnetwork pair that has a direct link in the network, and
merges at each step the subnetwork pair \(G_i, G_j\) that minimizes
the cost

\begin{equation}\label{eq:netreponsecost}
  \Delta\mathcal{C} = - 2(\L_{i,j} - (\L_i + \L_j)) + (q_{i,j} - (q_i + q_j))log(N). 
\end{equation}

The iteration continues until no improvement is obtained by merging
the subnetworks. The combination of modeling techniques yields a
scalable algorithm for genome- and organism-wide investigations:
First, the analysis focuses on those parts of the data that are
supported by known interactions, which increases modeling power and
considerably limits the search space. Second, the agglomerative scheme
finds a fast approximative solution where at each step the subnetwork
pair that leads to the highest improvement in cost function is
merged. Third, an efficient variational approximation is used to learn
the mixture models \citep{Kurihara07nips}. Note that the algorithm
does not necessarily identify a globally optimal solution. However,
detection of physiologically coherent and reproducible responses is
often sufficient for practical applications.

\subsubsection{Global view on network activation patterns}

The NetResponse algorithm introduced in Publication~\ref{NR} was
applied to investigate transcriptional activation patterns of a
pathway interaction network of 1800 genes based on the KEGG database
of metabolic pathways \citep{Kanehisa08} provided by the SPIA package
\citep{Tarca09} across 353 gene expression samples from 65
tissues. The two algorithms proposed in Publication~\ref{NR}, MATISSE+
and NetResponse were shown to outperform an unsupervised biclustering
approach in terms of reproducibility of the finding. The introduced
NetReponse algorithm, where the detection of transcriptional response
patterns is used as a search criterion for subnetwork identification,
was the best-performing method. The algorithm identified 106
subnetworks with 3-20 genes, with distinct transcriptional responses
across the conditions. One of the subnetworks is illustrated in
Figure~\ref{fig:netresponse}; the other findings are provided in the
supplementary material of Publication~\ref{NR}. The detected
transcriptional responses were physiologically coherent, suggesting a
potential functional role. The reproducibility of the responses was
confirmed in an independent validation data set, where 80\% of the
predicted responses were detected (\(p < 0.05\)). The findings
highlight context-specific regulation of the genes.  Some responses
are shared by many conditions, while others are more specific to
particular contexts such as the immune system, muscles, or the brain;
related physiological conditions often exhibit similar network
activation patterns. Tissue relatedness can be measured in terms of
shared transcriptional responses of the subnetworks, giving an
alternative formulation of the tissue connectome map suggested by
\cite{Greco08} in order to highlight functional connectivity between
tissues based on the number of shared differentially expressed genes.
In Publication~\ref{NR}, shared network responses are used instead of
shared gene count. The use of co-regulated gene groups is expected to
be more robust to noise than the use of individual genes.  The
analysis provides a global view on network activation across the
normal human body, and can be used to formulate novel hypotheses of
gene function in previously unexplored contexts.

\section{Conclusion}\label{sec:atlasconclusion}

Gene function and interactions are often subject to condition-specific
regulation \citep{Liang06, Rachlin06}, but these have been typically
studied only in particular experimental conditions. Organism-wide
analysis can potentially reveal new functional connections and help to
formulate novel hypotheses of gene function in previously unexplored
contexts, and to detect highly specialized functions that are specific
to few conditions. Changes in cell-biological conditions induce
changes in the expression levels of co-regulated genes, in order to
produce specific physiological responses, typically affecting only a
small part of the network. Since individual processes and their
transcriptional responses are in general unknown \citep{Lee08c,
  Montaner09}, data-driven detection of condition-specific responses
can provide an efficient proxy for identifying distinct
transcriptional states of the network, with potentially distinct
functional roles.

Publication~\ref{NR} provides efficient model-based tools for global,
organism-wide discovery and characterization of context-specific
transcriptional activity in genome-scale interaction networks,
independently of predefined classifications for genes and
conditions. The network is used to bring in prior information of gene
function, which would be missing in unsupervised models, and allows
data-driven detection of coordinately regulated gene sets and their
context-specific responses. The algorithm is readily applicable in any
organism where gene expression and pairwise interaction data,
including pathways, protein interactions and regulatory networks, are
available. It has therefore a considerably larger scope than previous
network-based models for global transcriptome analysis, which rely on
organism-specific annotations, but lack implementations for most
organisms \citep{Reiss06, Tanay04}.

While biomedical implications of the findings require further
investigation, the results highlight shared and reproducible responses
between physiological conditions, and provide a global view of
transcriptional activation patterns across the normal human
body. Other potential applications for the method include large-scale
screening of drug responses and disease subtype
discovery. Implementation of the algorithm is freely available through
BioConductor.\footnote{http://bioconductor.org/packages/devel/bioc/html/netresponse.html}

\newpage

\chapter{Human transcriptome and other layers of genomic information}\label{ch:integration}

\begin{quotation}
  \emph{The way to deal with the problem of big data is to beat it
    senseless with other big data.}
\begin{flushright}
J. Quackenbush (2006) 
\end{flushright}
\end{quotation}

This chapter presents the third main contribution of the thesis,
computational strategies to integrate measurements of human
transcriptome to other layers of genomic information.  Genomic,
transcriptomic, proteomic, epigenomic and other sources of measurement
data characterize different aspects of genome organization
\citep{Hawkins2010, Montaner2010, Sara2010}; any single source
provides only a limited view to the cellular system. Understanding
functional organization of the genome and ultimately the cell function
requires integration of data from the various levels of genome
organization and modeling of their dynamical interplay.  Such an
holistic approach, which is also called {\it systems biology}, is a
key to understanding living organisms, which are ``rich in emergent
properties because forever new groups of properties emerge at every
level of integration'' \citep{Mayr04}. Combining evidence across
multiple sources can help to discover functional mechanisms and
interactions, which are not seen in the individual data sets, and to
increase statistical power in noisy and incomplete high-throughput
experiments \citep{Huttenhower2010, Reed2006}.

Integration of heterogeneous genomic data comes with a variety of
technical and methodological challenges \citep{Hwang05,
  Troyanskaya05}, and the particular modeling approaches vary
according to the analysis task and particular properties of the
investigated measurement sources.  Integrative studies have been
limited by poor availability of co-occurring genomic observations, but
suitable data sets are now becoming increasingly available in both
in-house and public biomedical data repositories \citep{tcga08}.  New
observations highlight the need for novel, integrative approaches in
functional genomics \citep{Coe08}. Recent studies have proposed for
instance methods to integrate epigenetic modifications
\citep{Sadikovic2008}, micro-RNA \citep{Qin2008}, transcription factor
binding \citep{Savage2010}, as well as protein expression
\citep{Johnson2008}.  Given the complex stochastic nature of
biological systems, computational efficiency, robustness against
uncertainty and interpretability of the results are key issues. Prior
information of biological systems is often incomplete, and subject to
high levels of uncontrolled variation and complex interdependencies
between different parts of the cellular system \citep{Troyanskaya05}.
These issues emphasize the need for principled approaches requiring
minimal prior knowledge about the data, as well as minimal model
fitting procedures. Section~\ref{sec:traditionaldep} gives an overview
of the standard models for high-throughput data integration methods,
which have close connections to the modeling approaches developed in
this work.

\section{Standard approaches for genomic data integration}\label{sec:traditionaldep}

The integrative approaches can be roughly classified in three
categories: methods that (i) combine statistical evidence across
related studies in order to obtain more accurate inferences of target
variables, (ii) utilize side information in order to guide the
analysis of a single, primary data source, and (iii) detect and
characterize dependencies between the measurement sources in order to
discover new functional connections between the different layers of
genomic information. The contributions in Chapters~\ref{ch:preproc}
and~\ref{ch:atlases} are associated with the first two categories; the
contributions presented in this chapter, the regularized dependency
detection framework of Publication~\ref{MLSP}, and associative
clustering of Publications~\ref{ECML} and~\ref{AC}, belong to the
third category.

\subsection{Combining statistical evidence}

The first general category of methods for genomic data integration
consists of approaches where evidence across similar studies is
combined to increase statistical power, for instance by comparing and
integrating data from independent microarray experiments targeted at
studying the same disease. In Publications~\ref{RPA} and~\ref{NR},
joint analysis of a large number of commensurable microarray
experiments, where the observed data is directly comparable between
the arrays, helps to increase statistical power and to reveal weak,
shared signals in the data that can not be detected in more restricted
experimental setups and smaller datasets. 

However, the related observations are often not directly comparable,
and further methodological tools are needed for integration. {\it
Meta-analysis} provides tools for such analysis
\citep{Ramasamy08b}. Meta-analysis forms part of the microarray
analysis procedure introduced in Publication~\ref{PECA}, where methods
to integrate related microarray measurements across different array
platforms are developed. Meta-analysis emphasizes shared effects
between the studies over statistical significance in individual
experiments. In its standard form, meta-analysis assumes that each
individual study measures the same target variable with varying levels
of noise. The analysis starts from identifying a measure of {\it
effect size} based on differences, means, or other summary statistics
of the observations such as the Hedges' g, used in
Publication~\ref{PECA}. Weighted averaging of the effect sizes
provides the final, combined result. Weighting accounts for
differences in reliability of the individual studies, for instance by
emphasizing studies with large sample size, or low measurement
variance. Averaging is expected to yield more accurate estimates of
the target variable than individual studies. This can be particularly
useful when several studies with small sample sizes are available for
instance from different laboratories, which is a common setting in
microarray analysis context, where the data sets produced by
individual laboratories are routinely deposited to shared community
databases. Ultimately, the quality of meta-analysis results rests on
the quality of the individual studies. Modeling choices, such as the
choice of the effect size measure and included studies will affect the
analysis outcome.

{\it Kernel methods} \citep[see e.g.][]{Scholkopf02} provide another
widely used approach for integrating statistical evidence across
multiple, potentially heterogeneous measurement sources. Kernel
methods operate on similarity matrices, and provide a natural
framework for combining statistical evidence to detect similarity and
patterns that are supported by multiple observations. The modeling
framework also allows for efficient modeling of nonlinear feature
spaces.

{\it Multi-task learning} refers to a class of approaches where
multiple, related modeling tasks are solved simultaneously by
combining statistical power across the related tasks. A typical task
is to improve the accuracy of individual classifiers by taking
advantage of the potential dependencies between them \citep[see
e.g.][]{Caruana97}.

\subsection{Role of side information}

The second category of approaches for genomic data integration
consists of methods that are asymmetric by nature; integration is used
to support the analysis of one, primary data source. Side information
can be used, for instance, to limit the search space and to focus the
analysis to avoid overfitting, speed up computation, as well as to
obtain potentially more sensitive and accurate findings \citep[see
e.g.][]{Eisenstein06}. One strategy is to impose hard constraints on
the model, or model family, based on side information to target
specific research questions. In gene expression context, functional
classifications or known interactions between the genes can be used to
constrain the analysis \citep{Goeman07, Ulitsky09}. In factor analysis
and mixed effect models, clinical annotations of the samples help to
focus the modeling on particular conditions \citep[see
e.g.][]{Carvalho08}. Hard constraints rely heavily on the accuracy of
side information. Soft, or probabilistic approaches can take the
uncertainty in side information into account, but they are
computationally more demanding. Examples of such methods in the
context of transcriptome analysis include for instance the supervised
biclustering models, such as cMonkey and modified SAMBA, as well as
other methods that guide the analysis with additional information of
genes and regulatory mechanisms, such as transcription factor binding
\citep{Reiss06, Savage2010, Tanay04}. Publication~\ref{NR} uses gene
interaction network as a hard constraint for modeling transcriptional
co-regulation of the genes, but the condition-specific responses of
the detected gene groups are identified in an unsupervised manner.

A complementary approach for utilizing side information of the
experiments is provided by {\it multi-way learning}. A classical
example is the analysis of variance (ANOVA), where a single data set
is modeled by decomposing it into a set of basic, underlying effects,
which characterize the data optimally. The effects are associated with
multiple, potentially overlapping attributes of the measurement
samples, such as disease state, gender and age, which are known prior
to the analysis. Taking such prior knowledge of systematic variation
between the samples into account helps to increase modeling power and
can reveal the attribute-specific effects. An interesting subtask is
to model the interactions between the attributes, so-called {\it
interaction effects}. These are manifested only with particular
combinations of attributes, and indicate dependency between the
attributes. For instance, simultaneous cigarette smoking and asbestos
exposure will considerably increase the risk of lung cancer, compared
to any of the two risk factors alone \citep[see
e.g.][]{Nymark07}. {\it Factor analysis} is a closely related approach
where the attributes, also called {\it factors}, are not given but
instead estimated from the data. {\it Mixed effect models} combine the
supervised and unsupervised approaches by incorporating both {\it
fixed} and {\it random effects} in the model, corresponding to the
known and latent attributes, respectively \citep[see
e.g.][]{Carvalho08}. The standard factorization approaches for
individual data sets are related to the dependency-seeking approaches
in Publications~\ref{MLSP}-\ref{AC}, where co-occurring data sources
are decomposed in an unsupervised manner into components that are
maximally informative of the components in the other data set. 

\subsection{Modeling of mutual dependency}

Symmetric models for dependency detection form the third main category
of methods for genomic data integration, as well as the main topic of
this chapter. Dependency modeling is used to distinguish the {\it
shared} signal from {\it dataset-specific} variation. The shared
effects are informative of the commonalities and interactions between
the observations, and are often the main focus of interest in
integrative analysis. This motivates the development of methods that
can allocate computational resources efficiently to modeling of the
shared features and interactions.

{\it Multi-view learning} is a general category of approaches for
symmetric dependency modeling tasks. In multi-view learning, multiple
measurement sources are available, and each source is considered as a
different view on the same objects. The task is to enhance modeling
performance by combining the complementary views. A classical example
of such a model is canonical correlation analysis
\citep{Hotelling36}. Related approaches that have recently been applied
in functional genomics include for instance probabilistic variants of
meta-analysis \citep{Choi07, Conlon07}, generalized singular value
decomposition \citep[see e.g.][]{Alter03, Berger06} and simultaneous
non-negative matrix factorization \citep{Badea08}.

The dependency modeling approaches in this thesis make an explicit
distinction between statistical representation of data and the
modeling task.  Let us denote the representations of two co-occurring
multivariate observations, $\x$ and $\y$, with $f_x(\x)$ and
$f_y(\y)$, respectively. The selected representations depend on the
application task. The representation can be for instance used to
perform feature selection as in {\it canonical correlation analysis
(CCA)} \citet{Hotelling36}, capture nonlinear features in the data as
in kernelized versions of CCA \citep[see e.g.][]{Yamanishi03}, or
partition the data as in information bottleneck \citep{Friedman01} and
associative clustering (Publications~\ref{ECML}-\ref{AC}). {\it
Statistical independence} of the representations implies that their
joint probability density can be decomposed as \(p(f_x(\x), f_y(\y)) =
p(f_x(\x))p(f_y(\y))\). Deviations from this assumption indicate
statistical dependency. The representations can have a flexible
parametric form which can be optimized by the dependency modeling
algorithms to identify dependency structure in the data.

Recent examples of such dependency-maximizing methods include
probabilistic canonical correlation analysis \citep{Bach05}, which has
close theoretical connection to the regularized models introduced in
Publication~\ref{MLSP}, and the associative clustering principle
introduced in Publications~\ref{ECML}-\ref{AC}. Canonical
correlations and contingency table analysis form the methodological
background for the contributions in Publications~\ref{MLSP}-\ref{AC}.
In the remainder of this section these two standard approaches for
dependency detection are considered more closely.

\subsubsection{Classical and probabilistic canonical correlation analysis}

Canonical correlation analysis (CCA) is a classical method for
detecting linear dependencies between two multivariate random
variables \citep{Hotelling36}. While ordinary correlation
characterizes the association strength between two vectors with paired
scalar observations, CCA assumes paired vectorial values, and
generalizes correlation to multidimensional sources by searching for
maximally correlating low-dimensional representation of the two
sources, defined by linear projections \(\X\vx, \Y\vy\). Multiple
projection components can be obtained iteratively, by finding the most
correlating projection first, and then consecutively the next ones
after removing the dependencies explained by the previous CCA
components; the lower-dimensional representations are defined by
projections to linear hyperplanes. The model can be formulated as a
generalized eigenvalue problem that has an analytical solution with
two useful properties: the result is invariant to linear
transformations of the data, and the solution for any fixed number of
components maximizes mutual information between the projections for
Gaussian data \citep{Kullback59, Bach02}. Extensions of the classical
CCA include generalizations to multiple data sources
\citep{Kettenring71, Bach02}, regularized solutions with non-negative
and sparse projections \citep{Sigg07, Archambeau08, Witten09}, and
non-linear extensions, for instance with kernel methods \citep{Bach02,
Yamanishi03}. Direct optimization of correlations in the classical CCA
provides an efficient way to detect dependencies between data sources,
but it lacks an explicit model to deal with the uncertainty in the
data and model parameters.

Recently, the classical CCA was shown to correspond to the ML solution
of a particular generative model where the two data sets are assumed
to stem from a shared Gaussian latent variable \(\z\) and normally
distributed data-set-specific noise \citep{Bach05}. Using linear
assumptions, the model is formally defined as

\begin{flalign}\label{eq:genmodel}
     \left\{
   \begin{array}{cl}
	\x &\sim \Wx \z + \Epsx\\
	\y &\sim \Wy \z + \Epsy.
  \end{array}
\right.
\end{flalign}

\begin{figure}[ht!]
\centering{ 
\includegraphics[width=.4\textwidth]{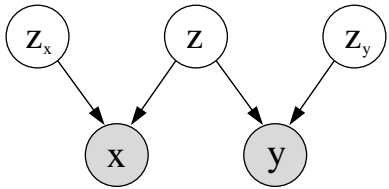}
}
\caption{A graphical representation of the generative shared latent
  variable model in Equation~(\ref{eq:genmodel}). The latent source
  $\z$ is shared by observations $\x$ and $\y$. The other effects that
  are specific to each observation are characterized by $\z_x$ and
  $\z_y$, respectively. Gray shading indicates observed variables.}
\label{fig:modelpic}
\end{figure}

\noindent The manifestation of the shared signal in each data set can
be different. This is parameterized by \(\W_x\) and \(\W_y\).
Assuming a standard Gaussian model for the shared latent variable,
\(\z \sim \N(\0,\I)\) and data set-specific effects where \(\Epsx \sim
\N(\0, \Psi_x)\) (and respectively for \(\y\)), the
correlation-maximizing projections of the traditional CCA introduced
in Section~\ref{sec:traditionaldep} can be retrieved from the ML
solution of the model \citep{Archambeau06, Bach05}. The model
decomposes the observed co-occurring data sets into {\it shared} and
{\it data set-specific} components based on explicit modeling
assumptions (Figure~\ref{fig:modelpic}). The dataset-specific effects
can also be described in terms of latent variables as \(\Epsx =
\B_x\z_x\) and \(\Epsy = \B_y\z_y\), allowing the construction of more
detailed models for the dataset-specific effects \citep{Klami08}. The
shared signal \(\z\) is treated as a latent variable and marginalized
out in the model, providing the marginal likelihood for the
observations:

\begin{equation}\label{eq:ccalikelihood}
    p(\X,\Y|\W,\Psi) = \int p(\X, \Y|\Z, \W, \Psi)p(\Z) d\Z,
\end{equation}

\noindent where \(\Psi\) denotes the block-diagonal matrix of
\(\Psi_x\), \(\Psi_y\), and \(\W = [\W_x; \W_y]\).  The probabilistic
formulation of CCA has opened up a way to new probabilistic extensions
that can treat the modeling assumptions and uncertainties in the data
in a more explicit and robust manner \citep{Archambeau06, Klami08,
Klami10uai}.

The general formulation provides a flexible modeling framework, where
different modeling assumptions can be used to adapt the models in
different applications. The connection to classical CCA assumes full
covariances for the dataset-specific effects. Simpler models for the
dataset-specific effects will not distinguish between the shared and
marginal effects as effectively, but they have fewer model parameters
that can potentially reduce overlearning and speed up computation.  It
is also possible to tune the dimensionality of the shared latent
signal. Learning of lower-dimensional models can be faster and
potentially less prone to overfitting. Interpretation of simpler
models is also more straightforward in many applications. The
probabilistic formulation allows rigorous treatment of uncertainties
in the data and model parameters also with small sample sizes that are
common in biomedical studies, and allows the incorporation of prior
information through Bayesian priors, as in the regularized dependency
detection framework introduced in Publication~\ref{MLSP}.

\subsubsection{Contingency table analysis}

Contingency table analysis is a classical approach used to study
associations between co-occurring categorical observations.  The
co-occurrences are represented by cross-tabulating them on a {\it
contingency table}, the rows and columns of which correspond to the
first and second set of features, respectively. Various tests are
available for measuring dependency between the rows and columns of the
table \citet{Yates34, Agresti92}, including the classical Fisher test
\citep{Fisher34}, a standard tool for measuring statistical enrichment
of functional categories in gene cluster analysis
\citep{Hosack03}. While the classical contingency table analysis is
used to measure dependency between co-occurring variables, more recent
approaches use contingency tables to derive objective functions for
dependency exploration tasks. The associative clustering principle
introduced in Publications~\ref{ECML}-\ref{AC} is an example of such
approach.

Other approaches that use contingency table dependencies as objective
functions include the {\it information bottleneck (IB)} principle
\citep{Tishby99} and {\it discriminative clustering (DC)}
\citep{Sinkkonen02ecml, Kaski05nc}. These are asymmetric,
dependency-seeking approaches that can be used to discover cluster
structure in a primary data such that it is maximally informative of
another, discrete auxiliary variable. The dependency is represented on
a contingency table, and maximization of contingency table
dependencies provides the objective function for clustering. While the
standard IB operates on discrete data, DC is used to discover cluster
structure in continuous-valued data. The two approaches also employ
different objective functions. In classical IB, a discrete variable
$\Xcal$ is clustered in such a way that the cluster assignments become
maximally informative of another discrete variable $\Ycal$. The
complexity of the cluster assignments is controlled by minimizing the
mutual information between the cluster indices and the original
variables. The task is to find a partitioning $\tilde{\X}$ that
minimizes the cost \(\L(p(\tilde{\X}|\X)) = I(\tilde{\X}; \X) - \beta
I(\tilde{\X}; \Y),\) where $\beta$ controls clustering resolution. In
DC, mutual information is replaced by a Bayes factor between the two
hypotheses of dependent and independent margins. The Bayes factor is
asymptotically consistent with mutual information, but provides an
unbiased estimate for limited sample size \citep[see
e.g.][]{Sinkkonen05tr}. The standard information bottleneck and
discriminative clustering are asymmetric methods that treat one of the
data sources as the primary target of analysis.

In contrast, the dependency maximization approaches considered in this
thesis, the associative clustering (AC) and regularized versions of
canonical correlation analysis are symmetric and they operate
exclusively on continuous-valued data.  CCA is not based on
contingency table analysis, but it has close connections to the
Gaussian IB \citep{Chechik05} that seeks maximal dependency between
two sets of normally distributed variables. The Gaussian IB retrieves
the same subspace as CCA for one of the data sets. However, in
contrast to the symmetric CCA model, Gaussian IB is a directed method
that finds dependency-maximizing projections for only one of the two
data sets.  The second dependency detection approach considered in
this thesis, the associative clustering, is particularly related to
the symmetric IB that finds two sets of clusters, one for each
variable, which are optimally compressed presentations of the original
data, and at the same time maximally informative of each other
\citep{Friedman01}. While the objective function in IB is derived from
mutual information, AC uses the Bayes factor as an objective function
in a similar manner as it is used in the asymmetric discriminative
clustering. Another key difference is that while the symmetric IB
operates on discrete data, AC employs contingency table analysis in
order to discover cluster structure in continuous-valued data spaces.

\section{Regularized dependency detection}\label{sec:pairwise}

Standard unsupervised methods for dependency detection, such as the
canonical correlation analysis or the symmetric information bottleneck,
seek maximal dependency between two data sets with minimal assumptions
about the dependencies. The unconstrained models involve high degrees
of freedom when applied to high-dimensional genomic observations. Such
flexibility can easily lead to overfitting, which is even worse for
more flexible nonparametric or nonlinear, kernel-based dependency
discovery methods.  Several ways to regularize the solution have been
suggested to overcome associated problems, for instance by imposing
sparsity constraints on the solution space \citep{Bie03, Vinod76}.

In many applications prior information of the dependencies is
available, or particular types of dependency are relevant for the
analysis task. Such prior information can be used to reduce the
degrees of freedom in the model, and to regularize dependency
detection. In the cancer gene discovery application of
Publication~\ref{MLSP}, DNA mutations are systematically correlated
with transcriptional activity of the genes within the affected region,
and identification of such regions is a biomedically relevant research
task. Prior knowledge of chromosomal distances between the
observations can improve the detection of the relevant spatial
dependencies. However, principled approaches to incorporate such prior
information in dependency modeling have been
missing. Publication~\ref{MLSP} introduces regularized models for
dependency detection based on classical canonical correlation analysis
\citep{Hotelling36} and its probabilistic formulation
\citep{Bach05}. The models are extended by incorporating appropriate
prior terms, which are then used to reduce the degrees of freedom
based on prior biological knowledge.

\subsubsection{Correlation-based variant}

In order to introduce the regularized dependency detection framework
of Publication~\ref{MLSP}, let us start by considering regularization
of the classical correlation-based CCA. This searches for arbitrary
linear projection vectors \(\vx, \vy\) that maximize the correlation
between the projections of the data sets \(\X, \Y\). Multiple
projection components can be obtained iteratively, by finding the most
correlating projection first, and then consecutively the next ones
after removing the dependencies explained by the previous CCA
components. The procedure will identify maximally dependent linear
subspaces of the investigated data sets. To regularize the solution,
Publication~\ref{MLSP} couples the projections through a
transformation matrix \(\T\) in such a way that \(\vy = \T \vx\). With
a completely unconstrained \(\T\) the model reduces to the classical
unconstrained CCA; suitable constraints on can be used to regularize
dependency detection.

To enforce regularization one could for instance prefer solutions for
\(\T\) that are close to a given transformation matrix, \(\T \sim
\M\), or impose more general constraints on the structure of the
transformation matrix that would prefer particular rotational or other
linear relationships. Suitable constraints depend on the particular
applications; the solutions can be made to prefer particular types of
dependency in a soft manner by appropriate penalty terms. In
Publication~\ref{MLSP} the completely unconstrained CCA model has been
compared with a fully regularized model with \(\T = \I\); this encodes
the biological assumption that probes with small chromosomal distances
tend to capture more similar signal between gene expression and copy
number measurements than probes with a larger chromosomal distance;
the projection vectors characterize this relationship, and are
therefore expected to have similar form, \(\vx \sim \vy\). Utilization
of other, more general constraints in related data integration tasks
provides a promising topic for future studies.

The correlation-based treatment provides an intuitive and easily
implementable formulation for regularized dependency
detection. However, it lacks an explicit model for the shared and
data-specific effects, and it is likely that some of the
dataset-specific effects are captured by the correlation-maximizing
projections. This is suboptimal for characterizing the shared effects,
and motivates the probabilistic treatment.

\subsubsection{Probabilistic dependency detection with similarity constraints}

The probabilistic approach for regularized dependency detection in
Publication~\ref{MLSP} is based on an explicit model of the
data-generating process formulated in Equation~(\ref{eq:genmodel}). In
this model, the transformation matrices \(\W_x\), \(\W_y\) specify how
the shared latent variable \(\Z\) is manifested in each data set
\(\X\), \(\Y\), respectively. In the standard model, the relationship
between the transformation matrices is not constrained, and the
algorithm searches for arbitrary linear transformations that maximize
the likelihood of the observations in
Equation~(\ref{eq:ccalikelihood}). The probabilistic formulation opens
up possibilities to guide dependency search through Bayesian priors.

In Publication~\ref{MLSP}, the standard probabilistic CCA model is
extended by incorporating additional prior terms that regularize the
relationship by reparameterizing the transformation matrices as \(\W_y
= \T\W_x\), and setting a prior on \(\T\). The treatment is analogous
to the correlation-based variant, but now the transformation matrices
operate on the latent components, rather than the observations. This
allows to distinguish the shared and dataset-specific effects more
explicitly in the model. The task is then to learn the optimal
parameter matrix \(\W = [\Wx; \Wy]\), given the constraint \(\W_y =
\T\W_x\). The Bayes' rule gives the model likelihood

\begin{equation}\label{eq:ccafull}
  p(\X, \Y, \W, \Psib) \sim p(\X, \Y | \W, \Psib) p(\W, \Psib).
\end{equation}

\noindent The likelihood term \(p(\X, \Y | \W, \Psib)\) can be
calculated based on the model in Equation~(\ref{eq:genmodel}). This
defines the objective function for standard probabilistic CCA, which
implicitly assumes a flat prior \(p(\W, \Psib) \sim 1\) for the model
parameters. The formulation in Equation~(\ref{eq:ccafull}) makes the
choice of the prior explicit, allowing modifications on the prior
term. To obtain a tractable prior, let us assume that the prior
factorizes as \(p(\W, \Psib) = p( \W ) p( \Psib )\). The first term
can be further decomposed as \(p(\W) \sim p(\W_x) p(\T)\), assuming
independent priors for \(\W_x\) and \(\T\). A convenient and tractable
prior for \(\T\) is provided by the matrix normal
distribution:\footnote{\(\Nm(\T | \M, \U, \Vb) \sim
  exp\left(-\frac{1}{2} Tr\{\U^{-1}(\T-\M)\Vb^{-1}(\T -
    \M)^T\}\right)\) where \(\M\) is the mean matrix, and \(\U\) and
  \(\Vb\) denote row and column covariances, respectively.}

\begin{equation}\label{eq:Nm}
  p(\T) = \Nm(\T | \M, \U, \Vb).
\end{equation}

\noindent For computational simplicity, let us assume independent rows
and columns with \(\U = \Vb = \sigma_T \I\). The mean matrix \(\M\)
can be used to emphasize certain types of dependency between \(\Wx\)
and \(\Wy\). Assuming uninformative, flat priors \(p(\Wx) \sim 1\) and
\(p(\Psib) \sim 1\), as in the standard probabilistic CCA model, and
denoting \(\bSigma = \W\W^T + \Psib \), the negative log-likelihood of
the model is

\begin{equation}\label{eq:simccacost}
  -logp(\X, \Y, \W, \Psib) \sim log|\bSigma| + Tr \bSigma^{-1} \tilde{\bSigma} +
  \frac{\parallel \T - \M \parallel_{F}^2}{2\sigt}.
\end{equation}

\noindent This is the objective function to minimize. Note that this
has the same form as the objective function of the standard
probabilistic CCA, except the additional penalty term
\(\frac{\parallel \T - \M \parallel_{F}^2}{2\sigt}\) arising from the
prior \(p(\T)\). This yields the cost function employed in
Publication~\ref{MLSP}. In our cancer gene discovery application the
choice \(\M = \I\) is used to encode the biological prior constrain
\(\T \approx \I\), which states that the observations with a small
chromosomal distance should on average show similar responses in the
integrated data sets, i.e., \(\Wx \approx \Wy\). The regularization
strength can be tuned with \(\sigt\). A fully regularized model is
obtained with \(\sigt \rightarrow 0\). When \(\sigt \rightarrow
\infty\), $\W_x$ and $\W_y$ become independent {\it a priori},
yielding the ordinary probabilistic CCA. The \(\sigt\) can be used to
regularize the solution between these two extremes. Note that it is
possible to incorporate also other types of prior information
concerning the dependencies into the model through \(p(\T)\).

The model parameters \(\W\), \(\Psib\) are estimated with the EM
algorithm. The regularized version is not analytically tractable with
respect to \(\W\) in the general case, but can be optimized with
standard gradient-based optimization techniques. Special cases of the
model have analytical solutions, which can speed up the model fitting
procedure. In particular, the fully regularized and unconstrained
models, obtained with \(\sigt = 0\) and \(\sigt = \infty\)
respectively, have closed-form solutions for \(\W\). Note that the
current formulation assumes that the regularization parameters \(\M,
\sigt\) are defined prior to the analysis. Alternatively, these
parameters could be optimized based on external criteria, such as
cancer gene detection performance in our application, or learned from
the data in a fully Bayesian treatment these parameters would be
treated as latent variables. Incorporation of additional prior
information of the data set-specific effects through priors on
\(\W_x\) and \(\Psib\) provides promising lines for further work.

\subsection{Cancer gene discovery with dependency detection}

The regularized models provide a principled framework for studying
associations between transcriptional activity and other regulatory
layers of the genome. In Publication~\ref{MLSP}, the models are used
to investigate cancer mechanisms. DNA copy number changes are a key
mechanism for cancer, and integration of copy number information with
mRNA expression measurements can reveal functional effects of the
mutations. While causation may be difficult to grasp, study of the
dependencies can help to identify functionally active mutations, and
to provide candidate biomarkers with potential diagnostic, prognostic
and clinical impact in cancer studies.

The modeling task in the cancer gene discovery application of
Publication~\ref{MLSP} is to identify chromosomal regions that show
exceptionally high levels of dependency between gene copy number and
transcriptional levels.  The model is used to detect dependency within
local chromosomal regions that are then compared in order to identify
the exceptional regions. The dependency is quantified within a given
region by comparing the strength of shared and data set-specific
signal. High scores indicate regions where the shared signal is
particularly high relative to the data-set-specific effects.  A
sliding-window approach is used to screen the genome for
dependencies. The regions are defined by the \(d\) closest probes
around each gene. Then the dimensionality of the models stays
constant, which allows direct comparison of the dependency measures
between the regions without additional adjustment terms that would be
otherwise needed to compensate for differences in model complexity.

Prior information of the dependencies is used to regularize cancer
gene detection. Chromosomal gains and losses are likely to be
positively correlated with the expression levels of the affected genes
within the same chromosomal region or its close proximity; copy number
gain is likely to increase the expression of the associated genes
whereas deletion will block gene expression.  The prior information is
encoded in the model by setting \(\M = \I\) in the prior term
\(p(\T)\). This accounts for the expected positive correlations
between gene expression and copy number within the investigated
chromosomal region. Regularization based on such prior information is
shown to improve cancer gene detection performance in
Publication~\ref{MLSP}, where the regularized variants outperformed
the unconstrained models.

\begin{figure}[ht!]
\centerline{
\includegraphics[width=1\textwidth]{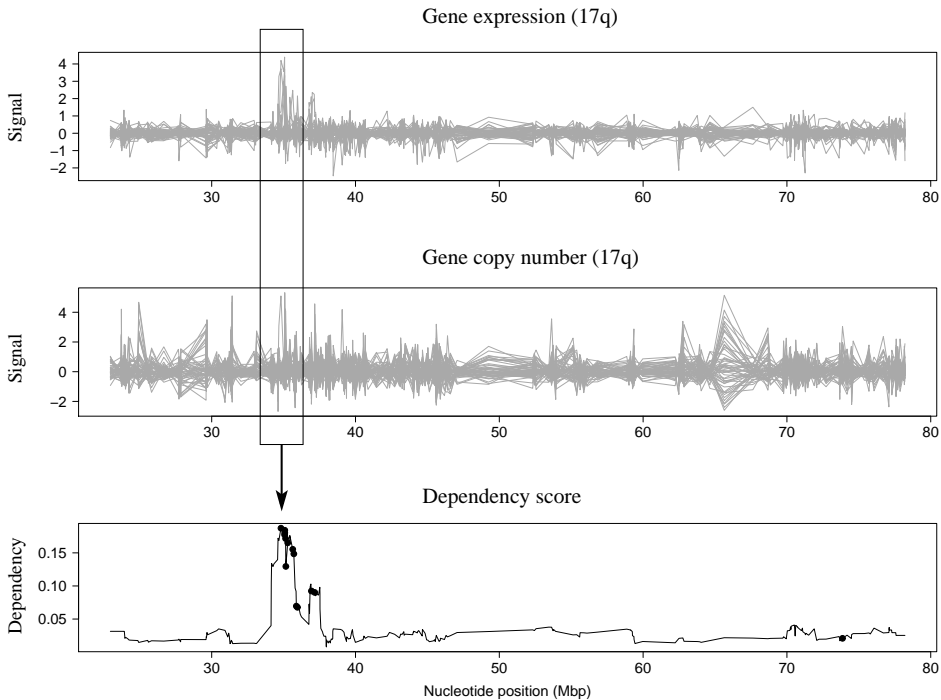}} 
\caption{Gene expression, copy number signal, and the dependency score
  along the chromosome arm 17q obtained with the regularized latent
  variable framework in Equation~\ref{eq:simccacost}. Known
  cancer-associated genes from an expert-curated list are marked with
  black dots.}
\label{fig:17q}
\end{figure}

A genome-wide screen of 51 gastric cancer patients
\citep{Myllykangas08jc} reveals clear associations between DNA copy
number changes and transcriptional activity. The Figure~\ref{fig:17q}
illustrates dependency detection on chromosome arm 17q, where the
regularized model reveals high dependency between the two data sources
in a known cancer-associated region.  The regularized and
unconstrained models were compared in terms of receiver-operator
characteristics calculated by comparing the ordered gene list from the
dependency screen to an expert-curated list of known genes associated
with gastric cancer \citep{Myllykangas08jc}. A large proportion of the
most significant findings in the whole-genome analysis were known
cancer genes; the remaining findings with no known associations to
gastric cancer are promising candidates for further study.

Biomedical interpretation of the model parameters is also
straightforward. A ML estimate of the latent variable values \(\Z\)
characterizes the strength of the shared signal between DNA mutations
and transcriptional activity for each patient. This allows robust
identification of small, potentially unknown patient subgroups with
shared amplification effects. These would remain potentially
undetected when comparing patient groups defined based on existing
clinical annotations. The parameters in \(\W\) can downweigh signal
from poorly performing probes in each data set, or probes that measure
genes whose transcriptional levels are not functionally affected by
the copy number change. This provides tools to distinguish between
so-called {\it driver} mutations having functional effects from less
active {\it passenger} mutations, which is an important task in cancer
studies.  On the other hand, the model can combine statistical power
across the adjacent measurement probes, and it captures the strongest
shared signal in the two sets of observations. This is useful since
gene expression and copy number data are typically characterized by
high levels of biological and measurement variation and small sample
size.

\subsubsection{Related approaches}

Integration of chromosomal aberrations and transcriptional activity is
an actively studied data integration task in functional genomics. The
first studies with standard statistical tests were carried out by
\citet{Hyman2002} and \citet{Phillips2001} when simultaneous
genome-wide observations of the two data sources had become available.
The modeling approaches utilized in this context can be roughly
classified in regression-based, correlation-based and latent variable
approaches. The regression-based models \citep{Adler2006, Bicciato09,
  Wieringen09} characterize alterations in gene expression levels
based on copy number observations with multivariate regression or
closely related models. The correlation-based approaches
\citep{Gonzalez2009, Schafer09, Soneson2010} provide symmetric models
for dependency detection, based on correlation and related statistical
models. Many of these methods also regularize the solutions, typically
based on sparsity constraints and non-negativity of the projections
\citep{LeCao2009, Waaijenborg2008, Witten09, Parkhomenko2009}. The
correlation-based approach in Publication~\ref{MLSP} introduces a
complementary approach for regularization that constrains the
relationship between subspaces where the correlations are
estimated. The latent variable models by \cite{Berger06, Shen09,
  Vaske2010}, and Publication~\ref{MLSP} are based on explicit
modeling assumptions concerning the data-generating processes. The
iCluster algorithm \citep{Shen09} is closely related to the latent
variable model considered in Publication~\ref{MLSP}. While our model
detects continuous dependencies, iCluster uses a discrete latent
variable to partition the samples into distinct subgroups.  The
iCluster model is regularized by sparsity constraints on \(\W\), while
we tune the relationship between \(\Wx\) and \(\Wy\). Moreover, the
model in Publication~\ref{MLSP} utilizes full covariance matrices to
model for the dataset-specific effects, whereas iCluster uses diagonal
covariances. The more detailed model for dataset-specific effects in
our model should help to distinguish the shared signal more
accurately. Other latent variable approaches include the iterative
method based on generalized singular-value decomposition
\citep{Berger06}, and the probabilistic factor graph model PARADIGM
\citep{Vaske2010}, which additionally utilizes pathway topology
information in the modeling.

Experimental comparison between the related integrative approaches can
be problematic since they target related, but different research
questions where the biological ground truth is often unknown.  For
instance, some methods utilize patient class information in order to
detect class-specific alterations \citep{Schafer09}, other methods
perform {\it de novo} class discovery \citep{Shen09}, provide tools
for gene prioritization \citep{Salari2010}, or guide the analysis with
additional functional information of the genes \citep{Vaske2010}. The
algorithms introduced in Publication~\ref{MLSP} are particularly
useful for gene prioritization and class discovery purposes, where the
target is to identify the most promising cancer gene candidates for
further validation, or to detect potentially novel cancer
subtypes. However, while an increasing number of methods are released
as conveniently accessible algorithmic tools \citep{Salari2010,
  Shen09, Schafer09, Witten09}, implementations of most models are not
available for comparison purposes. Open source implementations of the
dependency detection algorithms developed in this thesis have been
released to enhance transparency and reproducibility of the
computational experiments and to encourage further use of these models
\citep{Lahti10pint}.

\section{Associative clustering}

Functions of human genes are often studied indirectly, by studying
model organisms such as the mouse \citep{Davis04, Joyce06}. Orthologs
are genes in different species that originate from a single gene in
the last common ancestor of these species. Such genes have often
retained identical biological roles in the present-day organisms, and
are likely to share the function \citep{Fitch70}. Mutations in the
genomic DNA sequence are a key mechanism in evolution. Consequently,
DNA sequence similarity can provide hypotheses of gene function in
poorly annotated species. An exceptional level of conservation may
highlight critical physiological similarities between species, whereas
divergence can indicate significant evolutionary changes
\citep{Jordan05}. Investigating evolutionary conservation and
divergence will potentially lead to a deeper understanding of what
makes each species unique.  Evolutionary changes primarily target the
structure and sequence of genomic DNA. However, not all changes will
lead to phenotypic differences.  On the other hand, sequence
similarity is not a guarantee of functional similarity because small
changes in DNA can potentially have remarkable functional
implications.

Therefore, in addition to investigating {\it structural conservation}
of the genes at the sequence level, another level of investigation is
needed to study {\it functional conservation} of the genes and their
regulation, which is reflected at the transcriptome \citep{Jimenez02,
  Jordan05}. Transcriptional regulation of the genes is a key
regulatory mechanism that can have remarkable phenotypic consequences
in highly modular cell-biological systems \citep{Hartwell99} even when
the original function of the regulated genes would remain intact.

Systematic comparison of transcriptional activity between different
species would provide a straightforward strategy for investigating
conservation of gene regulation \citep{Bergmann04, Enard02, Zhou04}.
However, direct comparison of individual genes between species may not
be optimal for discovering subtle and complex dependency structures.
The associative clustering principle (AC), introduced in
Publications~\ref{ECML}-\ref{AC}, provides a framework for detecting
groups of orthologous genes with exceptional levels of conservation
and divergence in transcriptional activity between two species. While
standard dependency detection methods for continuous data, such as the
generalized singular value decomposition \citep[see e.g.][]{Alter03}
or canonical correlation analysis \citep{Hotelling36} detect global
linear dependencies between observations, AC searches for dependent,
local groupings to reveal gene groups with exceptional levels of
conservation and divergence in transcriptional activity.  The model is
free of particular distributional assumptions about the data, which
helps to allocate modeling resources to detecting dependent subgroups
when variation within each group is less relevant for the analysis.
The remainder of this section provides an overview of the associative
clustering principle and its application to studying evolutionary
divergence between species.

\begin{figure}[t]
\label{fig:acprinciple}
\centerline{
\includegraphics[width=0.9\textwidth]{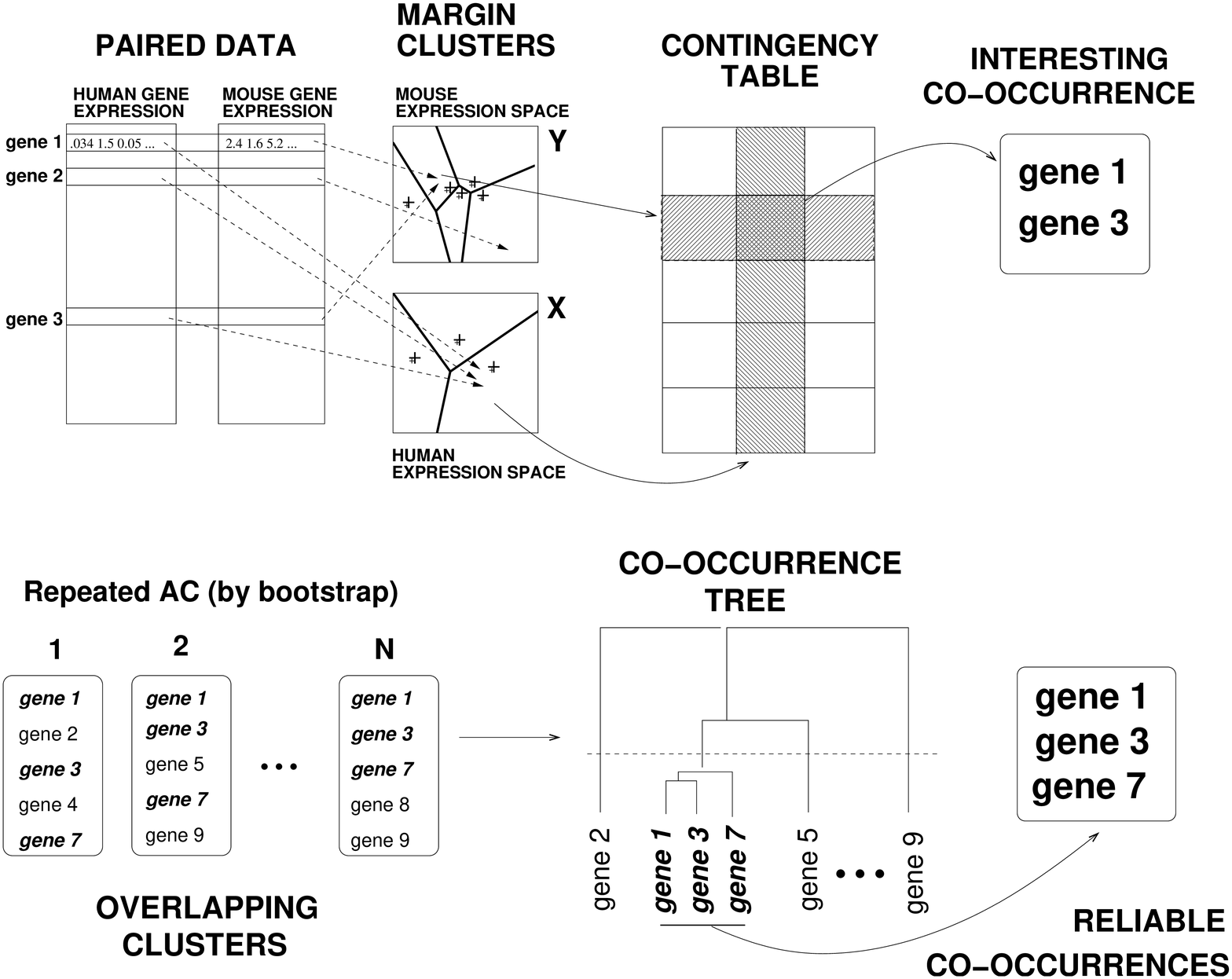}}
\caption{Principle of associative clustering (AC). AC performs
  simultaneous clustering of two data sets, consisting of paired
  observations, and seeks to maximize the dependency between the two
  sets of clusters.  The clusters are defined by cluster centroids in
  each data space.  The clustering results are represented on a
  contingency table, where clusters of the two data sets correspond
  with the rows and columns of the contingency table,
  respectively. These are called the margin clusters of the
  contingency table. The table cells are called cross clusters and
  they contain orthologous genes from the two data sets.  The cluster
  centroids are optimized to produce a contingency table with maximal
  dependency between the margin cluster counts.  Cross clusters that
  show significant deviation from the null hypothesis of independent
  margins indicate dependency. In order to enhance the reliability of
  the results, the clustering is repeated with slightly differing
  bootstrap samples. Then reliable co-occurrences are identified from
  a co-occurrence tree with a specified threshold. Frequently
  co-occurring orthologues are selected for further analyzes.}
\end{figure}

\subsubsection{The associative clustering principle}

The principle of associative clustering (AC) is illustrated in
Figure~\ref{fig:acprinciple}. AC performs simultaneous clustering of
two data sets to reveal maximally dependent cluster structure between
two sets of observations. The clusters are defined in each data space
by {\it Voronoi parameterization}, where the clusters are defined by
cluster centroids to produce connected, internally homogeneous
clusters.  Let us denote the two sets of clusters by
\(\{V^{(x)}_i\}_i\), \(\{V^{(y)}_j\}_j\).  A given data point \(\x\)
is then assigned to the cluster corresponding to the nearest centroid
\(\m_i\) in the feature space, with respect to a given distance
measure\footnote{\(\x \in V_i^{(x)}\) if \(d(\x, \m_i) \leq d(\x,
  \m_k)\) for all \(k\).}  \(d\).  This divides the space into
non-overlapping {\em Voronoi regions}. The regions define a clustering
for all points of the data space. The association between the clusters
of the two data sets can be represented on a contingency table, where
the rows and columns correspond to clusters in the first and second
data set, respectively. The clusters in each data set are called {\it
  margin clusters}. Each pair of co-occurring observations $(\xi,\yi)$
maps to one margin cluster in each data set, and each contingency
table cell corresponds to a pair of margin clusters. These are called
{\it cross clusters}.

AC searches for a maximally dependent cluster structure by optimizing
the Voronoi centroids in the two data spaces in such a way that the
dependency between the contingency table margins is maximized. Let us
denote the number of samples in cross cluster \(i, j\) by
\(n_{ij}\). The corresponding margin cluster counts are $n_{i \cdot} =
\sum_j n_{ij}$ and $n_{\cdot j} = \sum_i n_{ij}$.  The observed sample
frequencies over the contingency table margins and cross-clusters are
assumed to follow multinomial distribution with latent parameters
\(\bth_i, \bth_j\) and \(\bth_{ij}\), respectively.  Assuming the
model $M_I$ of {\it independent margin clusters}, the expected sample
frequency in each cross cluster is given by the outer product of
margin cluster frequencies.  The model $M_d$ of \emph{dependent margin
  clusters} deviates from this assumption. The {\it Bayes factor (BF)}
is used to compare the two hypotheses of dependent and independent
margins. This is a rigorously justified approach for model comparison,
which indicates whether the observations provide superior evidence for
either model. Evidence is calculated over all potential values of the
model parameters, marginalized over the latent frequencies. In a
standard setting, the Bayes factor would be used to compare evidence
between the dependent and independent margin cluster models for a
given clustering solution. AC uses the Bayes factor in a non-standard
manner; as an objective function to maximize by optimizing the cluster
centroids in each data space; the centroids define the margin clusters
and consequently the margin cluster dependencies.

The centroids are optimized with a conjugate-gradient algorithm after
smoothing the cluster borders with continuous parameterization. The
hyperparameters $n^{(d)}$, $n^{(x)}$, and $n^{(y)}$ arise from
Dirichlet priors of the two multinomial models \(M_I\), \(M_D\) of
independent and dependent margins, respectively.  Setting the
hyperparameters to unity yields the classical hypergeometric measure
of contingency table dependency \citep{Fisher34,Yates34}. With large
sample size, the logarithmic Bayes factor approaches mutual
information \citep{Sinkkonen05tr}. The Bayes factor is a desirable
choice especially with a limited sample size since a marginalization
over the latent variables makes it robust against uncertainty in the
parameter values, and because finite contingency table counts would
give a biased estimate of mutual information.  The number of clusters
in each data space is specified in advance, typically based on the
desired level of resolution. Nonparametric extensions, where the
number of margin clusters would be inferred automatically from the
data form one potential topic for further studies; a closely related
approach was recently proposed in \cite{Rogers2010}.

Publication~\ref{AC} introduces an additional, bootstrap-based
procedure to assess the reliability of the findings
(Figure~\ref{fig:acprinciple}). The analysis is repeated with similar,
but not identical training data sets obtained by sampling the original
data with replacement. The most frequently detected dependencies are
then investigated more closely. The analysis will emphasize findings
that are not sensitive to small variations in the observed data.

\subsubsection{Comparison methods}

Associative clustering was compared with two alternative methods:
standard K-means on each of the two data sets, and a combination of
K-means and information bottleneck (K-IB). K-means \citep[see
e.g.][]{Bishop06} is a classical clustering algorithm that provides
homogeneous, connected clusters based on Voronoi
parameterization. Homogeneity is desirable for interpretation, since
the data points within a given cluster can then be conveniently
summarized by the cluster centroid. On the other hand, K-means
considers each data set independently, which is suboptimal for the
dependency modeling task. The two sets of clusters obtained by
K-means, one for each data space, can then be presented on a
contingency table as in associative clustering.  The second comparison
method is K-IB introduced in Publication~\ref{ECML}. K-IB uses K-means
to partition the two co-occurring, continuous-valued data sets into
discrete atomic regions where each data point is assigned in its own
singleton cluster. This gives two sets of atomic clusters that are
mapped on a large contingency table, filled with frequencies of
co-occurring data pairs $(\x_k, \y_k)$. The table is then compressed
to the desired size by aggregating the margin clusters with the
symmetric IB algorithm in order to maximize the dependency between the
contingency table margins \citep{Friedman01}.  Aggregating the atomic
clusters provides a flexible clustering approach, but the resulting
clusters are not necessarily homogeneous and they are therefore
difficult to interpret.

\begin{figure}[t]
\centerline{
\begin{tabular}{cc}
{\bf A}&{\bf B}\\
\includegraphics[width=0.4\textwidth]{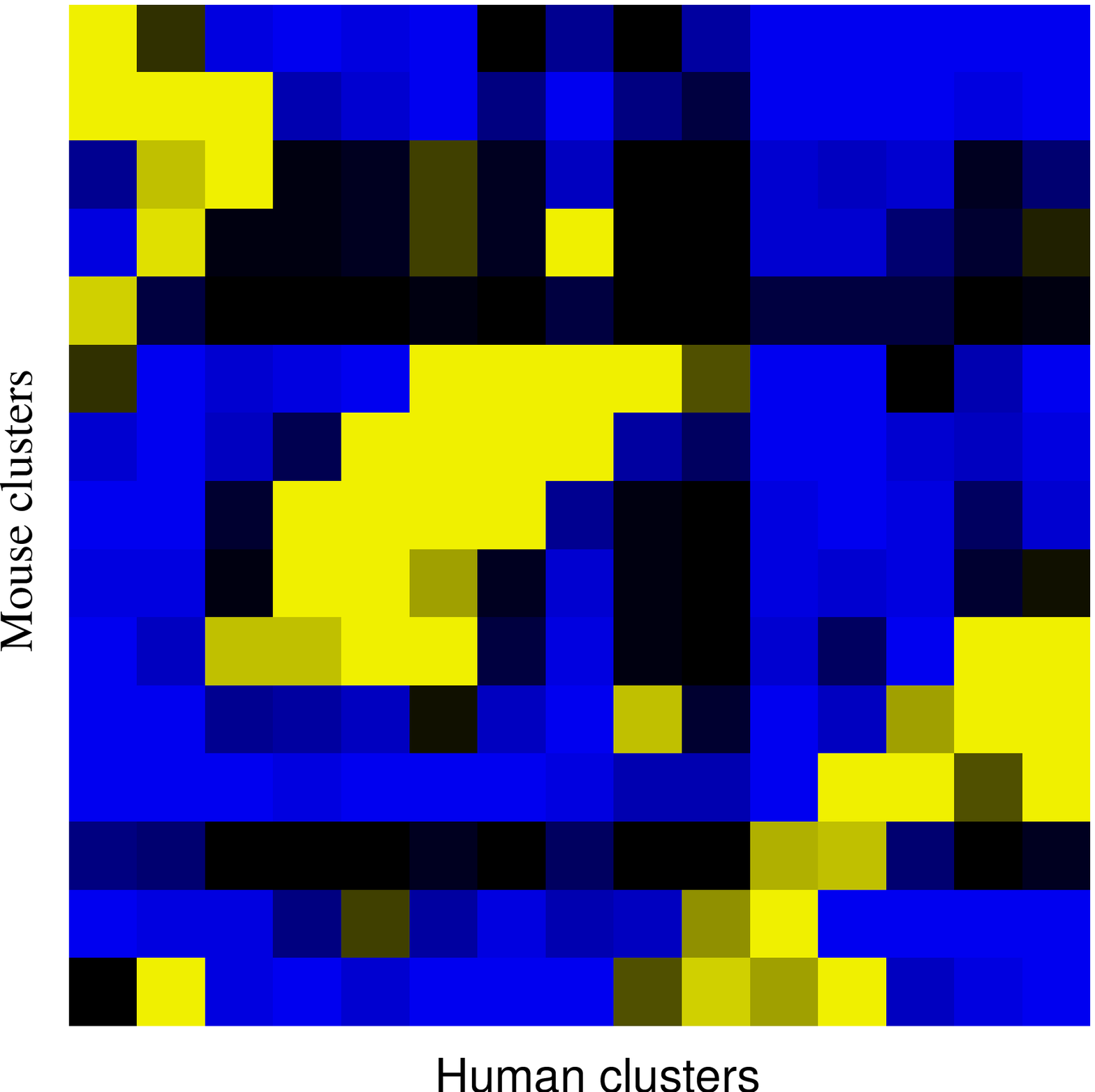}& 
\includegraphics[width=0.5\textwidth]{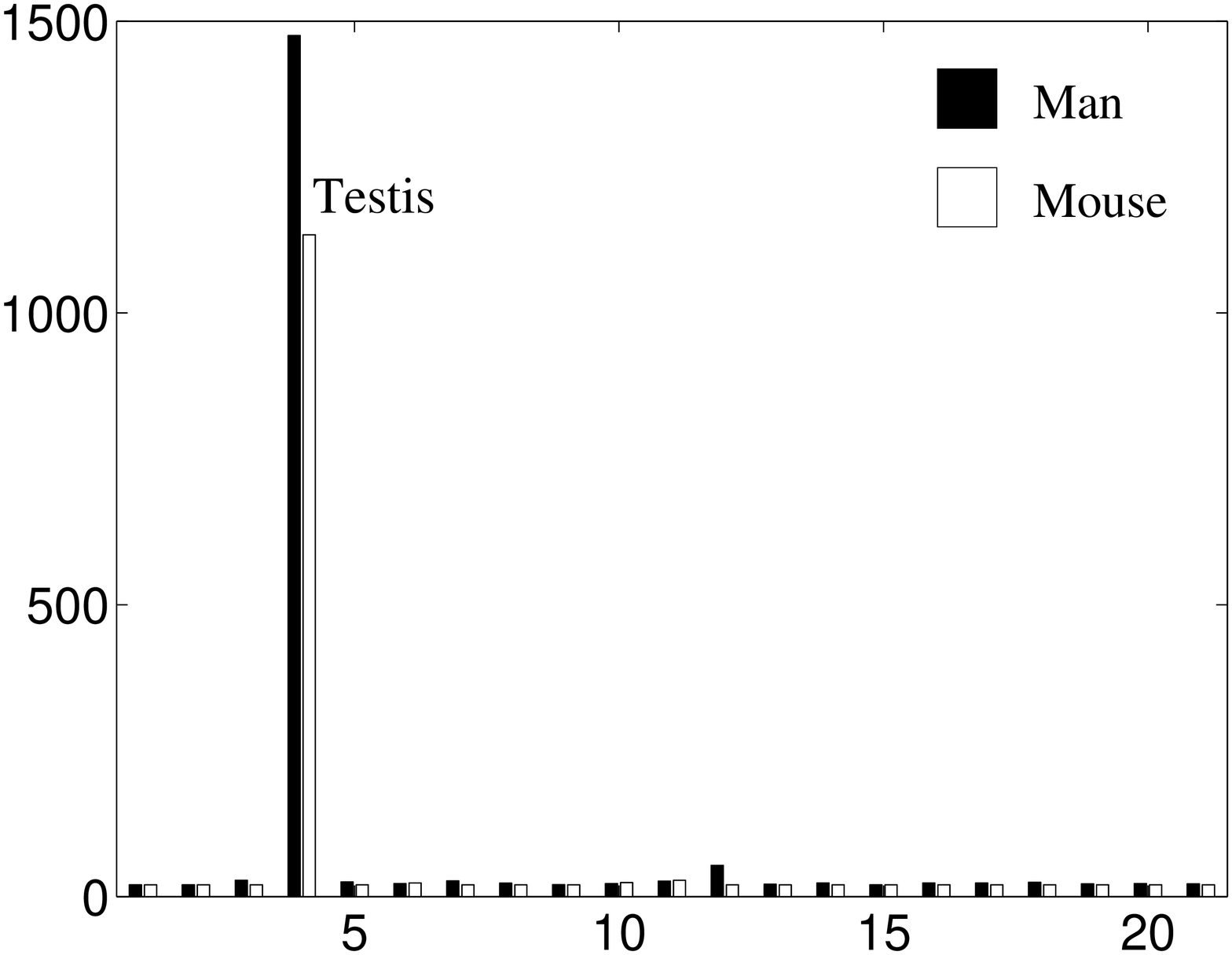}
\end{tabular}
}
\caption{{\bf A} The contingency table of associative clustering
  highlights orthologous gene groups in human (rows) and mouse
  (columns) with exceptional levels of conservation (yellow) or
  divergence (blue) in transcriptional activity between the two
  species. {\bf B} Average expression profiles of a highly conserved
  group of testis-specific genes across 21 tissues in man and
  mouse. \copyright IEEE. Reprinted with permission from
  Publication~\ref{AC}.}
\label{fig:ctable}
\end{figure}

AC compared favorably to the other methods.  While AC outperformed the
standard K-means in dependency modeling, the cluster homogeneity was
not significantly reduced in AC. The cross clusters from K-IB
\citep{Sinkkonen03tr} were more dependent than in AC. On the other
hand, AC produced more easily interpretable localized clusters, as
measured by the sum of intra-cluster variances in
Publication~\ref{AC}. Homogeneity makes it possible to summarize
clusters conveniently, for instance by using the mean expression
profiles of the cluster samples, as in Figure~\ref{fig:ctable}B.
While K-means searches for maximally homogeneous clusters and K-IB
searches for maximally dependent clusters, AC finds a successful
compromise between the goals of dependency and homogeneity.

\subsection{Exploratory analysis of transcriptional divergence between
  species}

Associative clustering is used in Publications~\ref{ECML} and~\ref{AC}
to investigate conservation and divergence of transcriptional activity
of 2818 orthologous human-mouse gene pairs across an organism-wide
collection of transcriptional profiling data covering 46 and 45 tissue
types in human and mouse, respectively \citep{Su02}. AC takes as input
two gene expression matrices with orthologous genes, one for each
species, and returns a dependency-maximizing clustering for the
orthologous gene pairs.  Interpretation of the results focuses on
unexpectedly large or small cross clusters revealed by the contingency
table analysis of associative clustering. Compared to plain
correlation-based comparisons between the gene expression profiles, AC
can reveal additional cluster structure, where genes with similar
expression profiles are clustered together, and associations between
the two species are investigated at the level of such detected gene
groups. The dependency between each pair of margin clusters can be
characterized by comparing the respective margin cluster centroids
that provide a compact summary of the samples within each cluster.

Biological interpretation of the findings, based on enrichment of Gene
Ontology (GO) categories \citep{Ashburner00}, revealed genes with
strongly conserved and potentially diverged transcriptional
activity. The most highly enriched categories were associated with
ribosomal functions, the high conservation of which has also been
suggested in earlier studies \citep{Jimenez02}; ribosomal genes often
require coordinated effort of a large group of genes, and they
function in cell maintenance tasks that are critical for species
survival.  An exceptional level of conservation was also observed in a
group of testis-specific genes, yielding novel functional hypotheses
for certain poorly annotated genes within the same cross-cluster
(Figure~\ref{fig:ctable}). Transcriptional divergence, on the other
hand, was detected for instance in genes related to embryonic
development.

While general-purpose dependency exploration tools may not be optimal
for studying the specific issue of transcriptional conservation, such
tools can reveal dependency with minimal prior knowledge about the
data. This is useful in functional genomics experiments where little
prior knowledge is available. In Publications~\ref{ECML} and~\ref{AC},
associative clustering has been additionally applied in investigating
dependencies between transcriptional activity and transcription factor
binding, another key regulatory mechanism of the genes.

\section{Conclusion}

The models introduced in Publications~\ref{MLSP}-\ref{AC} provide
general exploratory tools for the discovery and analysis of
statistical dependencies between co-occurring data sources and tools
to guide modeling through Bayesian priors. In particular, the models
consider linear dependencies (Publication~\ref{MLSP}) and
cluster-based dependency structures (Publications~\ref{ECML}-\ref{AC})
between the data sources. The models are readily applicable to data
integration tasks in functional genomics. In particular, the models
have been applied to investigate dependencies between chromosomal
mutations and transcriptional activity in cancer, and evolutionary
divergence of transcriptional activity between human and mouse.
Biomedical studies provide a number of other potential applications
for such general-purpose methods. An increasing number of co-occurring
observations across the various regulatory layers of the genome are
available concerning epigenetic mechanisms, micro-RNAs, polymorphisms
and other genomic features \citep{tcga08}. Simultaneous observations
provide a valuable resource for investigating the functional
properties that emerge from the interactions between the different
layers of genomic information. An open source implementation in
BioConductor\footnote{http://www.bioconductor.org/packages/release/bioc/html/pint.html}
provides accessible computational tools for related data integration
tasks, helping to guarantee the utility of the developed models for
the computational biology community.

\newpage

\chapter{Summary and conclusions}
\label{ch:conclusion}

\begin{quotation}
  \emph{Mathematics is biology's next microscope, only better; biology
    is mathematics' next physics, only better.}
\begin{flushright}
J.E. Cohen (2004)
\end{flushright}
\end{quotation}

Following the initial sequencing of the human genome \citep{Lander01,
  Venter01}, the understanding of structural and functional
organization of genetic information has extended rapidly with the
accumulation of research data. This has opened up new challenges and
opportunities for making fundamental discoveries about living
organisms and creating a holistic picture about genome
organization. The increasing need to organize the large volumes of
genomic data with minimal human intervention has made computation an
increasingly central element in modern scientific inquiry. It is a
paradox of our time that the historical scale of data in public and
proprietary repositories is only revealing how incomplete our
knowledge of the enormous complexity of living systems is. The
particular challenges in data-intensive genomics are associated with
the complex and poorly characterized nature of living systems, as well
as with limited availability of observations.  It is possible to solve
some of these challenges by combining statistical power across
multiple experiments, and utilizing the wealth of background
information in public repositories. Exploratory data analysis can help
to provide research hypotheses and material for more detailed
investigations based on large-scale genomic observations when little
prior knowledge is available concerning the underlying phenomena;
models that are robust to uncertainty and able to automatically adapt
to the data, can facilitate the discovery of novel biological
hypotheses. Statistical learning and probabilistic models provide a
natural theoretical framework for such analysis.

In this thesis, general-purpose exploratory data analysis methods have
been developed for organism-wide analysis of the human transcriptome,
a central functional layer of the genome. Integrating evidence across
multiple sources of genomic information can help to reveal mechanisms
that could not be investigated based on smaller and more targeted
experiments; this is a central aspect in all contributions.  In
particular, methods have been developed (i) in order to improve
measurement accuracy of high-throughput observations, (ii) in order to
model transcriptional activation patterns and tissue relatedness in
genome-wide interaction networks at an organism-wide scale, and (iii)
in order to integrate measurements of the human transcriptome with
other layers of genomic information. These results contribute to some
of the 'grand challenges' in the genomic era by developing strategies
to understand cell-biological systems, genetic contributions to human
health and evolutionary variation \citep{Collins03}. The computational
experiments in this thesis have been carried out based on publicly
available, anonymized data sets that follow commonly accepted ethical
standards in biomedical research. Open access implementations of the
key algorithms have been provided to guarantee wide access to these
tools and to spark new research beyond the original applications
presented in this thesis.

Methodological extensions and application of the developed algorithms
to new data integration tasks in functional genomics and in other
fields provide a promising line for future studies. The methods
developed in this thesis are readily applicable in genome-wide
screening studies in cancer and potentially other diseases. Increasing
amounts of co-occurring data concerning various aspects of the genome
have become available, including gene- and micro-RNA expression,
structural variation in the DNA, epigenetic modifications and gene
regulatory networks. It is expected that with small modifications the
introduced methodology can be applied to study further associations
between these and other layers of genome organization, as well as
their contributions to human health. The fundamental research
challenges in contemporary genome biology provide a wide array of
applications for statistical learning and exploratory analysis, and a
rich source of ideas for methodological research.



\chapter*{}\label{ch:references}
\addcontentsline{toc}{chapter}{References}
%
{\small

}



\newpage


\end{document}